\documentclass[journal]{IEEEtran}
\usepackage[style=ieee, maxbibnames=6, minbibnames=1]{biblatex}
\usepackage{graphicx}
\usepackage{algorithm}
\graphicspath{{figures/},{authors/}}
\DeclareGraphicsExtensions{.pdf,.jpg,.png}

\usepackage{physics}
\usepackage{subfigure}
\usepackage{amsfonts}
\usepackage{amsmath}
\usepackage{amssymb}
\usepackage{bbding}
\usepackage{tabularx}
\usepackage{listings}
\usepackage{xcolor}
\usepackage{tcolorbox}

\usepackage{siunitx}
\usepackage{colortbl} 
\usepackage{xcolor,pifont}
\usepackage{diagbox}
\usepackage{longtable}
\usepackage{booktabs}
\usepackage{mypythonhighlight}
\usepackage{makecell}
\usepackage{multirow}
\usepackage{algorithm}
\usepackage[colorlinks=true,linkcolor=blue,citecolor=blue]{hyperref} %
\usepackage{xspace}
\usepackage{algpseudocode}
\usepackage{algpseudocodex}

\usepackage{tcolorbox}
\tcbset{colback=white}
\colorlet{titleblue}{blue!80!black}
\colorlet{titlered}{red!80!black}
\colorlet{titlegreen}{green!80!black}

\usepackage[edges]{forest}
\usepackage{tikz}
\tikzstyle{my-box}=[
    rectangle,
    draw=hidden-draw,
    rounded corners,
    text opacity=1,
    minimum height=1.5em,
    minimum width=5em,
    inner sep=2pt,
    align=center,
    fill opacity=.5,
]
\tikzstyle{leaf}=[my-box, minimum height=1.5em,
    fill=hidden-blue!60, text=black, align=left,font=\scriptsize,
    inner xsep=2pt,
    inner ysep=4pt,
]
\definecolor{hidden-draw}{RGB}{0,51,102}
\definecolor{hidden-blue}{RGB}{194,232,247}
\definecolor{hidden-orange}{RGB}{243,202,120}
\definecolor{hidden-yellow}{RGB}{242,244,193}
\definecolor{tree-level-1}{RGB}{245,20,85}
\definecolor{tree-level-2}{RGB}{246,86,118}
\definecolor{tree-level-3}{RGB}{248,177,193}
\definecolor{tree-leaf}{RGB}{176,230,198}

\definecolor{Self}{RGB}{255,0,128}
\definecolor{Ensemble}{RGB}{0,127,255}
\definecolor{Iterative}{RGB}{153,51,255}

\definecolor{exemplar1}{RGB}{136,98,148}
\definecolor{exemplar2}{RGB}{148,210,242}
\definecolor{knowledge1}{RGB}{249,219,152}
\definecolor{knowledge2}{RGB}{255,245,220}

\usepackage{pifont}

\newcommand{\second}[1]{{\color{black}{#1}}}

\usepackage[capitalize]{cleveref}
\crefname{section}{Sec.}{Secs.}
\Crefname{section}{Section}{Sections}
\Crefname{table}{Table}{Tables}
\crefname{table}{Tab.}{Tabs.}

\newcommand{\first}[1]{{\color{black}{#1}}}

\ifCLASSINFOpdf
\else
\fi

\begin{document}
%

\title{\first{LLM4AD:} Large Language Models for Autonomous Driving — Concept, \first{Review,} Benchmark, Experiments, and \first{Future Trends}}

%

\author {Can~Cui,~\IEEEmembership{Student Member,~IEEE,}
        Yunsheng~Ma,~\IEEEmembership{Student Member,~IEEE,}
        Sung-Yeon~Park,
        Zichong~Yang, 
        Yupeng Zhou,~\IEEEmembership{Student Member,~IEEE,}
        Peiran Liu,
        Juanwu Lu,~\IEEEmembership{Student Member,~IEEE,}
        Juntong Peng,~\IEEEmembership{Student Member,~IEEE,}
        Jiaru Zhang,
        Ruqi Zhang,
        Lingxi~Li,~\IEEEmembership{Senior Member,~IEEE,}
        Yaobin~Chen,~\IEEEmembership{Senior Member,~IEEE,}
        Jitesh~H.~Panchal,
        Amr~Abdelraouf,~\IEEEmembership{Member,~IEEE,}
        Rohit~Gupta,~\IEEEmembership{Member,~IEEE,}
        Kyungtae~Han,~\IEEEmembership{Senior Member,~IEEE,}
        and~Ziran~Wang,~\IEEEmembership{Senior Member,~IEEE}


%
%

\thanks{C. Cui and Y. Ma contributed equally to this work.

C. Cui, Y. Ma, S. Park, Z. Yang, Y. Zhou, P. Liu, J. Lu, J. Peng, L. Li, Y. Chen, J. H. Panchal, and Z. Wang are with the College of Engineering, Purdue University, West Lafayette, IN 47907, USA.

J. Zhang is with the Institute for Physical Artificial Intelligence (IPAI), Purdue University, West Lafayette, IN 47907, USA.

R. Zhang is with the Department of Computer Science, Purdue University, West Lafayette, IN 47907, USA.

A. Abdelraouf, R. Gupta, and K. Han are with the InfoTech Labs, Toyota Motor North America, Mountain View, CA 94043, USA.

Corresponding author: Ziran Wang, e-mail: ziran@ieee.org.

Copyright (c) 2025 IEEE. Personal use of this material is permitted. However, permission to use this material for any other purposes must be obtained from the IEEE by sending a request to pubs-permissions@ieee.org.

}}

\markboth{Journal of \LaTeX\ Class Files,~Vol.~14, No.~8, August~2015}%
{Shell \MakeLowercase{\textit{et al.}}: Bare Demo of IEEEtran.cls for IEEE Journals}
%



\maketitle

\begin{abstract}

With the broader adoption and highly successful development of Large Language Models (LLMs), there has been growing interest and demand for applying LLMs to autonomous driving technology. Driven by their natural language understanding and reasoning capabilities, LLMs have the potential to enhance various aspects of autonomous driving systems, from perception and scene understanding to interactive decision-making. This paper first introduces the novel concept of designing Large Language Models for Autonomous Driving (LLM4AD), \first{followed by a review of existing LLM4AD studies}. Then, a comprehensive benchmark is proposed for evaluating the instruction-following and reasoning abilities of LLM4AD systems, which includes LaMPilot-Bench, CARLA Leaderboard 1.0 Benchmark in simulation \first{and NuPlanQA for multi-view visual question answering}. Furthermore, extensive real-world experiments are conducted on autonomous vehicle platforms, examining both on-cloud and on-edge LLM deployment for personalized decision-making and motion control. \first{Next, the future trends of integrating language diffusion models into autonomous driving are explored, exemplified by the proposed ViLaD (Vision-Language Diffusion) framework.} Finally, the main challenges of LLM4AD are discussed, including latency, deployment, security and privacy, safety, trust and transparency, and personalization.

\end{abstract}


%
\IEEEpeerreviewmaketitle




\section{Introduction}

\subsection{Background}
The recent development of Large Language Models (LLMs) has enhanced numerous cutting-edge technologies, introducing significant advancements across a broad spectrum of applications~\cite{zhao2023survey,wang2023survey}. Their applications span from classic natural language processing (NLP) tasks, such as document modification and information extraction, to emerging scenarios like LLM-based agents and evaluation frameworks~\cite{zhao2023survey}. 

A prominent application is the adoption of  LLMs for autonomous driving (LLM4AD). In this area, various advanced algorithms are continually enhancing autonomous driving technology, leveraging the potential of LLMs to drive innovation and efficiency. These models can contribute to autonomous systems ranging from high-level decision-making processes to meticulous low-level control.

On the high-level side, LLMs can actively engage in adjusting driving modes or decision-making process~\cite{cui_drive_2024}. Consider a scenario where a passenger abstractly expresses urgency, such as ``I do not want my friends to wait for me." An LLM4AD system can interpret these emotions and adjust the vehicle's control strategies accordingly, aligning with the current driving mood. By contrast, non-LLM-based systems lack the ability to accurately comprehend or interpret human intentions derived soly from vague expressions~\cite{cui2023large}. Furthermore, LLM-based systems offer significant potential for personalization due to their continuous learning capability. This enables the system to adapt to individual preferences, improving the driving experience for different users. Additionally, LLMs hold the potential to develop knowledge-driven systems capable of driving like experts by accumulating experience through continuous operation~\cite{fu_drive_2024}.

On the low-level side, LLMs play a crucial role in the tuning and control process. They have demonstrated the ability to analyze specific scenarios and convert gathered information into mathematical representations to guide low-level controllers~\cite{sha_languagempc_2023}. Additionally, LLMs can receive input data from the controller and provide performance updates to aid in analyzing the control loop's effectiveness~\cite{s23135899}, potentially suggesting improvements or detecting issues to enhance overall performance.

Beyond applying LLMs to conventional modular autonomous systems, they also play a significant role in end-to-end systems. By leveraging powerful reasoning and multimodal understanding capabilities, LLMs can directly process sensory data and ego-vehicle states to generate planning trajectories and fine-grained control actions~\cite{chen2024end}. This LLM-driven paradigm has attracted increasing attention from both industry~\cite{driveirl,hu2023imitation,pmlr-v164-scheel22a} and academia~\cite{hu_planning-oriented_2023}. Unlike conventional modular paradigms that handle perception, prediction, planning, and control separately, the LLM-based end-to-end system is fully differentiable and optimizes the entire driving process simultaneously.

The emerging developments in LLM4AD raise crucial questions: Why have LLMs become popular in autonomous driving, and what advantages do they offer over systems without LLM integration? Based on current research, compelling advantages of integrating LLMs include:

\begin{itemize}
\item \textbf{Intuitive Language Interaction:} LLMs enable intuitive communication between humans and vehicles. Users can express abstract commands and feelings, while the LLM accurately captures the intentions behind these expressions.

\item \textbf{Contextual Understanding and Reasoning:} LLMs provide contextual understanding from diverse sources, such as traffic rules and accident reports, allowing generated decisions to ensure safety and adherence to local regulations.

\item \textbf{Zero-Shot and Few-Shot Planning:} Zero-shot generalization enables LLMs to perform tasks on which they have not been explicitly trained~\cite{kojima_large_2022}. Consequently, LLM-based systems can handle uncommon situations with minimal prior experience, enabling confident navigation through ``corner cases.''

\item \textbf{Continuous Learning and Personalization:} LLMs learn and adapt continuously, facilitating the ability to follow individual user preferences and improve the driving experience over time.

\item \textbf{Interpretability and Trust:} LLMs can explain decisions in natural language, thereby improving trust and transparency between autonomous systems and users.
\end{itemize}

Despite these advantages, LLMs possess limitations. A primary concern is their applicability to real-time tasks; LLMs typically require seconds to process textual information. This latency poses safety risks in situations demanding immediate decision-making. Additionally, preventing hallucinations, where LLMs generate factually incorrect or nonsensical outputs, remains a challenge. As autonomous driving is safety-critical, hallucinations can undermine system reliability and user trust. Furthermore, data privacy and security present issues, as LLMs collect and process vast amounts of textual data, including potentially sensitive information regarding surroundings, passengers, and driving preferences.

\subsection{Overview of This Article}
The remainder of this paper is organized as follows: \first{\cref{sec:concept} introduces the concept and key elements of the proposed LLM4AD framework. \cref{sec:soa} reviews existing works related to the integration of LLMs into autonomous driving systems.} \cref{sec:benchmark} establishes open benchmarks designed to evaluate the performance of LLMs in various tasks, including datasets, evaluation metrics, and baseline models. \cref{sec:real-vehicle} and \cref{sec:onboard} present the implementation of the framework on a real-vehicle platform, discussing the results of extensive experiments conducted to validate driving and personalization performance. \cref{sec:vilad} explores future trends in leveraging large language diffusion models for autonomous driving. \cref{sec:discussion} analyzes current challenges and outlines promising research directions, and finally, \cref{sec:conclusion} summarizes the key contributions and findings of this work.

\begin{figure*}[!t]
    \centering
     \includegraphics[width=0.85\linewidth]{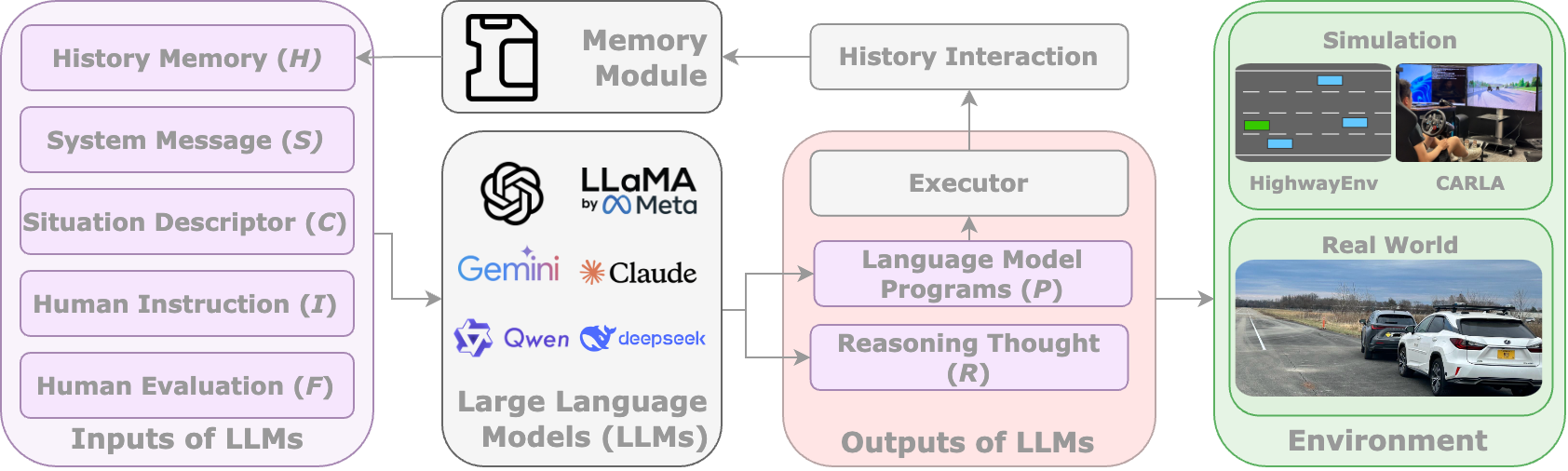}
    \caption{The conceptual illustration of the proposed LLM4AD system. The process begins with five distinct inputs being fed into the core large language models: historical memory ($H$) retrieved from a memory module, system messages ($S$) that define the role and constraints, a situation descriptor ($C$) providing context, human instructions ($I$), and human evaluations ($F$). These inputs are processed by models such as LLaMA, Gemini, or Claude to generate two primary outputs: reasoning thoughts ($R$) that explain the decision-making process and language model programs ($P$) which serve as executable policies. An executor then acts as a bridge, translating these programs into actions within the environment, which can range from simulators like CARLA and HighwayEnv to real-world autonomous driving platforms. Finally, the system incorporates a feedback loop where historical interactions are stored back into the memory module to support continuous learning and personalization.}
    \label{fig:concept_framework}
\end{figure*}



\section{Concept of LLM4AD}
\label{sec:concept}
This section introduces a perspective in which LLMs function as the decision-making ``brain" within the autonomous driving system. In the proposed framework, LLMs do not directly alter perception or localization modules (the vehicle's ``eyes"); instead, they utilize outputs from these modules as references to guide high-level decision-making. By processing this data, LLMs enhance informed decision-making, leading to significant performance improvements. Downstream, the vehicle's control module serves as the ``hands," executing the driving policy derived from the LLM-based process.


The overall LLM4AD framework is illustrated in \cref{fig:concept_framework}. Humans provide instructions $I$ and evaluations/feedback $F$. These, along with historical memories $H$, system messages $S$, and context information $C$, serve as inputs to the LLMs. The memory module stores the corresponding historical interactions $H$ for different users. Upon receiving these inputs, the LLMs engage in reasoning to produce outputs, including generated Language Model Programs (LMPs) or policies $P$ and reasoning thoughts $R$. The policy $P$ is sent to the executor for implementation in the environment, while the reasoning thoughts assist the LLMs in generating more logical driving strategies. While specific implementations may vary across applications, they all adhere to the principles of this general concept.

\subsection{Human Instruction and Evaluation}
Human instructions $I$ and evaluations $F$ are fed directly into the LLMs in natural language. $I$ represents the human's desired objectives for the autonomous agent, while the evaluation $F$ provides feedback on the effectiveness of the executed driving policies.


\subsection{System Message}
System messages $S$ provide instructions and context to the LLM4AD system at the initiation of a conversation or task. They allow developers to clearly define the role, constraints, and objectives the system must follow. In autonomous driving tasks, $S$ acts as a set of high-level guidelines framing how LLMs should operate, defining basic driving logic such as task definitions, adherence to traffic rules, decision state descriptions, and optimization metrics. These messages serve as a foundational framework to prevent incorrect assumptions or unintended strategies.

\subsection{Situation Descriptor}
The situation descriptor translates the current driving context $C$ into structured natural language descriptions. Its objective is to equip LLMs with situational awareness and a comprehensive representation of the driving scenario — such as ``You are positioned in the leftmost lane of a two-way highway" or ``A vehicle is located 50 meters ahead" — enabling appropriate decision-making. This descriptor intuitively translates complex spatial and temporal relationships between road users and the ego vehicle into a format the LLM can reason about.

\subsection{History Memory and Memory Module}
The memory module stores user profiles to enhance the driving experience. Whenever a human utilizes the LLM4AD system, relevant historical interactions $H$ associated with that user are logged. Subsequently, this historical data is transmitted to the LLMs as input, serving as a reference point for user preferences and guiding the system to improve the user experience. Interaction data is updated in the corresponding profile within the memory module after each trip.

\subsection{Large Language Models}
LLMs serve as the core module of the framework, receiving all inputs (history interactions $H$, system message $S$, situation description $C$, human instruction $I$, and human feedback $F$) to generate textual outputs (Generated Policies/LMPs $P$ and reasoning thoughts $R$). Notably, chain-of-thought prompting~\cite{wei_chain--thought_2022} is employed as a guiding signal to ensure alignment with human-like reasoning and practical driving considerations.


\subsection{Generated Policy/Program}
Inspired by the concept of ``Code as Policy"~\cite{liang_code_2023}, a primary output of the LLMs is the generated Policy $P$, consisting of executable code. These codes influence the ego agent's driving behavior within the environment. They possess the ability to generalize to new natural language commands and provide precise numerical values (e.g., velocities) based on vague linguistic descriptions (e.g., ``hurry," ``turn left") depending on the driving context.

\subsection{Output Thoughts}
Through chain-of-thought prompting, LLMs generate not only the program code but also step-by-step explanations of the thought process behind the solution. These chains of thought $R$ elucidate the reasoning behind decisions (e.g., ``Since the command is `hurry,' I will increase the target velocity"). The output thoughts $R$ accompany the generated policies $P$, offering insights into how natural language commands are interpreted within the specific driving context to produce precise control values, thereby improving system transparency and interpretability.

\subsection{Executor}
The executor bridges the gap between the LLMs' textual outputs and the autonomous driving policy. It receives the generated policies/LMPs $P$ and executes them in the corresponding environment, allowing the code to interact with the ego vehicle's status and deploy the intended behaviors in real or simulated settings. Execution specifics may vary depending on the underlying autonomous driving system.


\section{Review of Existing LLM4AD Studies}
\label{sec:soa}

\begin{table*}[!t]
\caption{Overview of state-of-the-art approaches on the applications of large language models for autonomous driving.}
\resizebox{\textwidth}{!}{%
\begin{tabular}{lllllll}
\toprule
Framework & Method & (M)LLM & Input & Output & Evaluation\\
\midrule
DiLu~\cite{wen_dilu_2024,fu_drive_2024} & RAG & GPT-4 & Prompt & Action+NL & HighwayEnv~\cite{leurent_environment_2018}\\
DriveGPT4~\cite{xu_drivegpt4_2024} & SFT & LLaMA-2 & Image+Prompt & Signal+NL & BDD-X~\cite{kim2018textual} \\
DriveMLM~\cite{wang_drivemlm_2023} & SFT & LLaMA & Image+Point+Prompt & Action+NL & CARLA~\cite{dosovitskiy_carla_2017} \\
DriveVLM~\cite{tian_drivevlm_2024} & SFT & Qwen-VL & Image+Prompt & Action+Trajectory+NL & nuScenes~\cite{caesar_nuscenes_2020}/SUP-AD~\cite{tian_drivevlm_2024} \\
Driving with LLMs~\cite{chen_driving_2024} & SFT & LLaMA & Vector+Prompt & Signal+NL & DrivingQA~\cite{chen_driving_2024} \\
GPT-Driver~\cite{mao_gpt-driver_2023} & SFT & GPT-3.5 & Vector+Prompt & Action+Trajectory+NL & nuScenes~\cite{caesar_nuscenes_2020} \\
LaMPilot~\cite{ma_lampilot_2024} & RAG & GPT-4 & Prompt & Code+NL & HighwayEnv~\cite{leurent_environment_2018} \\
LangProp~\cite{ishida_langprop_2024} & IL/RL & GPT-3.5 & Prompt & Code & CARLA~\cite{dosovitskiy_carla_2017} \\
Language Agent~\cite{mao_language_2024} & RAG+SFT & GPT-3.5 & Prompt & Action+Trajectory+NL & nuScenes~\cite{caesar_nuscenes_2020} \\
LanguageMPC~\cite{sha_languagempc_2023} & Prompt Engineering & GPT-3.5 & Prompt & NL & IdSim~\cite{fitzgibbons_idsim_2004} \\
LLaDA~\cite{li_driving_2024} & Prompt Engineering & GPT-4 & Prompt & NL & nuScenes~\cite{caesar_nuscenes_2020} \\
LMDrive~\cite{shao_lmdrive_2024} & SFT & LLaVA-v1.5 & Image+Point+Prompt & Singal & CARLA~\cite{dosovitskiy_carla_2017} \\
RAG-Driver~\cite{yuan_rag-driver_2024} & RAG+SFT & Vicuna-v1.5 & Image+Vector+Prompt & Signal+NL & BDD-X~\cite{kim2018textual}/Spoken-SAX~\cite{yuan_rag-driver_2024} \\
Talk2Drive~\cite{cui_drive_2024} & Prompt Engineering & GPT-4 & Prompt & Code+NL & Real World\\
VLP~\cite{pan_vlp_2024} & SFT & CLIP & Image & Trajectory & nuScenes~\cite{caesar_nuscenes_2020} \\
DriveLM~\cite{sima_drivelm_2024} & SFT & BLIP-2~\cite{li_blip-2_2023} & Image & Trajectory+NL & DriveLM~\cite{sima_drivelm_2024} \\
Emma~\cite{hwang_emma_2024} & SFT & Gemini 1.0 Nano-1~\cite{gemini_team_gemini_2024} & Image+Prompt & Action+Trajectory+NL &nuScenes ~\cite{caesar_nuscenes_2020}/WOMD~\cite{chen2024womdlidarrawsensordataset} \\
OpenEmma~\cite{10972521} & Prompt Engineering& LLaVA~\cite{meta_ai_llama_nodate,liu_visual_2023,wang2024qwen2vlenhancingvisionlanguagemodels} & Image+Prompt & Action+Trajectory+NL &nuScenes ~\cite{caesar_nuscenes_2020} \\
Dolphins~\cite{ma_dolphins_2024} & SFT& OpenFlamingo~\cite{alayrac2022flamingovisuallanguagemodel} & Image+Prompt & Action+NL &Custom Dataset \\
EM-VLM4AD~\cite{gopalkrishnan2024multiframelightweightefficient} & SFT& T5~\cite{raffel2023exploringlimitstransferlearning} & Image+Prompt & NL &DriveLM~\cite{sima_drivelm_2024} \\
OpenDriveVLA~\cite{zhou2025opendrivevlaendtoendautonomousdriving} & SFT& Qwen2.5VL~\cite{wang2024qwen2vlenhancingvisionlanguagemodels} & Image+Prompt & Action+Trajectory+NL &nuScenes~\cite{caesar_nuscenes_2020,qian_nuscenes-qa_2024} \\
HiLM-D~\cite{ding2025hilmdenhancingmllmsmultiscale} &Prompt Engineering& miniGPT-4~\cite{zhu_minigpt-4_2023} & Image+Prompt & NL &DRAMA~\cite{malla_drama_2023} \\
AdaThinkDrive~\cite{luo2025adathinkdrive} &RL& InternVL3~\cite{chen_internvl_2024} & Image+Prompt & Trajctory+NL &NIVSIM~\cite{dauner2024navsimdatadrivennonreactiveautonomous} \\
Auto-Drive$R^2$~\cite{yuan2025autodriver2incentivizingreasoningselfreflection} &RL& Qwen2.5VL~\cite{bai2025qwen25vltechnicalreport} & Image+Prompt & Trajctory+NL &nuScenesR2-6K~\cite{yuan2025autodriver2incentivizingreasoningselfreflection} \\
ReCogDrive~\cite{li2025recogdrive} &RL& InternVL3~\cite{chen_internvl_2024}/QwenVL2.5~\cite{bai2025qwen25vltechnicalreport} & Image+Prompt & Trajctory &NIVSIM~\cite{dauner2024navsimdatadrivennonreactiveautonomous}/Ben2dRIVE~\cite{jia_bench2drive_2024}\\
ReasonPlan~\cite{liu2025reasonplanunifiedsceneprediction} &RL& Qwen~\cite{bai_qwen_2023} & Image+Prompt & Trajctory & Ben2Drive~\cite{jia_bench2drive_2024}\\
MTRDrive~\cite{luo2025mtrdrivememorytoolsynergisticreasoning} &RL& Qwen2.5VL~\cite{bai2025qwen25vltechnicalreport} & Image+Prompt & Trajctory+Action+NL & NAVSIM~\cite{dauner2024navsimdatadrivennonreactiveautonomous}\\
Poutine~\cite{rowe2025poutine} &RL& Qwen2.5VL~\cite{bai2025qwen25vltechnicalreport} & Image+Prompt & Trajctory+NL & Waymo Open Dataset~\cite{Ettinger_2021_ICCV}\\
AgentThink~\cite{qian2025agentthink} &RL& Qwen2.5VL~\cite{bai2025qwen25vltechnicalreport} & Image+Prompt & NL & DriveMLLM~\cite{guo2025surds}\\
DriveAgent-R1~\cite{zheng2025driveagent} &RL& Qwen2.5VL~\cite{bai2025qwen25vltechnicalreport} & Image+Prompt & Action+NL & Drive-Internal/nuScenes~\cite{caesar_nuscenes_2020}\\
Auto-VLA~\cite{zhou2025autovla} &RL& Qwen2.5VL~\cite{bai2025qwen25vltechnicalreport} & Image+Prompt & Trajectory+NL & NAVSIM ~\cite{dauner2024navsimdatadrivennonreactiveautonomous}\\
\bottomrule
\end{tabular}%
}
\label{tab:sota}
\end{table*}


\first{

\first{This section reviews existing studies that apply LLMs for autonomous driving, which are grouped into four aspects: learning method, pipeline design, evaluation methodology, and deployment platform. On the learning side, recent work spans supervised fine-tuning with parameter-efficient variants, prompt engineering strategies, retrieval-augmented generation, imitation learning, and reinforcement learning from human or environment feedback, reflecting a progression from static adaptation to closed-loop policy optimization. The pipeline discussion highlights how these frameworks select and combine input modalities (textual scene descriptions, camera images, LiDAR point clouds, and vectorized driving states) and map them to diverse output representations, including natural language explanations, discrete meta-actions, low-level control signals, trajectories, and even policy code. For evaluation, existing studies rely on language-augmented driving datasets, vision-language benchmarks, simulation environments, and a small but growing set of real-vehicle deployments, using both task-specific driving metrics and natural-language quality metrics, yet still lack a unified benchmark for LLM4AD. Finally, this section summarizes real-world deployment platforms and compares their capabilities, showing that most current testbeds specialize in specific autonomy components, while only very recent efforts begin to support full-stack validation tailored to LLM-integrated autonomous driving.}

\subsection{Learning Method}
\label{sec:learning-method}
An analysis of the learning methods used in recent studies reveals that supervised fine-tuning (SFT) and retrieval-augmented generation (RAG) are the most commonly employed techniques. Other approaches include prompt engineering, imitation learning (IL), and reinforcement learning (RL).

\subsubsection{Supervised Fine-Tuning}
Supervised fine-tuning (SFT) is a well-established technique in machine learning that adapts a pre-trained model to a downstream task by further training it on task-specific data. In this paper's context, SFT allows LLMs to leverage their world knowledge acquired during pre-training while specializing it for driving tasks. However, full-weight SFT is a highly resource-intensive process, especially for LLMs, requiring substantial computational power and memory to store and update the model's billions of parameters. 

Researchers have proposed various parameter-efficient fine-tuning (PEFT) techniques to address these challenges. One notable example is Low-Rank Adaptation (LoRA)~\cite{hu_lora_2022}, which adds a small set of trainable parameters called low-rank adapters to the pre-trained model. These adapters are optimized during SFT, while the original model parameters remain frozen. This approach significantly reduces the computation and memory requirements compared to full SFT, making it more feasible for resource-constrained cases. Moreover, LoRA helps prevent catastrophic forgetting~\cite{kirkpatrick_overcoming_2017}, where the model loses its previously acquired knowledge during SFT. QLoRA~\cite{dettmers_qlora_2023} is a commonly used extension of LoRA that works by quantizing the precision of the weight parameters in the pre-trained LLM to 4-bit precision.

Several studies have successfully applied fine-tuned LLMs to various autonomous driving tasks, demonstrating their potential for enhancing the capabilities of autonomous vehicles. For instance, DriveGPT4~\cite{xu_drivegpt4_2024} fine-tuned LLaMA-2~\cite{touvron_llama_2023} as the backbone model on a custom visual instruction-following dataset. This enables the proposed framework to process multimodal input data, such as images and text, and generate both text responses and predicted control signals for the vehicle. Language Agent~\cite{mao_language_2024} fine-tuned GPT-3.5 on the nuScenes dataset~\cite{caesar_nuscenes_2020}. The framework excels at multi-round reasoning, taking into account traffic rules, past driving experiences, and environmental information to devise trajectories for driving. LMDrive~\cite{shao_lmdrive_2024} fine-tuned LLaVA~\cite{liu_visual_2023} as the backbone model and proposed a framework to process multimodal, multi-view sensor data alongside natural language instructions. This enables end-to-end autonomous driving, where the model directly maps input data to control commands without the need for intermediate representations.

\subsubsection{Prompt Engineering}
Prompt engineering is another approach for LLM4AD tasks, which does not involve updating the model parameters. Instead, it involves carefully designing prompts to feed task-specific information to LLMs, guiding them to generate relevant outputs for the given task. Its flexibility and ease of use make it a popular option for researchers and practitioners, particularly when SFT is not feasible due to resource constraints or limited access to task-specific data. There are several common practices for prompt engineering~\cite{brown2020language,wei_chain--thought_2022}:

\paragraph{Zero-Shot Prompting}
In zero-shot prompting, LLMs are expected to generate outputs for new tasks without being provided any task-specific examples. This approach is highly flexible and requires minimal effort. Leveraging the zero-shot generalizability of LLMs, LLaDA~\cite{li_driving_2024} proposes a training-free mechanism to assist human drivers and adapt autonomous driving policies to new environments. However, zero-shot inference may not always produce reliable or accurate results, especially for complex driving tasks. For example, DiLu~\cite{wen_dilu_2024} demonstrates that in the HighwayEnv simulator, no tasks are successfully conducted (with the median success steps below five), indicating that LLMs cannot directly perform closed-loop driving tasks without any adaptation.

\paragraph{Few-Shot Prompting}
Few-shot prompting addresses the limitations of zero-shot prompting by including a small number of task-specific examples in the prompt. These examples serve as a reference for LLMs, allowing them to adapt their outputs to the target task. Few-shot prompting has been shown to improve the performance of LLMs on various driving tasks. For instance, LaMPilot~\cite{ma_lampilot_2024} reports an increase in the task completion rate from 20.4\% to 63.3\% when using LLaMA-2 as the LLM with three-shot prompting.

\paragraph{Chain-of-Thought Prompting}
Chain-of-Thought~\cite{wei_chain--thought_2022} prompting involves providing LLMs with few-shot exemplars explaining the reasoning process. By demonstrating the multi-step reasoning process in the exemplars, LLMs are encouraged to show a similar reasoning process when generating the completion. This explanation of reasoning often leads to more accurate results. Chain-of-thought prompting has been particularly effective for tasks that require complex reasoning, such as arithmetic reasoning, commonsense reasoning, symbolic reasoning, and decision-making in driving scenarios. For example, LanguageMPC~\cite{sha_languagempc_2023} proposes a chain-of-thought framework for LLMs in driving scenarios, which divides the analysis and decision-making process into numerous sub-problems. This enables LLMs to comprehensively engage in logical reasoning and arrive at informed decisions, which are then used to direct the low-level controller using high-level textual decisions. Dong et al.~\cite{dong2024generalizing} used a zero-shot LLM with CoT prompt in the training phase to provide instructions to the standalone end-to-end model, showing the feasibility in real-world deployment. Dolphins~\cite{ma_dolphins_2024} extends multimodal LLMs to driving by integrating grounded CoT for visual reasoning over images, commands, and control history, enabling a unified model to perform perception, scene understanding, trajectory planning, and reflective self-correction in open-ended AV tasks.

While prompt engineering has demonstrated effectiveness in various studies~\cite{sha_languagempc_2023,li_driving_2024}, it also has some limitations. One major challenge is the limited context length, which restricts the amount of task-specific information that can be provided to the model. Another issue is the potential for generating incorrect outputs, often referred to as ``hallucinations"~\cite{huang_survey_2022}. Even with carefully crafted prompts, LLMs may sometimes produce outputs that are inconsistent or unrelated to the given task, which can be problematic in safety-critical scenarios like autonomous driving.

\subsubsection{Retrieval-Augmented Generation (RAG)}
Retrieval-augmented generation (RAG)~\cite{lewis_retrieval-augmented_2020} is a technique for enhancing the accuracy and reliability of LLMs with facts fetched from external sources. Similar to prompt engineering, RAG leverages in-context learning to provide task-specific information to LLMs. 
The process of RAG typically involves two modules: (1) an \textit{embedding model}, such as OpenAI's \texttt{text-embedding-ada-002}~\cite{openai_embeddings_nodate}, is used to map documents and queries (in text strings) into a shared vector space, and (2) a \textit{vector database}, such as Chroma~\cite{chroma_ai-native_2023}, is used to index the knowledge documents. 

Specifically, for each query, the embedding model is used to generate a vector representation. This query vector is then compared to the indexed document vectors, and the top-k most similar documents are retrieved. These documents are included in the prompt along with the query, providing LLMs with relevant context for generating the output~\cite{zhao2024retrievalaugmentedgenerationaigeneratedcontent}. RAG has shown remarkable effectiveness in scenarios where a corpus of task-specific documents is available. By leveraging this external knowledge, RAG can generate more accurate and informative outputs compared to prompt engineering alone. This is useful in autonomous driving tasks, as it enables autonomous vehicles to access knowledge from vast databases of driving experiences. This not only improves the performance and robustness of autonomous driving systems but also allows for greater adaptability and generalization to novel scenarios. 

Several studies have successfully applied RAG to various autonomous driving tasks. One example is RAG-Driver~\cite{yuan_rag-driver_2024}, which proposes a retrieval-augmented multi-modal large language model (MLLM) for autonomous driving. Given a query comprising a video of the current driving scenario and its corresponding control signal, RAG-Driver searches within a knowledge database for driving experiences that are similar to the current one. The MLLM then processes both the current query and the retrieved in-context learning samples to generate predictions for the current driving task. This approach enables RAG-Driver to leverage past experiences and adapt to new situations effectively. Another interesting application is LaMPilot~\cite{ma_lampilot_2024}, which draws inspiration from the Voyager framework~\cite{wang_voyager_2023}. LaMPilot utilizes LLM-generated function descriptions as database keys. When the system receives human instructions similar to those encountered in the past, it can quickly retrieve the appropriate policy code to perform the desired actions. 

While RAG offers potential, it also has some limitations. One challenge is the need for a high-quality knowledge base that covers the relevant information for the target task. Building and maintaining such a knowledge base can be nontrivial. Additionally, the performance of RAG heavily depends on the quality of the retrieval process. If the retrieved documents are not relevant or informative, the generated outputs may be suboptimal or even misleading.

\subsubsection{Reinforcement Learning}
Reinforcement Learning from Human Feedback (RLHF) addresses the limitations of SFT by incorporating dynamic human feedback into the training process. This approach consists of three crucial stages: initial model preparation through SFT, reward model training where human annotators provide comparative feedback on model outputs, and policy optimization using algorithms like Proximal Policy Optimization (PPO) ~\cite{schulman2017proximalpolicyoptimizationalgorithms} to maximize the predicted reward while maintaining semantic consistency with the initial model. For example, DeepSeek-R1~\cite{guo2025deepseek} utilized Group Relative
Policy Optimization (GRPO)~\cite{shao2024deepseekmathpushinglimitsmathematical} to reward different generations and improve reasoning performance. OpenAI o1~\cite{openai2024gpt4ocard} also employs its own reinforcement learning-based techniques to improve the reasoning process. 

In autonomous driving domain, this post-training method is widely used~\cite{zhou2024humancentricrewardoptimizationreinforcement,LI2025128736,pang2024largelanguagemodelguided,xu2025humancentricautonomousdrivingfastslow,10.1145/3675094.3677588,xu2025telldriveenhancingautonomousdriving}. AdaThinkDrive~\cite{luo2025adathinkdrive} proposes a dual-mode reasoning VLA framework for autonomous driving that distinguishes between ``fast answering” (no CoT) and ``slow thinking” (with CoT), and uses a reinforcement learning reward (Adaptive Think Reward) combined with Group Relative Policy Optimization (GRPO) to learn when to think. AutoDrive‑$R^2$~\cite{yuan2025autodriver2incentivizingreasoningselfreflection} introduces a VLA model with explicit CoT plus self-reflection supervision and applies RL (GRPO) in a physics-grounded reward setting to ensure trajectories are both interpretable and dynamically feasible. ReCogDrive~\cite{li2025recogdrive} presents a ``reinforced cognitive” framework that integrates a vision-language model with a diffusion-based planner, uses a three-stage training pipeline (QA pre-training, imitation learning, RL) to map from language reasoning to continuous driving actions, and improves performance on long-tail driving scenarios. ReasonPlan~\cite{liu2025reasonplanunifiedsceneprediction} proposes a closed-loop autonomous driving method that fine-tunes an MLLM with two complementary modules (a self-supervised next-scene-prediction task and a supervised decision chain-of-thought process) using the large PDR reasoning dataset to improve generalization and interpretability. MTRDrive~\cite{luo2025mtrdrivememorytoolsynergisticreasoning} introduces a VLA-based driving framework that integrates a memory bank of past driving experiences and a tool-calling engine to support CoT reasoning grounded in prior knowledge and detection tools, improving robustness in corner cases and out-of-distribution scenarios. Poutine~\cite{rowe2025poutine} presents a large 3 B-parameter VLM pretrained on vision-language-trajectory tokens, then fine-tuned via reinforcement learning (GRPO) to excel in long-tail autonomous driving scenarios.s. DriveAgent‑R1~\cite{zheng2025driveagent} proposes an autonomous driving agent that conducts active perception and uses a hybrid thinking framework (switching between fast text-only reasoning vs tool-augmented reasoning) trained via a cascaded RL strategy, enhancing interpretability and adaptability. AgentThink~\cite{qian2025agentthink} integrates dynamic, agent-style tool invocation directly into a VLM's CoT, allowing the model to autonomously call external tools to gather precise information and fill knowledge gaps. AutoVLA~\cite{zhou2025autovla} employs adaptive reasoning and reinforcement fine‑tuning to present an end-to-end VLA model that unifies scene reasoning and vehicle action generation within one autoregressive framework, and uses reinforcement learning GRPO to balance reasoning cost and trajectory planning performance.

\subsection{Pipeline}
\subsubsection{Input}
The choice of input modality and representation plays a crucial role in the effectiveness and capabilities of language-guided autonomous driving frameworks. In state-of-the-art language-guided autonomous driving frameworks, there is no standard form of input formats. The choice of input modality varies depending on the focus and capabilities of the employed backbone models.

Some studies focus on the language capabilities of LLMs and provide environment information in the form of descriptive text~\cite{wen_dilu_2024,ma_lampilot_2024}. For instance, DiLu~\cite{wen_dilu_2024} utilizes a scenario descriptor that transcribes the present scenario data into a comprehensive text description. The descriptor follows a standard sentence structure and employs natural language to provide a detailed depiction of the ongoing driving scenario, including static road details and dynamic information about the ego vehicle and surrounding vehicles.

Other studies leverage MLLMs that can integrate multiple data modalities, with sensor data such as images from on-board cameras being the most commonly used~\cite{xu_drivegpt4_2024,tian_drivevlm_2024,pan_vlp_2024,wang_drive_2024}. These approaches aim to exploit the visual understanding capabilities of MLLMs to enhance the perception and decision-making of autonomous driving systems. Some studies have also explored integrating point clouds from LiDAR sensors into MLLMs~\cite{wang_drivemlm_2023,shao_lmdrive_2024}, enabling the models to process and reason about 3D information. EM-VLM4AD~\cite{gopalkrishnan2024multiframelightweightefficient} proposes an efficient multi-frame vision-language architecture for driving-focused visual question answering (VQA), using gated attention fusion to achieve competitive perception and reasoning performance with 10× less compute, demonstrating real-time deployment feasibility in driving environments. HiLM-D~\cite{ding2025hilmdenhancingmllmsmultiscale} enhances current MLLMs with a dual-stream perception design, combining temporal video reasoning with high-resolution fine-grained spatial cues, yielding significant improvements in risk-object localization and explainable decision-making on ROLISP.

In addition to visual and textual inputs, some studies have taken a different path by integrating numeric vector modality, which is frequently used in robotics for representing speed, positions, and distance measurements~\cite{chen_driving_2024,yuan_rag-driver_2024}. For example, ``Driving with LLMs"~\cite{chen_driving_2024} fuses vectorized object-level 2D scene representations, commonly used in autonomous driving, into a pre-trained LLM using adapters. This approach enables the model to directly interpret and reason about comprehensive driving situations based on structured numeric data.

\subsubsection{Output}
Similar to input pipelines, there is no standard output format across recent language-guided autonomous driving frameworks. The choice of output representation varies depending on the specific goals and tasks of each work. The commonly used output modalities are \textit{natural language (NL)}, discrete \textit{actions}, \textit{code} snippets, and \textit{trajectories}, as shown in \cref{tab:sota}.

Many recent studies utilized natural language outputs for various purposes. Some frameworks generate chain-of-thought responses that provide a step-by-step explanation of the reasoning process behind the decisions to improve the accuracy and interpretability of the completion~\cite{wen_dilu_2024,cui2024onboardvisionlanguagemodelspersonalized}. Other studies focus on answering questions posed by human users, enhancing the explainability of the system~\cite{wayve_lingo1_2023,xu_drivegpt4_2024}. OpenDriveVLA~\cite{zhou2025opendrivevlaendtoendautonomousdriving} introduces a unified vision-language-action model built on structured 2D/3D scene representation and hierarchical alignment, supporting both natural-language reasoning and trajectory prediction and achieving strong generalization in open-loop planning tasks.

In \cref{tab:sota},  the term \textit{signal} is used to refer to control signals represented as numerical values. For instance, DriveGPT-4 predicts control signals, including the ego vehicle's speed and turning angle, for the subsequent step. 

The term \textit{action} represents the format of a set of predefined discrete actions that correspond to short-term decisions. As an example, DriveVLM~\cite{tian_drivevlm_2024} defines a set of 17 meta-action categories, including acceleration, deceleration, turning left, changing lanes, etc. These high-level actions provide a more abstract representation of the vehicle's behavior compared to low-level control signals. Mei et al.~\cite{mei_continuously_2024} proposes a dual-process decision-making framework, combining the strong but slow analytic process using a large language model and a fast and empirical process using a smaller model, and finially generate the driving actions. Jiang et al. \cite{jiang2024senna} introduces an integrated framework for VLM's usage in end-to-end autonomous driving, this framework leverages perceptual information gotten from the end-to-end model and provides meta-actions guidance to the end-to-end model.

\begin{table*}[!t]
  \centering
  \caption{Comparison with Real-world Autonomous Driving Deployment Platforms}
  \resizebox{0.8\textwidth}{!}{%
    \begin{tabular}{c|c|c|c|c|c}
      \hline
      \makecell{\textbf{Key}\\\textbf{Features}} &
      \makecell{\textbf{LLM4AD-}\\\textbf{Integrated}} &
      \makecell{\textbf{Holistic}\\\textbf{Testing}} &
      \makecell{\textbf{Open}\\\textbf{Source}} &
      \makecell{\textbf{Plug-Play}\\\textbf{Flexibility}} &
      \makecell{\textbf{Log}\\\textbf{Replay}} \\
      \hline
      Multi-agent Coordination \& CAV~\cite{burnett2020zeusdescriptiontwotimewinner},~\cite{kemsaram2022model}          & \XSolidBrush & \Checkmark &
      \XSolidBrush &  
      \XSolidBrush & 
      \XSolidBrush  \\
      
      Sensing \& Perception~\cite{testouri2024robocarrapidlydeployableopensource},~\cite{walling2017design},~\cite{gu2023end}         & \XSolidBrush & \XSolidBrush & \Checkmark & 
      \XSolidBrush & 
      \Checkmark \\ 
      
      Decision Making \& Planning ~\cite{9743646} & 
      \XSolidBrush &  \Checkmark & 
      \Checkmark & 
      \XSolidBrush &
      \Checkmark \\

      Motion Control \& Extreme Handling~\cite{9304728},~\cite{djeumou2024one},~\cite{10.1115/1.4045320},~\cite{7963824} & \XSolidBrush & \XSolidBrush & \XSolidBrush & \XSolidBrush & \Checkmark\\
      
      Full-stack Autonomy Validation~\cite{karle2023edgar},~\cite{fremont2020formalscenariobasedtestingautonomous},~\cite{wei2013towards},~\cite{guo2024developing}          & \XSolidBrush & \Checkmark & \Checkmark & \XSolidBrush & \Checkmark  \\
      Scalable Embedded Autonomy~\cite{evans2024unifyingf1tenthautonomousracing} & \XSolidBrush & \XSolidBrush & \Checkmark & \Checkmark & \Checkmark \\

      LLM4AD-Integrated Full-stack Autonomy Validation~\cite{zhou2025hierarchical}      & \Checkmark & \Checkmark & \Checkmark & \Checkmark & \Checkmark \\
      \hline
    \end{tabular}%
  }
  \label{tab:platform_comparison}
\end{table*}

Some studies adopt the \textit{code as policy} paradigm~\cite{liang_code_2023}, where the language model generates program code as the policy. For instance, LaMPilot~\cite{ma_lampilot_2024} generates code snippets instead of directly issuing low-level control signals. These code snippets guide the strategic navigation of the ego vehicle based on user instructions.

In \cref{tab:sota}, \textit{trajectory} refers to a sequence of waypoints that define the future path of the vehicle. Language Agent~\cite{mao_language_2024}, for example, instructs LLMs to generate text-based driving trajectories by reasoning on the input data. The trajectories provide a plan for the vehicle to follow, which can be further refined into final control signals such as throttle, brake, and steer with controllers like PID.

A more popular trend of the recent LLM-based agents usually provides multiple categories of outputs. Hwang et al. \cite{hwang2024emma} and Xing et al. \cite{xing2025openemma} introduce comprehensive frameworks for the use of multimodal large language models, outputing perception results, meta-actions and trajctories. And both of them demonstrate effectiveness, generalizability, and robustness across a variety of challenging driving scenarios when generating multiple outputs. Zhang et al. \cite{guo2025vdt} use VLM for scene understanding, meta-actions and potential paths to assist the diffusion-based path generation. Fu et al. \cite{fu2025orion} further enable the backpropagation from the generative planner to VLM while the VLMs output scene description, action reasoning and scene analysis.

\subsection{Evaluation}
Evaluating the performance of LLM4AD frameworks is crucial for assessing their effectiveness, robustness, and safety. However, due to the diverse range of tasks, input modalities, and output formats, there is no standardized evaluation methodology across recent studies. Researchers have employed various datasets, simulation environments, and real-world settings to evaluate their proposed frameworks, as shown in \cref{tab:sota}.

One common approach is to use existing autonomous driving datasets that have been augmented with natural language annotations. For instance, NuScenes-QA~\cite{qian_nuscenes-qa_2024} and NuPrompt~\cite{wu2025languagepromptautonomousdriving} provide perceptual information as text by describing the positions and states of surrounding objects. BDD-X~\cite{kim2018textual} provides reasons for the ego vehicle's actions in natural language descriptions. DriveLM~\cite{sima_drivelm_2024} organizes language annotations from the object level and task level with a graph structure. Additionally, some researchers collect new datasets and design vision-language tasks specifically for driving. For example, DRAMA~\cite{malla_drama_2023} and Rank2Tell~\cite{sachdeva_rank2tell_2024} identify crucial objects and provide corresponding driving suggestions. These datasets enable the evaluation of a framework's ability to generate human-like explanations and reasoning for its driving decisions. However, these datasets do not include closed-loop validation and do not fully capture the interactive nature of autonomous driving.

Simulation environments provide a safe and controlled setting for evaluating the performance of language-guided autonomous driving frameworks. PoPopular simulation platforms, including CARLA~\cite{dosovitskiy_carla_2017} and HighwayEnv~\cite{leurent_environment_2018}, allow researchers to test their frameworks under various driving scenarios and traffic densities.imulation also enables the collection of large-scale datasets for training and evaluation purposes. However, the sim-to-real gap remains a challenge, and the performance of such frameworks in simulation may not always translate directly to real-world settings.

To bridge the gap between simulation and real-world deployment, some studies have taken the bold step of deploying language-guided autonomous driving frameworks in real vehicles. Talk2Drive~\cite{cui_drive_2024} is a pioneering example, being the first of its kind to integrate LLMs on a real autonomous vehicle. This real-world evaluation demonstrates the practicality and potential of language-guided driving. However, real-world testing also poses significant challenges in terms of safety, reliability, and legal considerations.

In addition to task-specific evaluation metrics, such as success rate, collision rate, and traffic rule violations, some studies also assess the quality of the generated natural language outputs. Metrics such as BLEU~\cite{papineni_bleu_2002} and ROUGE~\cite{lin_rouge_2004}, commonly used in natural language generation tasks, can be adapted to evaluate the coherence and relevance of the generated explanations and responses. LINGO-Judge~\cite{marcu_lingoqa_2024} also proposes a more advanced trainable metric that demonstrates a 0.95 Spearman correlation coefficient with human evaluations, which could be used to evaluate the human-likeness of language outputs. Aiming to improve the reasoning capability, Wang et al.~\cite{wang_drivecot_2024} collected a dataset (DriveCoT) for chain-of-thought (CoT) reasoning, including labels for the reasoning process and final decisions.

As language-guided autonomous driving continues to evolve, developing standardized evaluation benchmarks and protocols will be crucial for fair comparison and progress tracking. These benchmarks should cover a wide range of driving scenarios, tasks, and input/output modalities, and include well-defined metrics for assessing performance, safety, and user interaction. Establishing such benchmarks will facilitate collaboration and accelerate advancements in this field. Additionally, as these technologies move towards real-world deployment, it will be essential to establish rigorous safety and performance standards to ensure their reliable and responsible implementation.

\subsection{Deployment Platform}
The rapid advancement of autonomous driving technology needs comprehensive real-world testing and evaluation to validate system safety and robustness, particularly when encountering edge cases related to unpredictable pedestrian and vehicle behaviors. Therefore, deployment platforms that provide real-world testing are important for validating state-of-the-art autonomous driving systems and enabling their safe deployment on public roads. As LLM4AD frameworks show significant potential across the entire autonomous driving pipeline, there is an increasing demand for the entire testing platform that supports comprehensive, full-stack evaluations from perception, reasoning, decision-making, and planning to final vehicle maneuvers.

As summarized in the Tab. \ref{tab:platform_comparison}, different autonomous driving deployment platforms have been developed with varying capabilities and specializations. EDGAR~\cite{karle2023edgar} provides full-stack autonomy validation. Zeus~\cite{burnett2020zeusdescriptiontwotimewinner} focuses exclusively on testing and deployment for multi-agent coordination scenarios. Kemsaram et al.~\cite{kemsaram2022model} proposed deployment platforms specifically for Connected Autonomous Vehicles (CAVs). RoboCar~\cite{testouri2024robocarrapidlydeployableopensource} and IseAuto~\cite{gu2023end} specialize in multimodal sensor fusion and dataset collection capabilities. Kessler et al.~\cite{9743646} design a real vehicle platform focusing on motion planning. Goel et al.~\cite{9304728} and Djeumou et al.~\cite{djeumou2024one} utilized modified Lexus LC500 and Toyota Supra platforms to validate autonomous drifting. For extreme vehicle handling scenarios, Goh et al.~\cite{10.1115/1.4045320} and Laurense et al.~\cite{7963824} introduce specialized real-world vehicle platforms. Comprehensive autonomous driving functionality testing has been addressed through platforms developed by Wei et al.~\cite{wei2013towards}, Fremont et al.~\cite{fremont2020formalscenariobasedtestingautonomous}, and Guo et al.~\cite{guo2024developing}. F1TENTH~\cite{evans2024unifyingf1tenthautonomousracing} offers scalable embedded autonomous testing and evaluation capabilities.

As shown in the Tab. \ref{tab:platform_comparison}, most existing real-world deployment and testing frameworks are inherently limited in their capacity to comprehensively evaluate LLM4AD frameworks due to their specialization in specific areas. To address this gap, Zhou et al.~\cite{zhou2025hierarchical} developed a hierarchical real-world deployment platform specifically designed for the evaluation of LLM4AD frameworks.


}

\newcommand{\pilot}{LaMPilot\xspace}
\newcommand{\pilotb}{LaMPilot-Bench\xspace}

\section{Benchmarking LLM4AD in Simulation}
\label{sec:benchmark}

This section introduces two comprehensive benchmarks for evaluating the instruction-following capabilities and performance of LLM-based agents in autonomous driving: One is for LLM-based agent planning evaluation (the \pilotb dataset on the HighwayEnv simulator and the CARLA Leaderboard 1.0~\cite{dosovitskiy_carla_2017}); Another is a multi-view visual question answering benchmark (NuPlanQA based on Nuplan dataset~\cite{caesar2022nuplanclosedloopmlbasedplanning,park2025nuplanqalargescaledatasetbenchmark}).

\subsection{\pilotb in HighwayEnv}
\subsubsection{Dataset}
The \pilotb dataset consists of 4.9K semi-human-annotated traffic scenes, with a subset of 500 samples split as the test set. Each data sample includes a high-level task description (instruction), an initial state for initializing the simulator, and goal state criteria aligned with the instruction. The dataset covers diverse driving scenarios, as shown in \cref{tab:stat}. \first{Our ``semi-human-annotated" protocol involved a two-stage process. First, for each scenario, \pilotb includes various situations, such as different ego vehicle initial positions and states, specific tasks (turning left/right or going straight), the number of other vehicles, and their positions and states. The driving model parameters for other vehicles are randomly initialized, and each scenario is assigned a random seed. Second, these scenarios were presented to a team of human annotators. These annotators were tasked with writing a clear, natural language instruction (e.g., ``Move to the middle lane and drive at 30 mph") that corresponded to the programmatic goal. To ensure the quality and consistency of these annotations, a quality control pass is then implemented. A separate team of senior evaluators reviewed the annotations, checking for ambiguity, grammatical errors, and, most importantly, a clear alignment between the natural language instruction and the scenario's programmatic goal. Any instruction that was found to be ambiguous or did not uniquely map to the intended goal was discarded and sent back for re-annotation.}

The \pilotb simulator is built upon HighwayEnv~\cite{leurent_environment_2018}, a widely used platform for autonomous driving research and tactical decision-making, which offers various driving models and simulates realistic multi-vehicle interactions. the HighwayEnv is extended with interfaces suitable for LLM-based agents and implement custom intersections to diversify the driving scenarios.
\subsubsection{Evaluator}
The \pilotb evaluator incorporates metrics to assess the safety and efficiency of the agent driving policies.
Time-to-collision (TTC) measures the vehicle's ability to maintain safe distances and avoid collisions. In a scenario with $n+1$ vehicles (ego vehicle labeled as 0), the TTC with vehicle $i$, denoted as $\tau_i$, is calculated using the ego vehicle's longitudinal velocity $\mathbf{v}_0$ and position $\mathbf{p}_0$, and vehicle $i$'s longitudinal velocity $\mathbf{v}_i$ and position $\mathbf{p}_i$ ($1 \leq i \leq n$):
\begin{equation}
\label{eq:tau}
\tau_i = -\frac{(\mathbf{p}_0 - \mathbf{p}_i) \cdot (\mathbf{v}_0-\mathbf{v}_i)}{\|\mathbf{v}_0-\mathbf{v}_i\|^2}
\end{equation}
The minimum positive TTC value $\tau_{\min}$ among all $n$ vehicles over all time steps $t$ (up to task completion time $T$) represents the nearest potential collision:
\begin{equation}
\label{eq:tau_min}
\tau_{\min} = \min_{1\leq i\leq n, 1\leq t\leq T} \tau^{t}_{i}
\end{equation}
TTC scores are empirically assigned based on a 2-second safety margin, with values above 2 seconds considered safe and given a score of 100:
\begin{equation}
\operatorname{TTC} =
\begin{cases}
100 & \text{if $\tau_{\min} > 2$} \\
\max(0, 100 - \frac{1}{\tau_{\min}}) & \text{if $0 < \tau_{\min} \leq 2$} \\
0 & \text{if $\tau_{\min} \leq 0$}
\end{cases}
\end{equation}

\first{The expression $100 - 1/\tau_{min}$ is a hyperbolic penalty function. This form is chosen because the penalty $1/\tau_{min}$ grows non-linearly toward infinity as $\tau_{min}$ approaches the unsafe boundary of 0. This ensures that critically dangerous, near-collision events are penalized far more severely than marginally unsafe ones. This ``inverse barrier" logic is a common method for enforcing safety constraints in robotics and is a core concept in Control Barrier Functions (CBFs)~\cite{ames2016control}.}


Speed variance (SV) is another safety metric. The ego vehicle's speed standard deviation $\sigma_{0}$ is calculated as:
\begin{equation}
\label{eq:sv}
\sigma_{0} = \sqrt{\frac{\sum_{t=1}^T{(\|\mathbf{v}^t_0\|-\mu_0)^2}}{T}},
\end{equation}
where $\mu_0$ is the mean speed:
\begin{equation}
\mu_{0} = \frac{\sum_{t=1}^T{\|\mathbf{v}^t_0\|}}{T}.
\end{equation}
The SV score is defined as:
\begin{equation}
\operatorname{SV} = \max(0, 100 \cdot (1 - {\sigma_{0}}/{\sigma_{\operatorname{safe}}})),
\end{equation}
where $\sigma_{\operatorname{safe}}$ is the maximum safe speed deviation, determined empirically.
The time efficiency (TE) score evaluates the policy's ability to complete the task within a predefined time limit $T_{\operatorname{limit}}$:
\begin{equation}
\operatorname{TE} = \max(0, 100 \cdot (1 - {T}/{T_{\operatorname{limit}}}))
\end{equation}
A TE score closer to 100 indicates a more efficient policy.
A task is considered successfully completed when the agent achieves the specified objectives while maintaining safety (i.e., avoiding collisions) and efficiency (i.e., finishing within the allotted time). The final score aggregates all individual metrics, weighted according to their importance:
\begin{equation}
\label{eq:score}
\operatorname{Score} = W_{\operatorname{TTC}} \cdot \operatorname{TTC} + W_{\operatorname{SV}} \cdot \operatorname{SV} + W_{\operatorname{TE}} \cdot \operatorname{TE}
\end{equation}

\subsubsection{Baselines and Results}
 Several baselines are estabilished to evaluate the performance of LLM-based agents across the benchmarks, as presented in \cref{tab:lampilot-result} for \pilotb.

\tcbset{colback=white}
\colorlet{titleblue}{blue!80!black}
\colorlet{titlered}{red!80!black}
\colorlet{titlegreen}{green!80!black}
\colorlet{darkgreen}{green!50!black}
\newcommand{\inc}[1]{{\color{darkgreen}#1}}
\newcommand{\dec}[1]{{\color{titlered}#1}}

\begin{table}[!t]
\centering
\caption{Main statistics of the \pilotb dataset.}
\resizebox{0.8\linewidth}{!}{
\begin{tabular}{l|l}
\hline
Statistic & Value\\
\hline
Total scenarios & 4,900 (100\%)\\
\hline
Distance-related & 1,200 (24.5\%) \\
Speed-related & 1,200 (24.5\%) \\
Pulling over & 200 (4.1\%) \\
Routing & 1,500 (30.2\%) \\
Lane changing & 400 (8.2\%)\\
Overtaking & 400 (8.2\%)\\
\hline
Highway & 3,400 (69.4\%)\\
Intersection & 1,500 (30.6\%)\\
\hline
Average instruction length & 7.6 words\\
Maximum instruction length & 14 words\\
Minimum instruction length & 2 words\\
\hline
\end{tabular}
}
\label{tab:stat}
\end{table}

\begin{table*}[!t]
    \centering
    \caption{Comparison of baselines on \pilotb under different learning settings.
    }
    \small
    \resizebox{0.95\linewidth}{!}{
    \begin{tabular}{l|c|llllll}
        \hline
        Method & Learning & Collision ($\downarrow$) & Completion $(\uparrow)$ & TTC Score $(\uparrow)$ & SV Score $(\uparrow)$ & TE Score $(\uparrow)$ & Driving Score $(\uparrow)$\\
        \hline
        IDM & \multirow{3}{3em}{N/A} & 0.0\% & 20.4\% & 92.8 & 80.9 & 71.0 & 17.3\\
        MOBIL & & 0.0\% & 15.3\% & 82.8 & 85.9 & 46.7 & 11.7\\
        Human Driver & & 0.0\% & 98.0\% & 93.4 & 86.3 & 81.3 & 84.6 \\
        \hline
        Llama 2 & \multirow{5}{3em}{0-Shot} 
        & 0.0\% & 20.4\% & 92.8 & 81.4 & 68.9 & 17.3\\
        PaLM 2 & 
        & 3.1\% & 35.7\% & 78.6 & 76.5 & 78.8 & 12.8 \\
        ChatGPT & 
        & 1.0\% & 40.8\% & 83.9 & 75.5 & 74.0 & 27.8\\
        GPT-4 & 
        & 4.1\% & 72.4\% & 61.3 & 70.5 & 74.0 & 28.5 \\
        GPT-4 Turbo & 
        & 3.1\% & 74.5\% & 73.2 & 70.1 & 73.9 & 39.1\\ 
        \hline
        Llama 2 & \multirow{5}{3em}{3-Shot} 
        & 1.0\% (\dec{+1.0}) & 63.3\% (\inc{+42.9}) & 68.3 (\dec{-24.5}) & 71.4 (\dec{-10.0}) & 73.8 (\inc{+4.9}) & 39.6 (\inc{+22.3}) \\
        PaLM 2 &  
        & 3.1\% (0.0) & 71.4\% (\inc{+35.7}) & 73.4 (\dec{-5.2}) & 69.7 (\dec{-6.8}) & 72.0 (\dec{-6.8}) & 36.6 (\inc{+23.8}) \\
        ChatGPT &  
        & 4.1\% (\dec{+3.1}) & 83.7\% (\inc{+42.9}) & 70.9 (\dec{-13.0}) & 73.7 (\dec{-1.8}) & 77.7 (\inc{+3.7}) & 41.2 (\inc{+13.4}) \\
        GPT-4 &  
        & 1.0\% (\inc{-3.1}) & 84.7\% (\inc{+12.3}) & 69.4 (\inc{+8.1}) & 72.0 (\inc{+1.5}) & 76.7 (\inc{+2.7}) & 55.9 (\inc{+27.4})\\
        GPT-4 Turbo &  
        & 1.0\% (\inc{-2.1}) & 80.6\% (\inc{+6.1}) & 69.1 (\dec{-4.1}) & 71.5 (\inc{+1.4}) & 74.9 (\inc{+1.0}) & 52.6 (\inc{+13.5})\\ 
        \hline
        Llama 2 & \multirow{5}{4em}{Human Feedback} 
        & 0.9\% (\inc{-0.1}) & 32.7\% (\dec{-30.6}) & 83.9 (\inc{+15.6}) & 74.6 (\inc{+3.2}) & 74.6 (\inc{+0.8}) & 21.4 (\dec{-18.2}) \\
        PaLM 2 &  
        & 0.0\% (\inc{-3.1}) & 45.5\% (\dec{-25.9}) & 81.1 (\inc{+7.7}) & 74.4 (\inc{+4.7}) & 77.6 (\inc{+5.6}) & 35.6 (\dec{-1.0})\\
        ChatGPT &  
        & 0.9\% (\inc{-3.2}) & 80.9\% (\dec{-2.8}) & 70.2 (\dec{-0.7}) & 71.7 (\dec{-2.0}) & 77.5 (\dec{-0.2}) & 54.2 (\inc{+13.0})\\
        GPT-4 &  
        & 0.9\% (\inc{-0.1}) & 92.7\% (\inc{+8.0}) & 67.6 (\dec{-1.8}) & 72.2 (\inc{+0.2}) & 76.8 (\inc{+0.1}) & 64.0 (\inc{+8.1})\\
        GPT-4 Turbo &  
        & 0.9\% (\inc{-0.1}) & 87.3\% (\inc{+6.7}) & 74.1 (\inc{+5.0}) & 73.4 (\inc{+1.9}) & 75.5 (\inc{+0.6}) & 60.5 (\inc{+7.9})\\ 
        \hline
    \end{tabular}
    }
    \label{tab:lampilot-result}
\end{table*}

\begin{table*}[!ht]
    \centering
    \small
    \caption{Driving performance of expert drivers on CARLA leaderboard 1.0 testing routes. $\uparrow$: Higher values are better.}
    \resizebox{0.75\linewidth}{!}{
    \begin{tabular}{l|l|c|ccc}
        \toprule
        Method & Learning & Navigation & DS $(\uparrow)$ & RC $(\uparrow)$ & IP $(\uparrow)$ \\
        \midrule
        Roach Expert~\cite{zhang_end--end_2021} & Reinforcement Learning & Waypoint & 63.4 & 100.0 & 0.63 \\
        TCP Expert~\cite{wu_trajectory-guided_2022} & Reinforcement Learning & Waypoint &  68.1 & 95.8 & 0.71 \\
        \midrule
        LLM Planner & 0-Shot CoT & Instruction & 0.00 & 48.7 & 0.00\\
        LLM Planner & 3-Shot CoT & Instruction & 17.3 & 65.3 & 0.31\\
        LLM Planner & Human-Feedback CoT & Instruction & 65.4 & 95.1 & 0.68
        \\
        \bottomrule
    \end{tabular}
    }

    \label{tab:carla-result}
\end{table*}

\paragraph{Heuristic baselines} Rule-based methods such as the Intelligent Driver Model (IDM)\cite{treiber_congested_2000} and Minimizing Overall Braking Induced by Lane Changes (MOBIL)\cite{kesting_general_2007}. Both methods achieve a zero-collision rate in \pilotb, as shown in \cref{tab:lampilot-result}. However, it is important to note that these methods operate independently of the provided instructions. Without considering human instructions, IDM and MOBIL achieve success rates of 20.4\% and 15.3\%, respectively. These results provide a reference point for assessing the effectiveness of LLM-based agents in following human instructions.
\paragraph{Zero-shot and few-shot baselines} LLMs are prompted with task instructions and API documentation, with or without in-context examples. For the few-shot setting, human programmers proficient in Python create in-context examples. They are given API documentation and allowed to write and test their code on the training routes. The same examples are used across all test routes. In the zero-shot setting, GPT models~\cite{openai_gpt-4_2023} and PaLM 2~\cite{anil_palm_2023}, given only the API Docs, driving context, and instructions, achieve an obvious performance advantage compared to rule-based methods in terms of task completion, as shown in \cref{tab:lampilot-result}. When provided with training examples that include exemplar code (three-shot setting), all the evaluated LLMs exhibit notable improvements in completion rates. For instance, Llama 2~\cite{touvron_llama_2023}'s completion rate increases from 20.4\% in the zero-shot setting to 63.3\% in the three-shot setting.
\paragraph{Human feedback baselines} LLMs are integrated with a human-in-the-loop learning approach, leveraging human feedback to refine generated policies. After executing a generated policy code ($P$), the human evaluators provide natural language feedback ($F$), which is fed back into the LLM along with $P$. This feedback loop enables continuous learning. If the feedback is positive (i.e., the human is satisfied with the execution), the code ($P$) is committed to the database for future retrieval and reuse. Otherwise, the feedback serves as guidance for iterative improvement. The integration of GPT models with human-in-the-loop learning further enhances their performance in driving tasks. Notably, GPT-4 achieves the highest driving score of $64.0$, as shown in \cref{tab:lampilot-result}. This improvement highlights the great potential of LLMs in following instructions in the driving context when combined with human feedback.
\paragraph{Human performance baseline} A licensed human driver controlling the vehicle in the simulation using keyboard inputs. This baseline provides a reference for human-level performance.

\subsection{CARLA Leaderboard 1.0 Benchmark}
\subsubsection{Dataset}
In addition to \pilotb in HighwayEnv, the CARLA Leaderboard 1.0 benchmark~\cite{carla_team_autonomous_2020,dosovitskiy_carla_2017} is used to evaluate LLM-based agents in a more realistic and complex driving environment. The Leaderboard consists of 76 routes (50 for training, 26 for testing), totaling over 170 km across six towns. It also includes predefined scenarios covering ten types of challenging traffic situations, such as control loss and traffic negotiation.
To test the instruction-following capabilities of LLM-powered systems, the Leaderboard setup is modified accordingly to provide natural language navigation instructions based on the agent's current position instead of GPS coordinates. To eliminate perception influence on our results, a language generator is used to craft the driving context information that leverages CARLA's privileged information.
Executing an LMP in CARLA involves using Python's \pyth{exec} function, which takes the LMP code as an input string and two dictionaries defining the execution scope: (i) \pyth{apis}, containing all driving APIs the code may call, and (ii) \pyth{local_vars}, initially empty but later holding a generator variable named {\ttfamily policy} once executed.
The Leaderboard evaluates agent performance using three key metrics:
\begin{itemize}
\item Route Completion (RC): measures the percentage of the route distance completed by the agent.
\item Infraction Penalty (IP): tracks various infractions (e.g., collisions, running red lights) committed by the agent, aggregated as a geometric series starting from an ideal base score of 1.0 that is reduced with each infraction.
\item Driving Score (DS): the product of RC and IP, serving as the principal evaluation metric.
\end{itemize}


\subsubsection{Baselines and Results}

\cref{tab:carla-result} is for the CARLA Leaderboard 1.0. \first{It is important to note that the scores reported in Tab.~\ref{tab:carla-result} represent the mean performance of each agent aggregated across all 26 official CARLA Leaderboard 1.0 testing routes with different random seeds.}

\begin{figure*}[!t]
    \centering
    \includegraphics[width=1.0\textwidth]{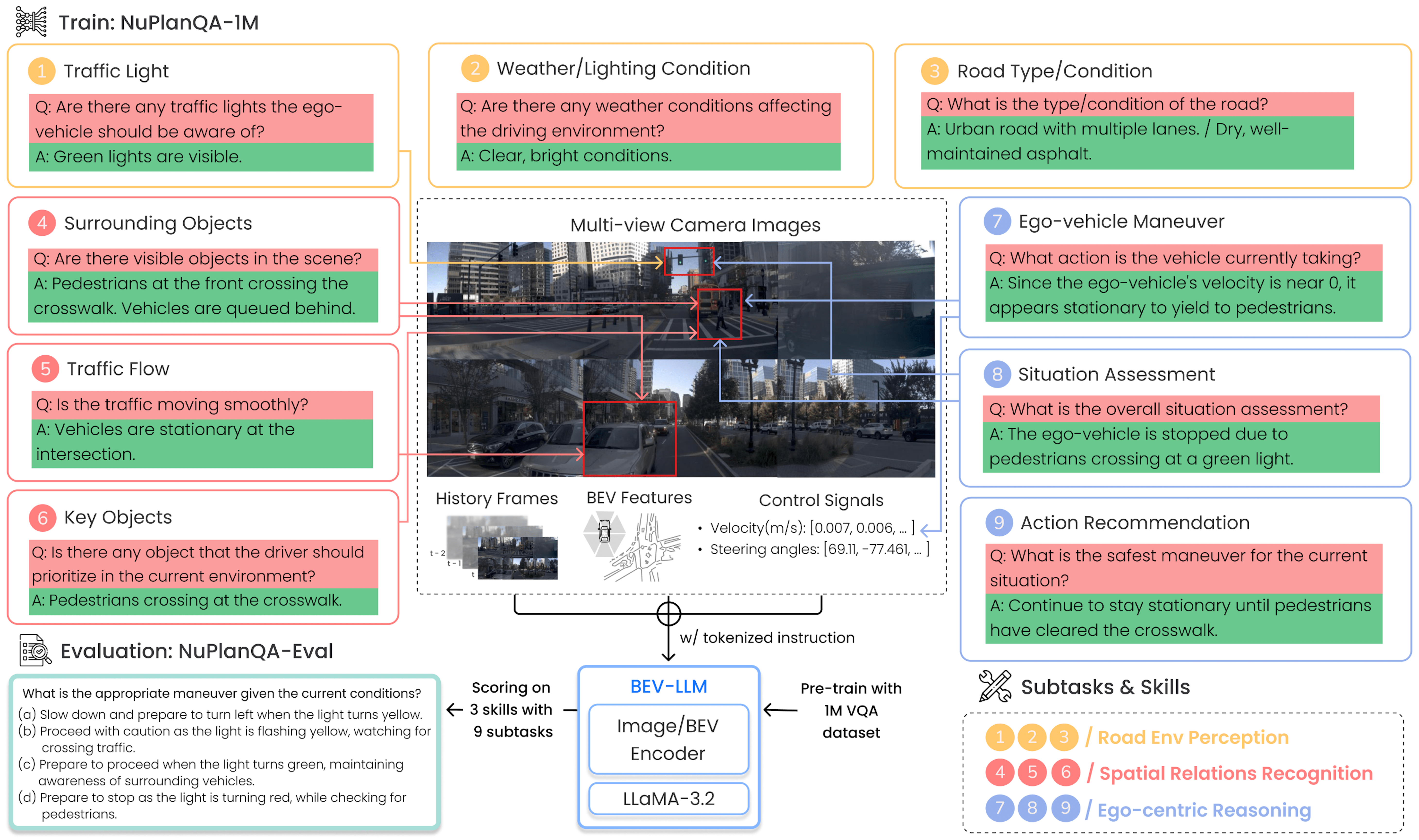}
\caption{An overview of NuPlanQA. NuPlanQA comprises nine subtasks across three skill areas to support the context-aware analysis of traffic scenes. The proposed baseline, \textit{\textbf{BEV-LLM}}, is trained on \textit{\textbf{NuPlanQA-1M}} using historical frames, BEV features from multi-view images, and control signals as inputs. Finally, MLLMs are evaluated using \textit{\textbf{NuPlanQA-Eval}}, a multiple-choice QA benchmark for driving scene understanding.}
    \label{fig:nuplanqa}
\end{figure*}

The human-feedback baseline with reinforcement learning experts that have access to privileged information, such as Roach~\cite{zhang_end--end_2021} and TCP~\cite{wu_trajectory-guided_2022} in the CARLA leaderboard~\cite{ma_learning_2024} are compared, as shown in \cref{tab:carla-result}. The comparison aims to provide a reference to previous works. Based on these results, the follow conclusions are observed:
\begin{itemize}
\item Without few-shot examples, the out-of-the-box LLMs struggle with the precise reasoning required for complex closed-loop driving in CARLA.
\item The three-shot baseline also falls short, resulting in significantly lower scores according to CARLA's standard metrics.
\item However, the LLM with 50 code snippets learned from human feedback performs on par with the Roach Expert reinforcement learning baseline.
\end{itemize}
The experiments demonstrate that off-the-shelf LLMs can generate code policies for driving tasks based on human instructions, outperforming rule-based methods in task completion. Few-shot learning with exemplar code further improves the performance of LLM-based agents, and the integration of human feedback enables iterative refinement of generated policies, leading to the highest driving scores among the evaluated methods.

\subsection{NuPlanQA Benchmark}
\label{sec:nuplanqa}

\subsubsection{Dataset}
\first{
While \pilotb and CARLA Leaderboard evaluate LLM-based agents in closed-loop simulation, they are limited to low-level control policies and rule-based task execution. To further assess the high-level reasoning and perception capabilities of MLLMs in realistic driving environments, the recently proposed NuPlanQA benchmark is incorporated~\cite{park2025nuplanqalargescaledatasetbenchmark}. 

NuPlanQA provides a large-scale, multi-view VQA framework tailored for driving-scene understanding. It is built upon the \textit{nuPlan}~\cite{caesar2022nuplanclosedloopmlbasedplanning} dataset, offering 1M VQA pairs (NuPlanQA-1M) for training and 8K multiple-choice VQA pairs (NuPlanQA-Eval) for evaluation. Each VQA sample corresponds to multi-view frames with annotated control signals and scene descriptions. The dataset supports context-aware reasoning over both spatial and temporal dimensions of traffic scenes, bridging the gap between low-level control evaluation (as in LaMPilot-Bench) and high-level situational reasoning.

\begin{table*}[!ht]
  \centering
  \caption{\textbf{Performance of MLLMs on NuPlanQA-Eval.} The metric used is accuracy (\%)—calculated as the number of correct responses over the total number of questions. The best-performing model in each task is \textbf{bolded}, while the second-best is \underline{underlined}. LLaVA-OV: LLaVA-OneVision, LLaVA-NV: LLaVA-Next-Video.}
  \small 
\begin{tabularx}{\linewidth}{p{2.7cm} *{13}{>{\centering\arraybackslash}X}}
    \toprule
    \multicolumn{1}{c}{} & \multicolumn{4}{c}{\textbf{Road Env. Perception}} & \multicolumn{4}{c}{\textbf{Spatial Relations Recog.}} & \multicolumn{4}{c}{\textbf{Ego-centric Reasoning}} & \multicolumn{1}{c}{}\\ 
    \cmidrule(lr){2-5} \cmidrule(lr){6-9} \cmidrule(lr){10-13}
    \textbf{\makecell{Method\\ \\}} & \textit{\makecell{Trfc.\\Light}} & \textit{\makecell{Wea\\-ther}} & \textit{\makecell{Road\\Type}} & \textit{\makecell{Avg.}}
    & \textit{\makecell{Sur.\\Obj.}} & \textit{\makecell{Trfc.\\Flow}} & \textit{\makecell{Key\\Obj.}} & \textit{\makecell{Avg.}}
    & \textit{\makecell{Ego\\Ctrl.}} & \textit{\makecell{Situ.\\Asse.}} & \textit{\makecell{Act.\\Rec.}} & \textit{\makecell{Avg.}} & \textbf{\makecell{Total\\ \\}}\\
    \midrule
    \multicolumn{14}{c}{\textbf{Multi-frame as Input}} \\
    \midrule
    GPT-4o {\scriptsize{\cite{openai2024gpt4ocard}}}& 68.5 & 91.7 & 95.0 & 85.1 & 79.6 & 76.5 & 67.7 & 74.6 & 86.8 & 85.1 & 82.4 & 84.8 & 81.5 \\
    Gemini-1.5-Pro {\scriptsize{\cite{gemini_team_gemini_2024}}}& 64.5 & 93.5 & 95.8 & 84.6 & 69.1 & 78.6 & 73.2 & 73.6 & 73.8 & 73.8 & 80.4 & 76.0 & 78.1 \\
    \midrule
    VideoLLaMA2{\tiny7B}{\scriptsize{\cite{damonlpsg2024videollama2}}}& \underline{54.2} & 59.4 & \underline{93.8} & 69.1 & 70.7 & \textbf{78.6} & 66.4 & 71.9 & 72.3 & \underline{82.7} & 81.4 & \underline{78.8} & 73.3 \\
    Qwen2.5-VL{\tiny7B} {\scriptsize{\cite{bai2025qwen25vltechnicalreport}}}& 51.7 & 71.9 & 69.3 & 64.3 & 49.2 & 23.0 & 30.9 & 34.4 & 30.4 & 39.6 & 10.1 & 26.7 & 41.8 \\
    LLaVA-OV{\tiny7B} {\scriptsize{\cite{li2024llavaonevisioneasyvisualtask}}}& 53.2 & \textbf{89.4} & \textbf{96.4} & \underline{79.7} & \underline{77.9} & \underline{77.6} & \textbf{73.2} & \textbf{74.3} & \underline{75.4} & 76.2 & \underline{82.9} & 77.5 & \underline{77.8} \\
    LLaVA-NV{\tiny7B} {\scriptsize{\cite{zhang2024llavanext-video}}}& 44.3 & 46.5 & 72.4 & 54.4 & 57.5 & 52.0 & 44.1 & 51.2 & 57.6 & 65.8 & 71.4 & 64.9 & 56.8 \\
    InternVL-1.5{\tiny20B} {\scriptsize{\cite{internvl1.5}}}& 40.9 & 70.5 & 75.5 & 62.3 & 48.6 & 50.0 & 44.5 & 47.7 & 57.1 & 55.0 & 60.8 & 57.6 & 55.9 \\
    LLaVA-NV{\tiny32B} {\scriptsize{\cite{zhang2024llavanext-video}}}& 47.8 & \underline{74.7} & \underline{93.8} & 72.1 & 69.1 & 74.0 & 63.6 & 68.9 & 72.8 & 69.3 & 76.9 & 73.0 & 71.3 \\
    \midrule
    \rowcolor{gray!25} BEV-LLM (Ours)& \textbf{61.1} & \textbf{89.4} & 89.6 & \textbf{80.0} & \textbf{78.5} & 75.5 & \underline{68.2} & \underline{74.1} & \textbf{79.1} & \textbf{83.2} & \textbf{83.4} & \textbf{81.9} & \textbf{78.7} \\
    \midrule
    \multicolumn{14}{c}{\textbf{Single-frame as Input}} \\
    \midrule
    GPT-4o {\scriptsize{\cite{openai2024gpt4ocard}}} & 66.5 & 92.2 & 94.4 & 84.4 & 81.2 & 77.0 & 68.5 & 75.6 & 86.4 & 86.6 & 81.4 & 84.8 & 81.6 \\
    Gemini-1.5-Pro {\scriptsize{\cite{gemini_team_gemini_2024}}}& 64.5 & 93.1 & 95.8 & 84.5 & 71.1 & 79.6 & 72.7 & 74.5 & 79.6 & 76.7 & 80.9 & 79.1 & 79.4 \\
    \midrule
    VideoLLaMA2{\tiny7B}{\scriptsize{\cite{damonlpsg2024videollama2}}}& 50.2 & 64.1 & \underline{93.8} & 69.4 & 71.3 & \textbf{79.6} & 64.1 & 71.7 & 69.6 & \textbf{84.2} & \underline{81.4} & \underline{78.4} & 73.2 \\
    Qwen2.5-VL{\tiny7B} {\scriptsize{\cite{bai2025qwen25vltechnicalreport}}}& 51.2 & 70.5 & 62.5 & 61.4 & 47.0 & 16.8 & 23.2 & 29.0 & 27.2 & 40.6 & 12.1 & 26.6 & 39.0 \\
    LLaVA-OV{\tiny7B} {\scriptsize{\cite{li2024llavaonevisioneasyvisualtask}}}& \underline{52.2} & \textbf{93.5} & \textbf{96.4} & \textbf{80.7} & \underline{72.9} & 74.0 & \textbf{68.6} & \underline{71.8} & \underline{72.3} & 72.3 & 79.9 & 74.8 & \underline{75.8} \\
    LLaVA-NV{\tiny7B} {\scriptsize{\cite{zhang2024llavanext-video}}}& 48.8 & 41.5 & 63.5 & 51.3 & 51.9 & 50.5 & 41.8 & 48.1 & 56.5 & 60.4 & 69.8 & 62.2 & 53.9 \\
    InternVL-1.5{\tiny20B} {\scriptsize{\cite{internvl1.5}}}& 39.4 & 68.7 & 72.4 & 60.2 & 54.7 & 48.5 & 41.4 & 48.2 & 62.3 & 58.9 & 58.8 & 60.0 & 56.1 \\
    LLaVA-NV{\tiny32B} {\scriptsize{\cite{zhang2024llavanext-video}}} & 46.8 & 73.3 & 91.7 & 70.6 & 70.7 & 71.9 & 58.2 & 66.9 & 71.2 & 68.8 & \textbf{82.9} & 74.3 & 70.6  \\
    \midrule
    \rowcolor{gray!25} BEV-LLM (Ours)& \textbf{58.1} & \underline{87.1} & 84.9 & \underline{76.7} & \textbf{75.1} & \underline{75.0} & \underline{66.8} & \textbf{72.3} & \textbf{76.0} & \underline{81.2} & \underline{81.4} & \textbf{79.5} & \textbf{76.2} \\
    \bottomrule
  \end{tabularx}
    \vspace{3mm}
  \label{tab:nuplanqa}
  \vspace{-5mm}
\end{table*}

\begin{figure}[!b]
    \centering
    \includegraphics[width=\linewidth]{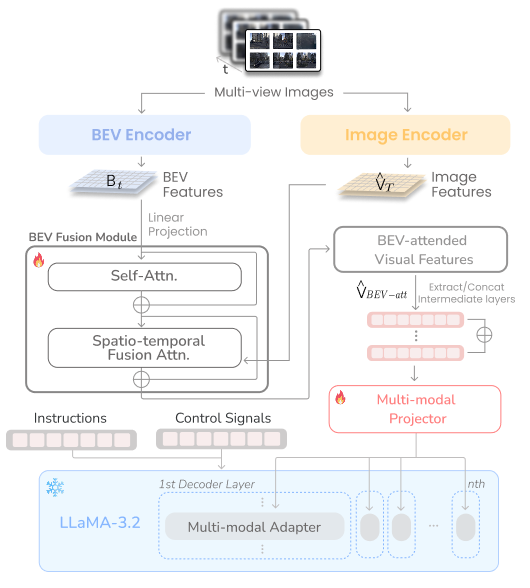}
\caption{Architecture of BEV-LLM. BEV features are integrated into LLaMA via the BEV-Fusion module.}
    \label{fig:nuplan_model}
\end{figure}

As shown in Fig.~\ref{fig:nuplanqa}, NuPlanQA decomposes the driving-scene understanding problem into nine subtasks spanning three hierarchical skill domains:
\begin{itemize}
    \item \textit{Road Environment Perception:} traffic light recognition, weather/lighting condition, and road type/condition;
    \item \textit{Spatial Relations Recognition:} identification of surrounding objects, traffic flow, and key objects critical to the ego vehicle;
    \item \textit{Ego-Centric Reasoning:} inference of ego-vehicle maneuver, situation assessment, and action recommendation.
\end{itemize}
This categorization enables fine-grained evaluation of MLLMs across low-level perception, spatial understanding, and high-level reasoning, facilitating diagnostic analysis of LLMs' strengths and weaknesses.

Each question in NuPlanQA-Eval is associated with four answer choices and evaluated by accuracy. To ensure annotation quality, all QA pairs undergo both GPT-4o-assisted generation with chain-of-thought prompting and human verification for semantic consistency and contextual correctness. The evaluation set maintains an approximately uniform distribution across all nine subtasks to prevent bias toward particular skill domains.

\subsubsection{NuplanQA Baseline and Results}


To establish a reference model, BEV-LLM is proposed as a multi-view, Bird's-Eye-View(BEV)-enhanced MLLM that integrates spatially aligned BEV features from multiple cameras into the vision encoder of LLaMA-3.2-Vision~\cite{grattafiori2024llama3herdmodels}. As shown in Fig. \ref{fig:nuplan_model}, the LLaMA-3.2-Vision encoder first extracts per-frame visual features \( \hat{V}_t\) from time-sequential multi-view images. The features across all timesteps are denoted as $V_t$.

To capture temporal dependencies, BEV features ($B_t$) are constructed by attending to historical BEV embeddings ($B_{t-1}$) using the temporal self-attention (TSA) from BEVFormer \cite{li_bevformer_2022}:

\vspace{-5pt}

\begin{equation}
    B_{\text{t}} = \text{TSA}(Q, \{Q, B_{t-1}\}) + B_{t-1}
\end{equation}

\vspace{3pt}


\noindent where $Q$ represents BEV queries. This TSA mechanism encodes motion-aware scene representations by incorporating past information.

These refined BEV features are then fused with the multi-view visual features ($\hat{V}_T$) using spatio-temporal fusion attention (SFA). In SFA, visual features act as queries, while the temporally-enriched BEV embeddings ($B_{\text{proj}}$, projected from $B_t$) serve as keys and values:


\vspace{-5pt}

\begin{equation}
    \hat{V}_{\text{BEV-att}} = \text{SFA}(\hat{V}_T, B_{\text{proj}}) + \hat{V}_T
\end{equation}

This cross-attention allows the visual features to integrate the spatial-temporal BEV context, implicitly transferring temporal awareness.

The features from multiple layers of $\hat{V}_{\text{BEV-att}}$ (every fourth layer and the final hidden state \cite{grattafiori2024llama3herdmodels})  are extracted and concatenated to create a robust multi-scale representation. A multi-modal projector then aligns this representation with the language model's text embedding space.

Finally, the projected vision-BEV embeddings are fed into the LLaMA decoder alongside tokenized text instructions and vehicle data (e.g., velocity, steering). The decoder's cross-attention layers function as a multi-modal adapter \cite{grattafiori2024llama3herdmodels, alayrac2022flamingovisuallanguagemodel}, enabling BEV-LLM to integrate motion, depth, and control signals for a comprehensive multi-modal understanding.

As shown in Tab.~\ref{tab:nuplanqa}, on NuPlanQA-Eval, BEV-LLM achieves an average accuracy of 78.7\%, outperforming other open-source models in six of nine subtasks, particularly in ego-centric reasoning and traffic light detection. These results highlight the effectiveness of BEV integration in enabling models to reason about motion and ego-vehicle context.}

\subsection{Summary}
The \pilotb dataset and the CARLA Leaderboard 1.0 benchmark provide a comprehensive evaluation framework for assessing the instruction-following capabilities and performance of LLM-based agents in autonomous driving. The experiments highlight the potential of LLMs in enhancing the decision-making capabilities of autonomous vehicles when combined with human feedback and domain-specific APIs. As LLMs continue to evolve and integrate with more modalities, their significant contribution is envisioned to the development of safe, efficient, and user-friendly autonomous vehicles. Future research should focus on addressing the challenges of real-time decision-making, reducing collision rates, and incorporating multi-modal inputs to further enhance the performance and robustness of LLM-based autonomous driving systems.

However, the notable collision rates in both benchmark indicate the need for further research to fully capture the complexities and safety requirements of real-world driving scenarios. Most failures stem from the trade-off made for the LLM low-level planning frequency, as illustrated in the example below:

\begin{python}
# Command: Alright, you can start driving.
# Context Info: 
# There is a vehicle at a distance of 6.8 m. 
# It is moving at a speed of 0.0 m/s. 
# It is at an angle of 1.9 degrees. 
# The angle between the headings is 0.0 degrees.
# My current speed is 0.0 m/s.
def start_driving_with_front_vehicle_check():
    (is_front_vehicle, 
     distance_to_front_vehicle, 
     speed_of_front_vehicle)
     = check_front_vehicle()
    if (is_front_vehicle 
        and distance_to_front_vehicle < 5 
        and speed_of_front_vehicle == 0.0):
        yield stop()
    else:
        speed_limit = check_speed_limit()
        yield proceed(speed_limit)
\end{python}

In this case, the vehicle could collide with a front vehicle if it starts to move. This issue arises because the LLM reasoning is based on the current context information and does not account for all potential future actions of surrounding agents. Due to the latency of LLM reasoning, it is not feasible to employ LLMs to reason at every single timestamp, which is a challenge to be addressed in future research.

\first{These findings, particularly the non-zero collision rates observed even in our best-performing models like the GPT-4 with human feedback (0.9\% in LaMPilot-Bench), highlight a critical safety gap. While these models outperform rule-based methods in task completion, any persistent collision rate, no matter how small, is unacceptable for deployment in real-world, safety-critical autonomous driving.}

\first{This result strongly implies that the current generation of LLMs, when used in a zero-shot or few-shot manner, are not sufficiently reliable for direct, real-time motion control. The root cause, as discussed, often lies in the inherent latency and the non-real-time reasoning frequency of these large models. They struggle to account for the dynamic evolution of a driving scene between reasoning cycles, leading to failures in time-sensitive ``corner cases."}

\first{The safety implications of this are clear: LLM-based agents should not be granted direct, low-level control over a vehicle's safety-critical functions. This limitation is a primary motivator for the architectural designs explored in our real-world experiments in Section~\ref{sec:real-vehicle} and~\ref{sec:onboard}. In those frameworks, the LLM is intentionally relegated to high-level, non-instantaneous tasks such as interpreting human preferences, personalizing driving style, or adjusting controller parameters while a separate, robust, low-latency control system maintains ultimate authority over real-time vehicle safety.}

\first{Additionally, NuPlanQA complements LaMPilot-Bench and CARLA Leaderboard by extending evaluation beyond closed-loop policy control to encompass perceptual understanding and reasoning in complex, real-world driving scenes. Together, these benchmarks form a holistic assessment pipeline for LLM4AD systems, spanning high-level perception and reasoning (NuPlanQA) to mid- and low-level decision execution (LaMPilot-Bench and CARLA).}

\section{Real-Vehicle Experiments: On-Cloud LLM for Personalized Decision-Making} 
\label{sec:real-vehicle}

\begin{figure*}[!t]
    \centering
    \includegraphics[width=\textwidth]{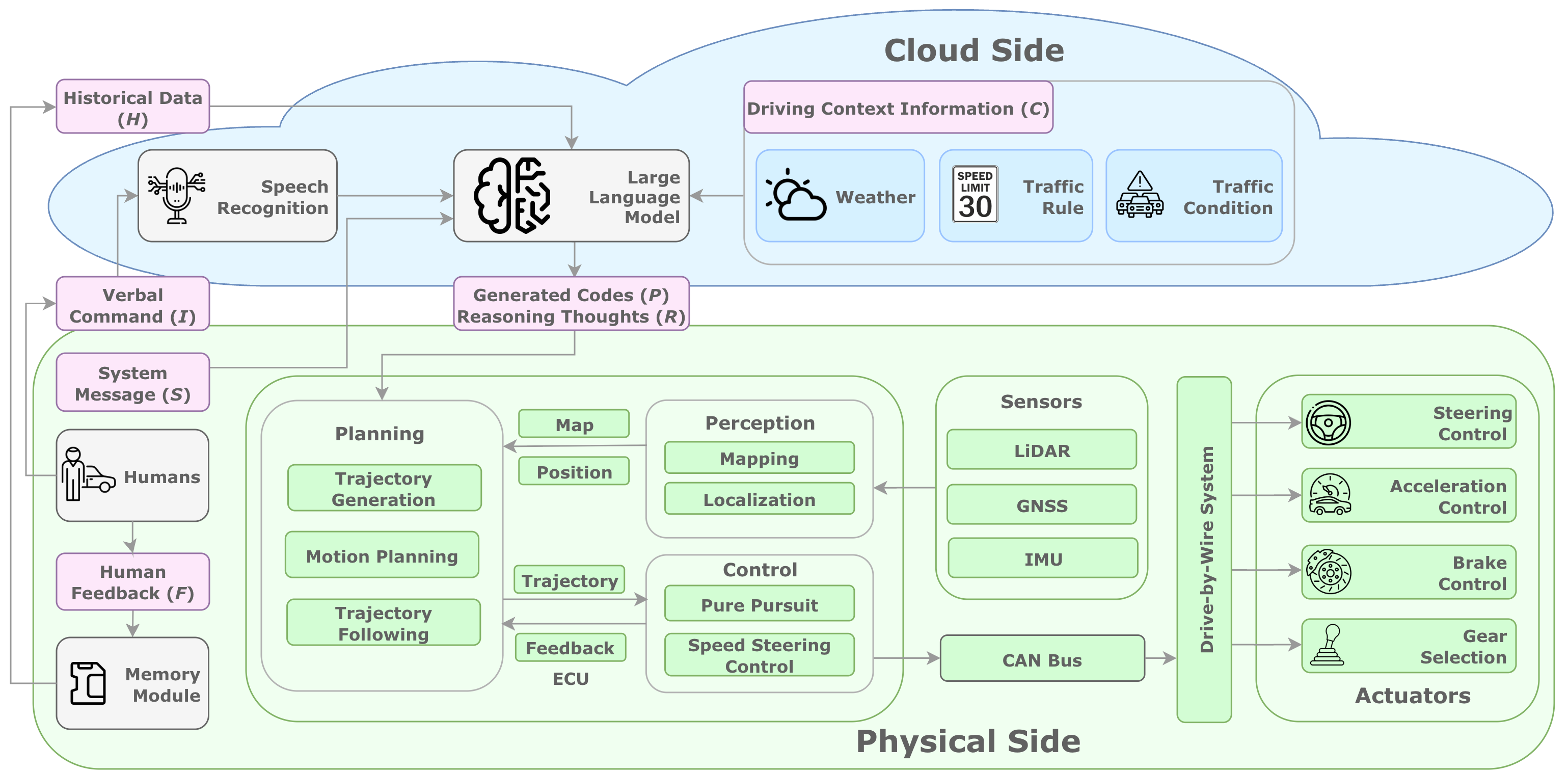}
    \caption{The proposed framework for personalized decision-making with an on-cloud large language model in real-vehicle experiments. On the cloud side, human verbal commands are transcribed and combined with real-time environmental context—such as weather, traffic rules, and road conditions—to serve as inputs for a large language model. This model also accesses historical interaction data to generate executable code policies and reasoning thoughts tailored to the user. These outputs are transmitted to the physical vehicle, where an electronic control unit executes them within the autonomous stack to adjust trajectory generation and motion control parameters. The process concludes with a feedback loop, where human evaluations of the driving maneuvers are stored in a memory module to continuously refine the system's personalization capabilities.}
    \label{fig:framework}
\end{figure*}

To further assess the effectiveness of the LLM4AD system and validate its applicability in real-world scenarios, we integrate LLMs into practical autonomous driving systems, introducing a framework called Talk2Drive. Similar to the concept in \cref{sec:concept}, this system has outstanding command understanding and reasoning performance. It also can also provide a personalized driving experience by leveraging records of previous interactions. The demo for the real-vehicle experiment is publicly available\footnote{Real-vehicle experiment demo video with on-cloud LLM: \url{https://youtu.be/4BWsfPaq1Ro}}.

\subsection{Talk2Drive Framework}

\begin{figure}[t!]
    \centering
    \includegraphics[width=0.8\linewidth]{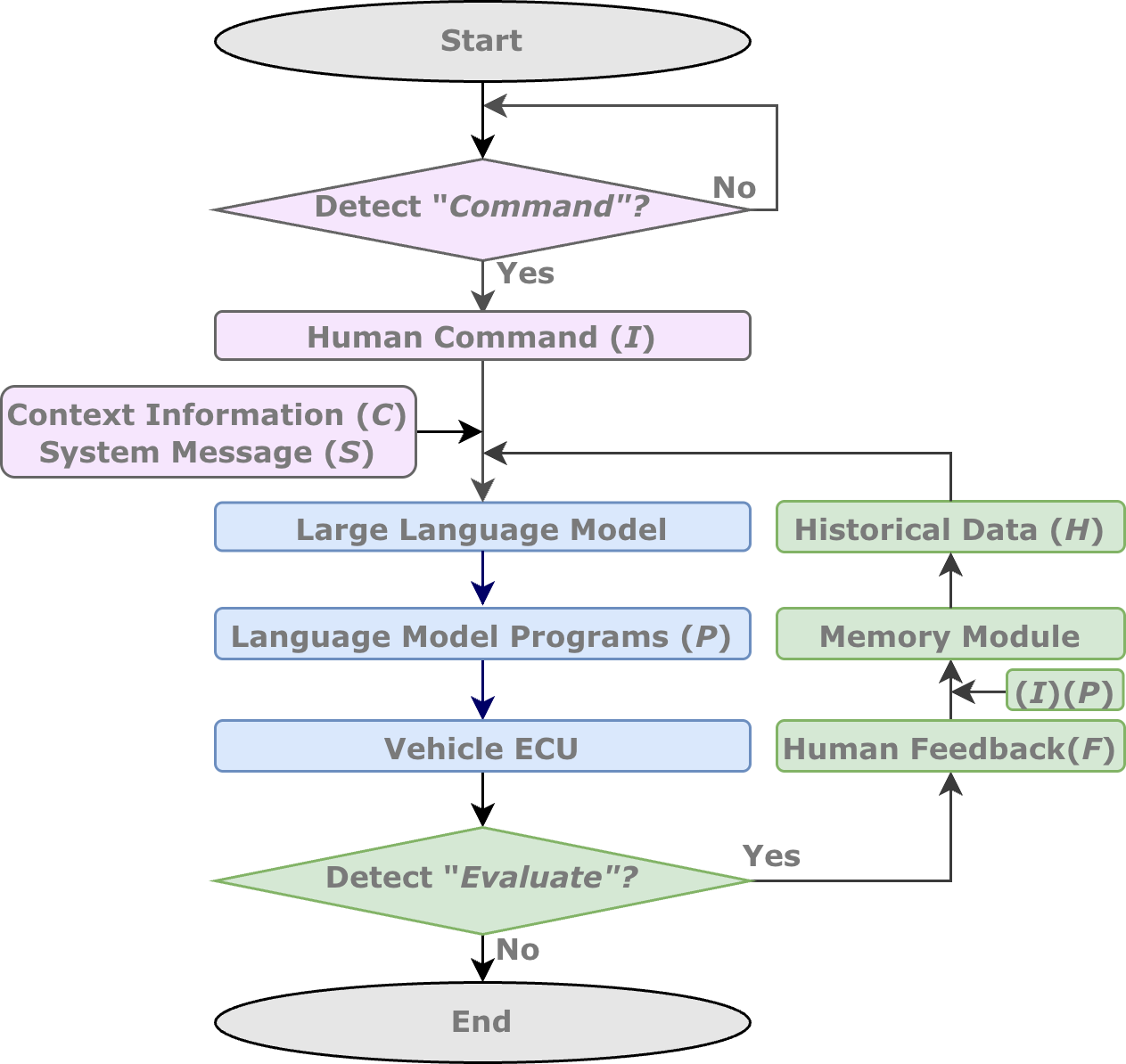}
    \caption{The flowchart of Talk2Drive. After the speech recognition module detects the keyword ``command,'' the inputs ($I,C,S,H$) are sent to the LLM. Then, the LLM generates corresponding LMPs to be executed by the ECU. If the speech recognition module detects the keyword ``evaluate,'' the system receives human feedback ($F$), and both $F$ and its corresponding $I$ and $P$ are updated in the memory module.}
    \label{fig: flowchart}
\end{figure}

This section proposes Talk2Drive (see Fig.~\ref{fig:framework}), an innovative approach to leveraging LLMs to enhance command interpretation and enable personalized decision-making in autonomous vehicles. It integrates cloud-based LLMs to enable personalized understanding and translation of human commands into executable control sequences with real-time vehicle dynamic inputs. This section begins with a problem statement and then articulates the distinct roles of each cloud-side and vehicle-side operation. The flowchart of the Talk2Drive system is shown in Fig.~\ref{fig: flowchart}.

\subsubsection{Problem Statement}In Talk2Drive, followed by the concept proposed in Sec. \ref{sec:concept}. The model aims to translate verbal commands into executable control sequences for the vehicle. The cloud-based LLM acts as a translating function $f:I \rightarrow P$ that generates corresponding LMPs as the policy ($P$) for maneuvers.

For personalization, the human feedback on execution $F(I,P)$ is needed to evaluate if the generated policy addresses the preferences of the driver or the passengers. The extra memory module (see~\ref{sec:mm}), which includes instructions, LMPs, and their evaluations, allows the LLM to learn user preferences. Therefore, there are two stages in the Talk2Drive workflow:
\begin{equation}
    \begin{aligned}
        \text{Execution}:\quad & P \leftarrow f(I,S,C,H); \\
        \text{Evaluation}:\quad & H \leftarrow \left[I,P,F(I,P)\right].
    \end{aligned}
\end{equation}

\subsubsection{Command Translation and Contextual Data Integration}  The initial step in the Talk2Drive framework involves directly receiving arbitrary verbal commands from humans. Utilizing cutting-edge voice recognition technology, specifically the open-source API Whisper~\cite{radford2023robust}, these verbal commands are accurately captured and then translated into textual instructions ($I$). An instance for $I$ is:

\begin{align}
&I \to \label{eq:prompt_command}
\text{Could you drive more conservatively?}
\end{align}

Simultaneously, LLMs access additional cloud-based real-time environment data, including weather updates, traffic conditions, and local traffic rules. In this product, LLMs can be informed by the weather information through Openweather API~\cite{openweather}, the map information (such as road type and speed limits) through OpenStreetMap API~\cite{OpenStreetMap}, and traffic information through TomTom API~\cite{tomtom}. We use a predefined structured language generator for both contextual data ($C$) and system messages ($S$), and this system will supply the appropriate values (the \textcolor{red}{red} contents) to the generator based on the context information, as shown below:
\begin{align}
&C \to \label{eq:prompt} \\
&\left\{\begin{array}{l}
\text{A vehicle in front of you is running at \textcolor{red}{38.0} km/h.}\\
\text{Your current speed is \textcolor{red}{40.0} km/h.} \\
\text{The speed limit is \textcolor{red}{60.0} km/h.} \\
\text{The weather is \textcolor{red}{sunny}.} \\
\vdots
\end{array}\right. \nonumber
\end{align}

\subsubsection{Processing and Reasoning with LLMs}


The LLMs are prompted using in-context learning coupled with chain-of-thought prompting. Specifically, the chain of thought's prompt in Talk2Drive consists of triples: $\{query, thought, output\}$. An example can be found in Eq. \ref{eq:syste_mess}. 

Once a command is translated into texts, it is uploaded to LLMs hosted on the cloud. In experiments,  GPT-4~\cite{openai_gpt-4_2023} is utilized and access it through ChatGPT API. The LLMs engage in a reasoning process to interpret the instruction ($I$), the contextual data ($C$), the system messages ($S$), and the history data ($H$). A simplified example of $S$ is presented below:


\begin{align}
&S \to \label{eq:syste_mess} \\
&\left\{\begin{array}{l}
\text{You are an autonomous vehicle with Adaptive cruise}\\
\text{control (ACC) and Lane Keeping Assist (LKA) always}\\
\text{enabled.} \\
\text{You are using Pure Pursuit Controller to do the waypoint} \\
\text{following.} \\
\vdots\\
\text{Here are some examples of how you need to react.}\\
\text{Query: You drive too aggressively.}\\
\text{Thought: The drivers think I drive too fast which looks}\\
\text{aggressive and the drivers do not ask me to change lanes,}\\
\text{so I need to slow down my speed.}\\
\text{Action: ...}\\
\vdots
\end{array}\right. \nonumber
\end{align}
Additionally, the LLMs also access the memory module, a repository of historical interactions ($H$), to consider the human's past behaviors and preferences. More details about the memory module are in Sec. \ref{sec:mm}. 


\subsubsection{Actionable Code Generation} 
Same as in \cref{sec:concept}, the LLM is utilized to generate corresponding LMPs ($P$) based on this interpretation. These LMPs include complex driving behaviors and parameters that need to be adjusted in the vehicle’s high-level controllers. Specifically, the LMPs adjust control parameters like the look-ahead distance and look-ahead ratio to optimize pure pursuit~\cite{pure_pursuit} performance in Talk2Drive. Additionally, LMPs also modify the target velocity of the vehicle to meet humans' commands. These LMPs take the form of ROS topic commands, directing the autonomous driving system based on Autoware \cite{autoware} to modify its trajectory following configuration.

One simple example code is shown as follows:
\begin{flalign}
     P \to && 
  \end{flalign}
{\small
\begin{lstlisting}{bash}
  $ timeout 1s rostopic pub /vehicle/engage 
  std_msgs/Bool "data: true"
  $ rostopic pub /autoware_config_msgs
  /ConfigWaypointFollower 
  "{\"param_flag\": 1, \"velocity\": 40, 
  \"lookahead_distance\": 12, 
  \"lookahead_ratio\": 2.0}"
  ...
\end{lstlisting}
}

In the meantime, LLMs also generate thoughts $R$ to explain why they make a particular decision. According to ~\cite{wei_chain--thought_2022}, even though the thoughts $R$ are not executed in thissystems, they help LLMs make more reasonable and feasible decisions. An example of the thought $R$ is shown below:

\vspace{-5mm}
\begin{align}
&R \to \label{eq:thought} \\
&\left\{\begin{array}{l}
\text{The driver wants me to drive more conservatively but}\\
\text{does not have any specific lane preference. Therefore, I} \\
\text{should reduce my speed and maintain my current lane.} \\
\end{array}\right. \nonumber
\end{align}

The generated LMPs ($P$) are then sent back from the cloud to the vehicle's electronic control unit (ECU), where they are executed. Additionally,  two kinds of safety checks are set for the generated LMPs. The first check involves verifying the format of the generated code. If the code does not conform to a valid format, this system will not execute or respond to the code. The second safety check is parameter verification, which assesses the feasibility and safety of the parameters within the context of the given policy, preventing the execution of potentially risky policies. For example, if the generated code instructs our agent to move to the left lane while the ego agent is already in the leftmost lane, this system will identify this abnormality and prevent the execution.




\label{sec:exp:setup}

\subsubsection{Memory Module and Personalization}
\label{sec:mm}

 As described in \cref{sec:concept}, the memory module is used to store the historical interactions ($H$) between humans and vehicles to provide a personalized driving experience. One important difference to note about this specific system, compared to the general framework, is that within this system, human feedbacks $F$ are archived within the historical memory, and the history memory $H$ includes three key linked components: the past commands ($I$) issued by the human, the corresponding LMP ($P$) generated by the LLMs in response, and the humans' subjective evaluations or feedback ($F$) on the system's performance for each command. Each interaction between the human and the vehicle is recorded and saved into a memory module in a text format within the ECU. This historical data in the memory module is updated after every trip. The example contents in the memory module are shown in Eq. \ref{eq:mm}:

\vspace{-5mm}
\begin{align}
&H \to \label{eq:mm} \\
&\left\{\begin{array}{l}
\text{Apart from the requirements I provided before. Here I}\\
\text{will provide you with the history dialogues between the}\\
\text{driver and the vehicle. You will need to learn what}\\
\text{drivers' wants and needs are.}\\
\text{For example, if the history driver's command is ``You}\\
\text{drive too conservatively.''}\\
\text{The history output action is : ...} \\
\text{After the trip, the driver's feedback is ``A little bit too}\\
\text{fast for me.''}\\
\text{Then the next time your action should adjust according}\\
\text{to the driver's feedback, which is ...}\\
\vdots\\
\text{The history command, action, and driver's feedback are:} \\
\text{Command: I'm on my way to the urgent care.}\\
\text{Action: ...}\\
\text{Evaluation: A little bit too fast.}\\
\vdots
\end{array}\right. \nonumber
\end{align}

Given the adaptive nature of LLMs, if users respond differently to similar commands, the LLMs will prioritize the most recent response as a reference point for their current decision-making process. When a command from the users is issued, the LLMs access the memory module and take stored information ($H$) as part of the input prompts for the decision-making process. Additionally, each user has their own profile in the memory module, ensuring that this framework can deliver personalized driving strategies for everyone.


\begin{figure}[!t]
    \centering
    \includegraphics[width=0.9\linewidth]{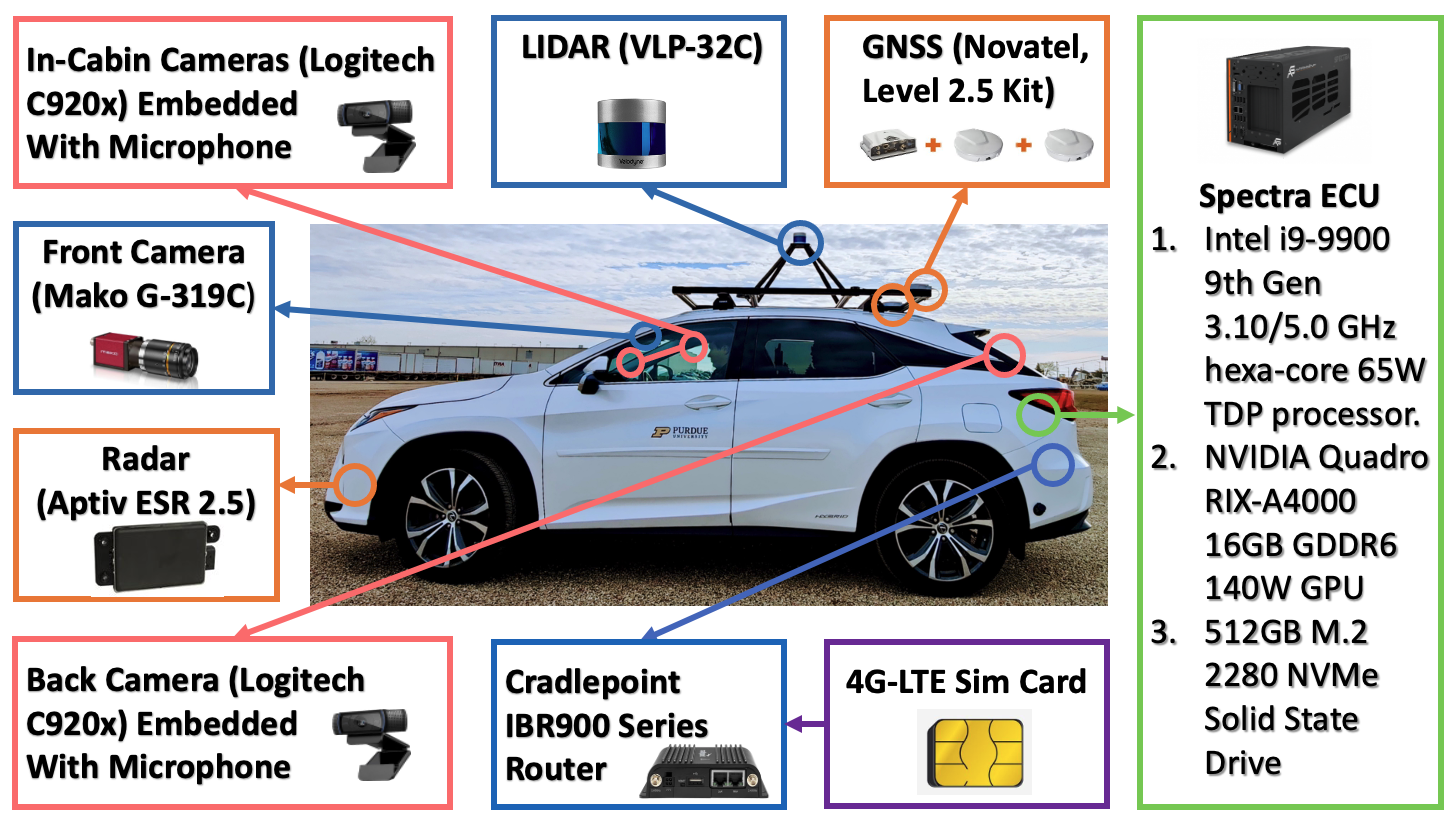}
    \caption{Setup of the real-world autonomous vehicle in the experiment.}
    \label{fig:Vehicle_Setup}
\end{figure}



\subsection{Autonomous Driving System}
\label{subsec:testing_platform}
As shown in Fig. \ref{fig:Vehicle_Setup}, an autonomous vehicle is used to conduct real-world experiments, which is a drive-by-wire enabled 2019 Lexus RX450h with perception sensors, localization module, and communication module. We deploy the open-source autonomous driving software Auoware.AI~\cite{autoware} with ROS Melodic in Ubuntu 18.04. We employ 3D-NDT~\cite{3Dndt} for mapping and localization, and pure pursuit~\cite{pure_pursuit} is utilized for trajectory following. Network communication is enabled by the Cradlepoint IBR900 Series Router, which is equipped with an AT\&T sim card with a 4G-LTE cellular connection.


\begin{figure*}[!t]
    \centering
    \includegraphics[width=\textwidth]{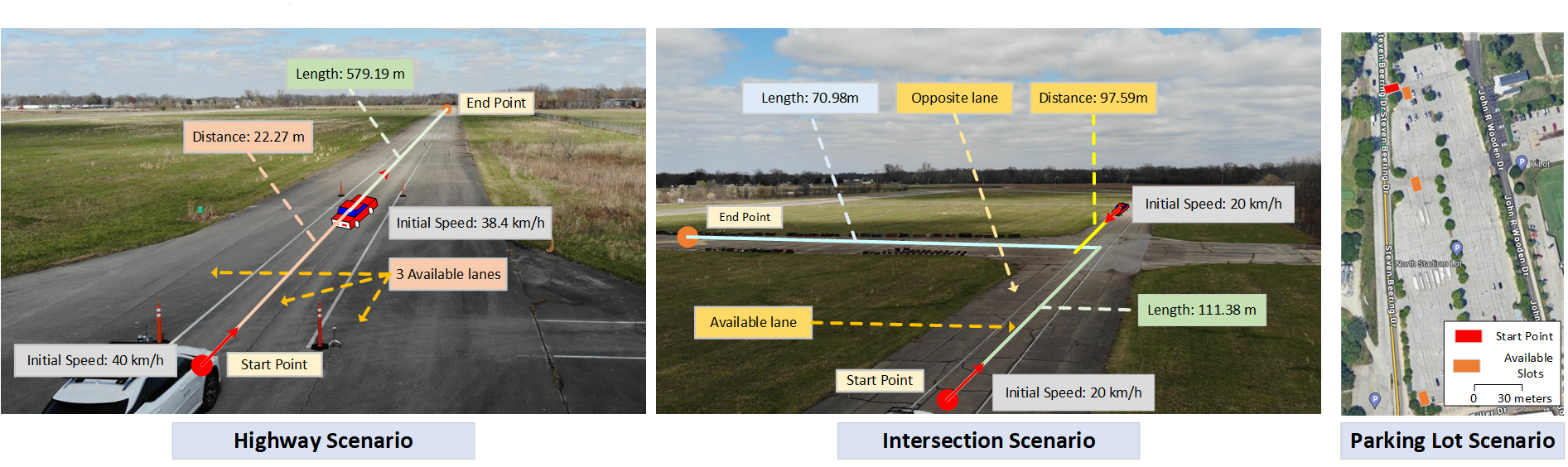}
    \caption{Experiment scenarios: car following on a three-lane highway, unprotected left turn at an intersection, and event parking at a parking lot.}
    \vspace{-3mm}
    \label{fig:mapstat}
\end{figure*}

\subsection{Experiment Setup}
\subsubsection{Scenarios Setup and Participants}

\begin{figure}[!t]
\centering
\includegraphics[width=\linewidth]{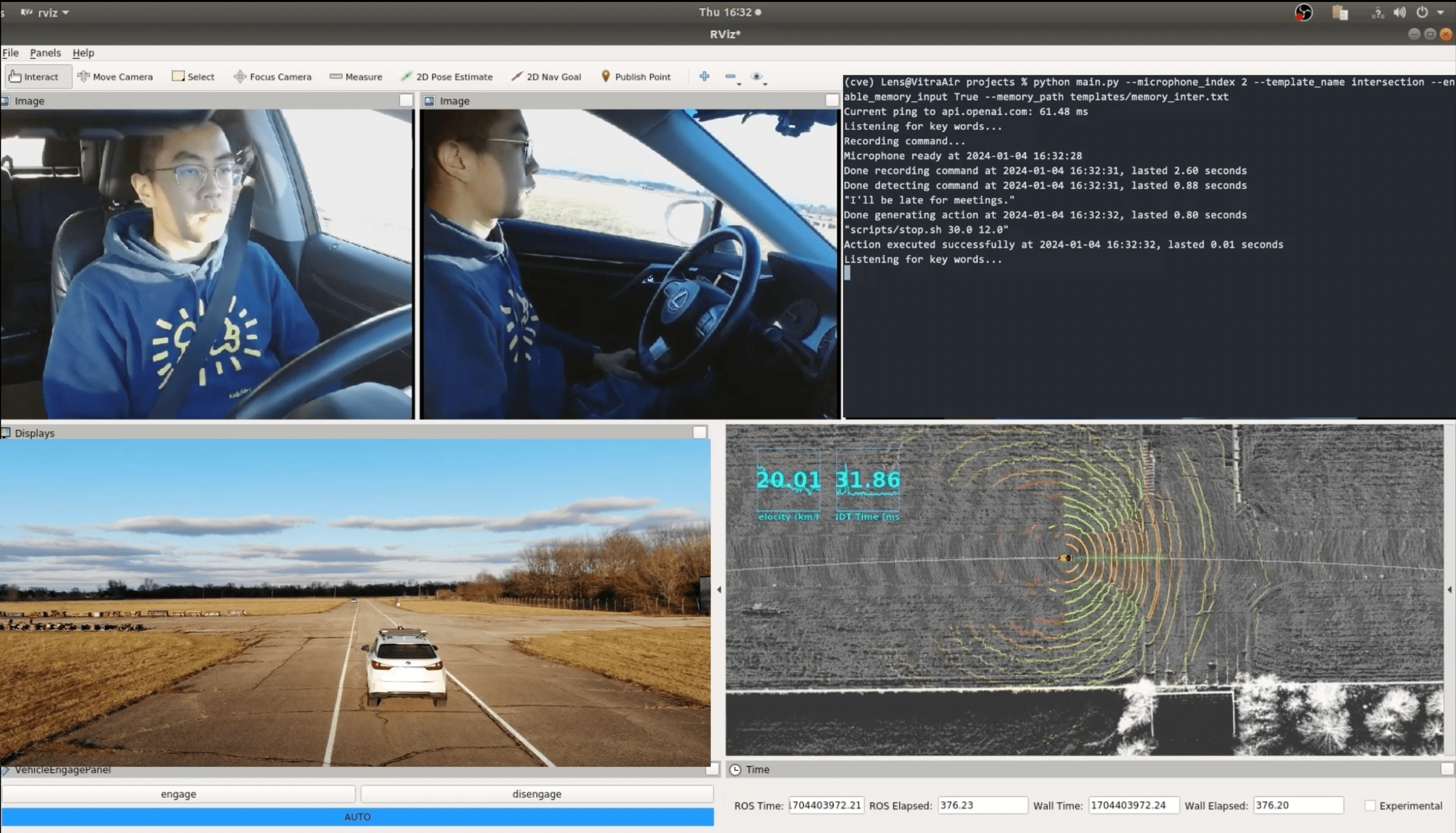}
\caption{The experiment visualization: In the upper left corner is the in-cabin view, while the lower left corner displays the exterior view. The upper right corner shows the console, and the lower right corner presents the lidar map.}
\label{fig:exp_vis}
\end{figure}

Our experimental trials include three different scenarios: highway, intersection, and parking lot. \footnote{The experiments conducted in this study satisfy all local traffic guidelines and guarantee the safety of the participants. A human always sits in the driver's seat of the autonomous vehicle to monitor its status and get ready to take over.} The field experiments for the highway and intersection scenarios are conducted at a proving ground in Columbus, IN, USA, to validate the efficacy of the proposed Talk2Drive framework. The highway scenario involves a three-lane highway, while the intersection contains a two-way junction. Additionally, the parking lot scenario is evaluated at the North Stadium Parking Lot in West Lafayette, IN, USA. The overview visualization and statistics of the test tracks are shown in Fig. \ref{fig:mapstat}, while the visualization for the experiments is shown in Fig. \ref{fig:exp_vis}. \first{In this study, a diverse range of drivers are recruited, including seven individuals of different genders (61.4\% male, 28.6\% female), ages (mean=26.71, std=4.11), and driving experiences(mean=6.79, std=5.08).} All participants have valid driving licenses.

\subsubsection{Input Instructions}
In the field of linguistics, giving instructions can be categorized as making a request, which falls under the division of directives, one of the five speech act types according to the general speech act classification theory~\cite{yule2022study}. More specifically, the scale of directness in requests can be characterized by three distinct strategies~\cite{shoshana1989cross}:

\textbf{Direct Strategies:} In this scenario, the human explicitly states the desired action, such as ``Increase the speed of the vehicle.'' This is usually in the form of imperative sentences.

\textbf{Conventionally Indirect Strategies:} This scenario involves phrasing requests in a manner that is socially and culturally acknowledged as polite and/or standard. An example would be ``Could you please speed up a little bit?''

\textbf{Non-Conventionally Indirect Strategies:} Requests in this scenario are more implicit and rely on contextual understanding. Within this category, hints can be further divided into strong and mild hints. In the context of requesting speed changes in an autonomous vehicle, strong hints may include explicit comments on the current speed, such as ``You are driving too aggressively.'' Conversely, mild hints might be more oblique references to time or urgency, for example, ``I hope we will not be late for the meeting.''

Given that conventionally indirect strategies primarily modify the politeness level of command of direct strategies, this section instead divides the non-conventionally indirect requests into two categories based on the strength of the hints. We define levels of directness in Tab. \ref{tab:directlevel} as the way to classify varying degrees of implicitness in the commands. To gather data, the test drivers are requested to generate commands based on their normal speaking preferences and subsequently categorize their commands into defined levels. Examples of the commands generated are also presented in Tab. \ref{tab:directlevel}.

\begin{table*}[!t]
\centering
\caption{Level of Command Directness and Examples}
\begin{tabular}{c|c|c}
    \toprule
    Command Level & Linguistic Category & Example Commands \\
    \midrule
    \multirow{3}{*}{Level \uppercase\expandafter{\romannumeral1}} & \multirow{3}{*}{Direct and conventionally indirect commands} & Drive as fast as you can. \\
    & & Can you drive faster? \\
    & & You should drive slower. \\
    \midrule
    \multirow{3}{*}{Level \uppercase\expandafter{\romannumeral2}} & \multirow{3}{*}{Non-conventionally indirect commands with strong hints} & You are driving too conservatively.\\
    & & You are driving a little bit slow now. \\
    & & I do not think we should drive this slow. \\
    \midrule
    \multirow{3}{*}{Level \uppercase\expandafter{\romannumeral3}} & \multirow{3}{*}{Non-conventionally indirect commands with mild hints} & I am going to be late for my work. \\
    & & I feel a bit motion-sick right now. \\
    & & I'm really in a hurry now. \\
    \bottomrule
\end{tabular}
\label{tab:directlevel}
\end{table*}
\begin{table*}
\caption{Driving Performance Validation}
\label{tab:drivingperformance}
\centering
\resizebox{2\columnwidth}{!}{%
\begin{tabular}{c|c|c|cc|cc|c|c}
\toprule
\multirow{3}{*}{\begin{tabular}[c]{@{}c@{}}Driving \\ Secnario\end{tabular}} & \multirow{3}{*}{\begin{tabular}[c]{@{}c@{}}Expected Driving \\ Behavior\end{tabular}} & \multirow{3}{*}{\begin{tabular}[c]{@{}c@{}}Command \\ Directness\end{tabular}} & \multicolumn{2}{c|}{Safety Metrics} & \multicolumn{2}{c|}{Comfort Metrics} & Time Efficiency & \multirow{3}{*}{Driving Score$\uparrow$} \\ \cline{4-8}
&  &  & \begin{tabular}[c]{@{}c@{}}Time to \\ Collision($s$)$\uparrow$\end{tabular} & \multicolumn{1}{c|}{\begin{tabular}[c]{@{}c@{}}Speed \\ Variance($m^2/s^2$)$\downarrow$\end{tabular}} & \begin{tabular}[c]{@{}c@{}}Mean Absolute\\ Acceleration($m/s^2$)$\downarrow$\end{tabular} & \multicolumn{1}{c|}{\begin{tabular}[c]{@{}c@{}}Mean Absolute\\ Jerk($m/s^3$)$\downarrow$\end{tabular}} & \begin{tabular}[c]{@{}c@{}}LLM \\ Latency($s$)$\downarrow$\end{tabular} &  \\
\midrule
\multirow{12}{*}{Highway} & \multirow{4}{*}{Overtake} & I & 2.88 & 5.02 & 0.24 & 2.64 & 1.68 & \textcolor{red}{85.05} \\
 &  & II & 1.94 & 4.05 & 0.24 & 2.81 & 1.87 & \textcolor{green}{86.12} \\
 &  & III & 3.07 & 1.26 & 0.18 & 2.64 & 1.86 & \textcolor{green}{91.12} \\
 &  & Baseline & 3.26 & 2.91 & 0.35 & 2.83 & - & 86.00 \\ \cline{2-9}
 & \multirow{4}{*}{Following} & I & 6.52 & 0.94 & 0.15 & 2.35 & 1.61 & \textcolor{green}{87.10} \\
 &  & II & 7.84 & 1.11 & 0.15 & 2.38 & 1.64 & \textcolor{green}{86.23} \\
 &  & III & 6.78 & 1.37 & 0.09 & 2.31 & 1.64 & \textcolor{green}{86.26} \\
 &  & Baseline & 4.02 & 0.78 & 0.22 & 2.50 & - & 86.00 \\ \cline{2-9}
 & \multirow{4}{*}{Right Lane} & I & 8.77 & 1.69 & 0.17 & 2.39 & 1.32 & \textcolor{green}{90.88} \\
 &  & II & 4.54 & 1.18 & 0.15 & 2.44 & 1.83 & \textcolor{green}{91.51} \\
 &  & III & 7.29 & 0.23 & 0.13 & 2.61 & 1.22 & \textcolor{green}{92.18} \\
 &  & Baseline & 4.70 & 7.39 & 0.22 & 2.77 & - & 86.00 \\
\midrule
\multirow{8}{*}{Intersection} & \multirow{4}{*}{Not Yield} & I & 0.89 & 0.29 & 0.26 & 2.28 & 1.65 & \textcolor{green}{59.67} \\
 &  & II & 0.89 & 0.22 & 0.28 & 2.55 & 1.78 & \textcolor{green}{59.60} \\
 &  & III & 1.04 & 0.21 & 0.26 & 2.32 & 1.52 & \textcolor{green}{60.32} \\
 &  & Baseline & 1.14 & 0.46 & 0.46 & 2.34 & - & 56.00 \\ \cline{2-9}
 & \multirow{4}{*}{Yield} & I & - & 0.29 & 0.52 & 2.27 & 1.47 & \textcolor{green}{91.53} \\
 &  & II & - & 0.22 & 0.82 & 2.54 & 1.43 & \textcolor{green}{89.50} \\
 &  & III & - & 0.21 & 0.48 & 2.28 & 1.38 & \textcolor{green}{91.92} \\
 &  & Baseline & - & 1.67 & 0.90 & 2.32 & - & 86.00 \\
\bottomrule
\end{tabular}%
}
\end{table*}

\subsubsection{Evaluation Metrics}
\label{subsec:evalationmetrics}
Our evaluation framework for autonomous vehicles includes driving performance, time efficiency, and personalization. We analyze Talk2Drive's driving performance in terms of safety, using time-to-collision ($\tau$) and speed variance \first{($\sigma_0$)}, and comfort, \first{using mean absolute acceleration ($|\bar a|$)} and mean absolute jerk ($|\bar J|$). Time efficiency is measured by the LLM latency ($Lcy$), while human satisfaction with personalization is assessed using the takeover rate ($R$).

\textbf{Driving Performance:} 
The overall driving performance of Talk2Drive is reflected by a driving score ($Score$), which is a weighted sum of four sub-scores: TTC ($Score_{\tau}$), speed variance ($Score_{\sigma}$), \first{mean absolute acceleration ($Score_{|\bar a|}$)}, and mean absolute jerk ($Score_{|\bar J|}$).

\begin{equation}
Score = \sum{w_k \cdot Score_k}
\end{equation}

For TTC $\tau$, the critical threshold $\tau_c$ is based on human reaction time to take over and brake. A TTC greater than 1.5 seconds is often considered safe as the driver will have enough time to react to avoid rear-end collisions \cite{van1994time}. Therefore, it can be considered as a hard threshold where any value greater than $\tau_c$ has a score of zero.
\begin{equation}
    Score_{\tau} = 
        \begin{cases}
        0,  & \text{if  } \tau_{min} < \tau_c \\
        100, & \text{if  } \tau_{min} \geq \tau_c
        \end{cases}
\end{equation}
where $\tau_c$ is the critical threshold. \first{While 2.0 s is a common simulation margin, a 1.5 s threshold was selected for this real-world test based on existing research results on TTC value \cite{van1994time}.} For other metrics like speed variance, acceleration, and jerk, the thresholds are related to individual perception and would vary in different driving scenarios. Therefore, the score to quantify the corresponding performance is defined, with a higher score indicating better performance relative to the baseline value. For example, the sub-score for mean absolute jerk $|\bar J|$ can be denoted as:
\begin{equation}
    Score_{|\bar J|} = 100 - \gamma\cdot\frac{|\bar J|}{|\bar J|_{\text{baseline}}}
\end{equation}
 where $\gamma$ is the sensitivity factor, which is set empirically. The sub-scores for mean absolute acceleration and speed variance are obtained in the same way.


\textbf{LLMs' Latency for Time Efficiency:} 
Latency $Lcy$ measures the response time of the LLM, which is important for time-sensitive vehicle applications. To measure latency, the time difference between the moment a command is sent to the LLMs and when the LLMs return the LMPs is calculated.


\textbf{Takeover Rate for Personalization:}
The frequency of manual interventions indicates the model's ability to adapt to personalized preferences from different humans~\cite{wang2022gaussian}.

\subsection{Experiment Results}
\first{All participants issue random verbal commands to either the baseline or the Talk2Drive system, reflecting their current feelings. To avoid bias, the participants will be unaware of which system is in use.} They can decide when to take over the driving system, and the takeover rate will be recorded. Takeovers in this context are noted when humans find the current autonomous system unsatisfactory. Additionally, all driving data will be logged, and the ping latency for the experiments is at 200$\sim$400 ms.

\subsubsection{The Validation of Driving Performance of Talk2Drive}


To validate the driving performance of the Talk2Drive system, comprehensive experiments are conducted to compare the driving performances between the baseline and the Talk2Drive through the metrics defined in \cref{subsec:evalationmetrics}. As discussed earlier, the baseline values are from the average data of human drivers. The average driving performance for each scenario within each command directness is shown in \cref{tab:drivingperformance}. Note that for each expected driving behavior, the driving pattern is distinct. Therefore, a baseline is created for every driving behavior separately.

The driving score is the weighted sum of all the safety and comfort metrics, reflecting the overall driving performance. The driving scores are calculated and recorded in \cref{tab:drivingperformance}, with green texts highlighting scores higher than the baseline of the same driving behavior and red texts showing the lower scores. As shown in \cref{tab:drivingperformance},  regardless of the command's directness or the human's behavior, nearly all driving scores are better than the baseline, showing that this framework consistently provides relatively safe and comfortable driving experiences for humans in all scenarios.

The latency of the internet remains mostly consistent (around 300 ms) under the same network conditions. The empirical results from the tests reveal that the latency for the most responsive LLM falls between approximately 1.2 to 1.8 seconds. \first{While the observed 1.2 s to 1.8 s latency is acceptable for high-level, non-urgent decisions (e.g., responding to ``drive more conservatively''), it is necessary to emphasize that this latency makes the cloud-based LLM entirely unsuitable for any safety-critical, time-sensitive maneuvers (e.g., emergency braking or collision avoidance.)} This finding suggests considerable potential for integration into certain real-time applications where such response times are within acceptable limits. Specifically, LLMs demonstrate strong suitability for making high-level decisions or modifying driving behaviors in response to human commands. These scenarios require timely responses but do not typically demand extremely low latency.

\first{Additionally, the Talk2Drive framework is designed to guarantee safety by architecturally separating the high-latency LLM from the low-latency, real-time vehicle control. As shown in Fig.~\ref{fig:framework}, the LLM's role is strictly limited to that of a high-level parameter adjuster. It does not engage in direct, real-time driving; instead, it generates LMPs that modify configuration parameters (e.g., target velocity, look-ahead distance) for the underlying, mature autonomous driving software such as Autoware, which handles the actual trajectory following and control. Furthermore, as detailed before, the system does not execute LLM-generated LMPs blindly. Every generated command is passed through a two-stage safety-check mechanism before execution. This includes: (1) a format check to ensure the code is valid and (2) a parameter verification step to assess the feasibility and safety of the proposed values. For example, any generated parameter that violates predefined safety bounds (e.g., a target velocity above the road's speed limit) or is contextually impossible (e.g., an instruction to ``move left" when the vehicle is already in the leftmost lane) is automatically identified and blocked from execution. This rule-based safety layer ensures that the LLM can only influence the vehicle's driving style within a predefined, safe operational envelope, while all safety-critical responses remain the responsibility of the low-latency on-board controllers.}

\subsubsection{The Improvement on Personalization Performance}

This section explores the impact of integrating the Talk2Drive framework into autonomous driving systems to enhance personalization based on the LLM model GPT-4~\cite{openai_gpt-4_2023}. One of the central focuses of the system is its ability to offer drivers a personalized driving experience. Through Talk2Drive, individuals can express their preferences or feelings with varying degrees of directness, prompting the system to make corresponding adjustments. The commands collected are divided into three levels of directness, as explained in Tab. \ref{tab:directlevel}. 

For each command directness within each expected driving behavior, the system's responses to various responses given by human participants are collected and calculated into the defined metrics. Here, the takeover rate is utilized as the primary metric and is evaluated both pre- and post-integration of Talk2Drive. We incorporate a rule-based system with a keyword-trigger logic as the baseline for comparison. For instance, in the highway scenario, commands like ``accelerate," ``left," and ``right" prompt the system to execute relevant actions such as speeding up or changing lanes accordingly. The results are gathered in diverse driving scenarios on human drivers with diverse driving styles. 

\begin{table}[t]
\caption{Takeover Rate Improvement through Personalization.}
\label{tab:personalization2}
\centering
\resizebox{\columnwidth}{!}{%
\begin{tabular}{c|cccc}
\toprule
\begin{tabular}[c]{@{}c@{}}Driving\\ Scenario\end{tabular} & \begin{tabular}[c]{@{}c@{}}Command \\ Directness\end{tabular} & Baseline & Talk2Drive & Reduction \\
\midrule
\multirow{3}{*}{Highway} & I & 0.33 & 0.07 & 78.8\% \\
 & II & 0.63 & 0.20 & 68.3\% \\
 & III & 0.77 & 0.31 & 59.7\% \\
\midrule
\multirow{3}{*}{Intersection} & I & 0.33 & 0.11 & 66.7\% \\
 & II & 0.71 & 0.29 & 59.2\% \\
 & III & 0.48 & 0.21 & 56.3\% \\
\midrule
\multirow{3}{*}{Parking} & I & 0.07 & 0 & 100\% \\
 & II & 0.20 & 0 & 100\% \\
 & III & 0.67 & 0.24 & 64.2\% \\
\bottomrule
\end{tabular}%
}
\end{table}

As shown in Tab. \ref{tab:personalization2}, the takeover rate in all driving scenarios decreases significantly, ranging from a 56.3\% reduction to the complete elimination of takeover cases, demonstrating the Talk2Drive system's ability to personalize the driving experience based on human preference. Additionally, one important finding from the results is that for all levels of command directness, the Talk2Drive system shows significant improvements compared to the baseline systems.

\subsubsection{The Effectiveness of the Memory Module}


To further investigate the performance of the memory module, experiments comparing personalization across two settings: conditions without the memory module, and conditions with the memory module are conducted. These experiments take place in the parking lot, where the vehicle follows a pre-defined trajectory and adjusts speed based on human inputs. Humans take over in this context when they find the adjusted speed unsatisfactory. As demonstrated in Tab. \ref{tab:takeoverrate}, there are significant reductions in takeover rate with Talk2Drive, regardless of the driver's aggression or conservatism levels. 

The results in Tab. \ref{tab:takeoverrate} reveal that the inclusion of the memory module leads to a marked reduction in the takeover rate. The utilization of Talk2Drive framework without the memory module brings takeover rates between 0.14 and 0.29. When implementing Talk2Drive framework with the memory module, the takeover rate can be further decreased to only 0.07. Compared to their personalization performance, Talk2Drive with a memory module can reduce takeover rates by up to 65.2\% compared to those without the memory module, which illustrates the benefits of a history-recording module in achieving a more personalized driving experience.

\begin{table}[!t]
\centering
\caption{Effectiveness of Memory Modules in Influencing Takeover Rates}
\begin{tabular*}{0.9\linewidth}{@{\extracolsep{\fill}}c|cc@{\extracolsep{\fill}}}
    \toprule
     \makecell{Driver } &  \makecell{Without Memory Module} & \makecell{With Memory Module}  \\
    \midrule
    A & 0.14  & 0.07 (50.0\%$\downarrow$)\\
    B & 0.23  & 0.08 (65.2\%$\downarrow$) \\
    C & 0.29  & 0.18 (37.9\%$\downarrow$)\\
    \bottomrule
\end{tabular*}
\label{tab:takeoverrate}
\end{table}

\section{Real-Vehicle Experiments: On-Board VLM for Personalized Motion Control}
\label{sec:onboard}
\begin{figure*}[!t]
    \centering
    \includegraphics[width=0.9\textwidth]{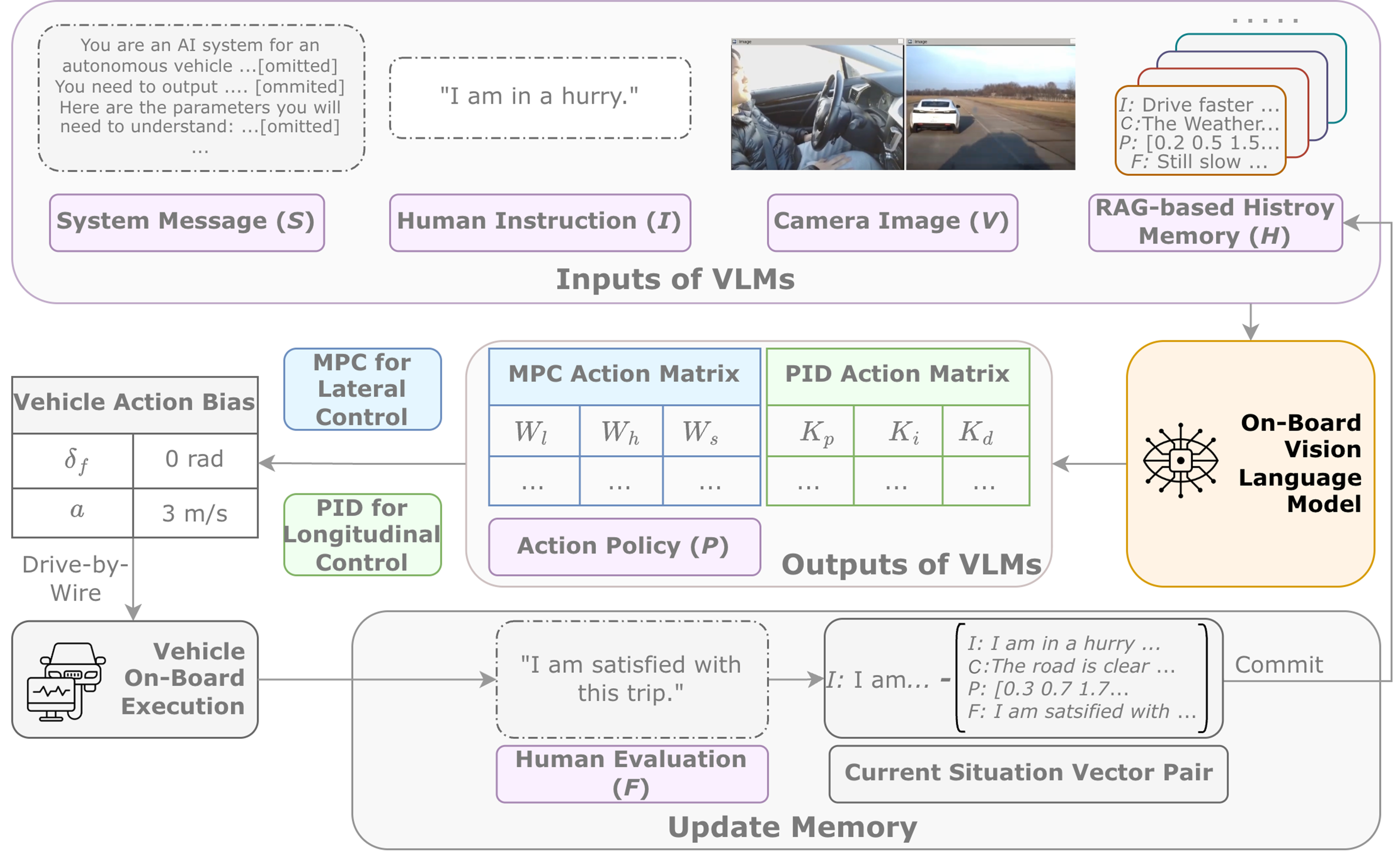}
    \caption{The proposed framework for personalized motion control with an on-board vision language model in real-vehicle experiments. The system processes four primary inputs: system messages, human instructions, camera images, and historical context retrieved from a retrieval-augmented generation (RAG) memory module. Based on these inputs, the model generates an action policy consisting of specific matrices that adjust the parameters for model predictive control (lateral) and PID control (longitudinal). These adjusted parameters regulate steering and acceleration through a drive-by-wire system to execute the driving maneuvers. The process concludes with a feedback loop where human evaluations are committed to the memory module, updating the historical database to refine future personalization.}
    \vspace{-5mm}
    \label{fig:main}
\end{figure*}

In this section, we present our on-board VLM for personalized motion control in autonomous driving, designed to accommodate individual driving styles. Our approach leverages a compact 8B-parameter VLM, fine-tuned from Qwen-VL~\cite{bai2023qwenvlversatilevisionlanguagemodel}, which processes both visual information (including weather conditions, road types, and traffic conditions) and verbal commands to generate personalized control strategies for each user. The reduced scale of this VLM enables edge deployment while maintaining command interpretation and reasoning capabilities, allowing the system to effectively understand and respond to implicit human instructions. The overall framework is shown in Fig. \ref{fig:main}.   

\subsection{Problem Statement}

We adopt the traditional module-based autonomous driving framework that includes the full pipeline from perception to motion control, and our focus is specifically on enhancing the decision-making process at the motion control level, adapting the control performance to accommodate different human driving styles. The goal of the proposed system is to translate both verbal commands $I$ and visual inputs $V$ into executable control sequences for the motion control process. The on-board VLM acts as a translation mechanism $f:(I,V)\rightarrow P$ that generates a policy $P$, which is then fed into predefined maneuver programs.

Additionally, system messages $S$ are sent to our VLM to specify both the tasks and adjustment strategies. In our setting, we follow the \cref{sec:real-vehicle}, $S$ is generated through a predefined language generator.

Simultaneously, to further enhance personalization, we implement a RAG-based memory module to build a database storing historical human-vehicle interactions. Whenever a human activates this system, only relevant historical scenarios $H$ are retrieved and provided to the VLM as reference. After each trip, users can provide feedback $F$ on the generated control policy $P$ for the current situations (including instructions $I$ and visual input $V$), which helps refine the VLM's reasoning process. Subsequently, the instructions $I$, scene description $C$ (generated by on-board VLM), policy $P$, and feedback  $F$ are packaged as a historical vector-based data entry and stored in the RAG database. Therefore, there are three procedures in our system:
\vspace{-2mm}
\begin{equation}
    \begin{aligned}
        \text{VLM Execution}:\quad & P \xleftarrow{\text{VLM}} f(I,S,V,H); \\
        \text{Human Feedback}:\quad & F \xleftarrow{\text{Human}} [I,V,P] ; \\
        \text{Memory Update}:\quad & H \leftarrow \left[I,C,P,F\right]
    \end{aligned}
\end{equation}

\subsection{Our Framework}
\subsubsection{System Input}

As illustrated in Fig. \ref{fig:main}, our fine-tuned on-board VLM processes four critical inputs for reasoning. The primary inputs consist of visual data $V$ from the on-board camera and natural language commands $I$ which are converted from verbal human instructions $I$ using the open-source local speech recognition model ``Whisper~\cite{radford2022robustspeechrecognitionlargescale}.'' Notably, our system leverages contextual and environmental information captured in the visual inputs $V$, including weather conditions, traffic conditions, and road characteristics. For instance, the system automatically adopts a more conservative driving policy during nighttime operations or adverse weather conditions. 

Additionally, a pre-defined system message generator is employed to produce a customized system message $S$, which is then simultaneously sent to the VLM. 


Furthermore, the VLM incorporates relevant interaction history $H$ extracted from our RAG-based memory module as contextual input, which includes previous human instructions $I$, scene descriptions $C$, executed actions $A$, and user feedback $F$. Note that the scene descriptions $C$ are generated by the on-board VLM to describe the image $V$. This historical context enables the VLM to generate more appropriate responses by considering past interactions and human feedback. A detailed discussion of how our memory module works will be presented in subsection \ref{subsec:mm}.


\subsubsection{Reasoning and Action Generation}

In our approach, reasoning within the VLM framework enables the interpretation of diverse driving scenarios and user instructions to generate actionable outputs. Traditional controllers in motion control typically rely on a default set of parameters; however, following the approach in \cite{sha2023languagempclargelanguagemodels}, our VLM will generate two distinct action matrices to separately manage the PID controller for longitudinal movement and the model predictive control (MPC) for lateral movement. These matrices translate the model’s understanding of the environment and user preferences into precise control actions, guiding the autonomous vehicle's behavior. Specifically, they are used by the controllers to generate acceleration $a$ and steering angle $\delta_f$, which are executed by the vehicle’s ECU. The ECU then sends low-level control signals to the drive-by-wire system developed by AutonomousStuff~\cite{autostuff}, enabling smooth and responsive vehicle operation. The general process of this subsection can be shown below:
\begin{equation}
    \begin{aligned}
        \text{Output Action } &P:= 
        \begin{bmatrix}
            K_p & K_i &K_d \\
            W_l & W_h & W_s
        \end{bmatrix}    \\
        \text{Action Execution }& P  \xrightarrow{Controllers}\left[\delta_f,a\right] \rightarrow ECU
    \end{aligned}
\end{equation}

\subsubsection{RAG-Enhanced Memory Module}
\label{subsec:mm}

Given that our 8B-parameter VLM lacks the extensive reasoning capabilities of larger, 100B-200B parameter models, we employ a RAG-based approach~\cite{lewis2020retrieval} and integrate it to enhance reasoning and enable human feedback learning. This system is built upon the Chroma vector database~\cite{chroma_ai-native_2023}, enabling efficient storage history interactions and retrieval of similar driving scenarios.

The memory module is uniquely created for each user, ensuring a personalized driving experience follows individual preferences and patterns. It stores driving-related information in a structured format comprising commands paired with corresponding context tuples: 
\begin{equation}
    \{I - ( I, C, P, F)\}
\end{equation}
When processing a new driving scenario, the instruction $I$ is used for similarity matching to retrieve the top-k similar prior situations. The associated data values are then sent to the VLM, enhancing decision-making with relevant context and supporting personalization. This RAG-enhanced memory enables the VLM to handle similar situations with greater accuracy and consistency, improving the vehicle’s responsiveness to unique user preferences and enhancing the overall driving experience.

\subsubsection{Multi-Controller Joint Motion Control}

As shown in Fig. \ref{fig:main}, we implement a decoupled control strategy that separates lateral and longitudinal vehicle motion control. The lateral control is handled by MPC calculating the front steering angle $\delta_f$, while longitudinal control is managed through a PID controller calculating longitudinal acceleration $\alpha$. The motion planning module in the upper layer provides trajectories consisting of waypoints, which our motion control system tracks. The calculated $\alpha$ and $\delta_f$ are then transmitted to the drive-by-wire system developed by AutonomousStuff~\cite{autostuff} for precise control of throttle, braking, and steering.

For the longitudinal control, the PID controller calculates the $\alpha$ for each time step $\Delta t$ to minimize the velocity error $e_v$, which is the difference between the current velocity $V_{\text{current}}$ and the desired velocity $V_{\text{ref}}$. 
 \begin{equation}
     \begin{aligned}
      \alpha(t) = K_p e_v(t) + K_i \sum_{i=0}^{t} e_v(i) \Delta t + K_d \frac{\Delta e_v(t)}{\Delta t}
      \end{aligned}
 \end{equation}
where $K_p$, $K_i$, and $K_d$ are the proportional terms, integration terms, and derivative terms that will be contained in the action matrix generated by our VLM.

For the lateral control, our MPC approach utilizes a linear vehicle dynamics model \cite{snider2009automatic} to predict future states and optimize the front steering angle, \(\delta_f\), over a finite prediction horizon. With the prediction model \cite{snider2009automatic}, the control increment is then obtained by solving a Quadratic Program (QP) \cite{bemporad2000explicit} to minimize the cost function \(J\) in the MPC:
\begin{equation}
    J = E^T W_Q E + \Delta_f^T W_R \Delta_f
\end{equation}
where \(E\) is the predicted state calculated by the prediction model, and \(\Delta_f\) is the future control input. The \(W_Q\) and \(W_R\) are weighting matrices that penalize tracking state deviations and control effort. Our VLM generates the action matrix that primarily considers three key components: the weight for lateral error ($W_l \in W_Q$), the weight for heading error ($W_h \in W_Q$), and the weight for the squared terms of speed and steering inputs ($W_s \in W_R$). These weighting factors are selected as they demonstrate the most significant impact on lateral control performance.

\subsubsection{Efficient On-Board VLM Module}

We generate a dataset of 10,000 image-instruction pairs, each labeled with the desired action, to create a comprehensive training set for fine-tuning our on-board VLM. This VLM is based on the Qwen-VL architecture~\cite{bai2023qwenvlversatilevisionlanguagemodel}, which we fine-tune using the Low-Rank Adaptation (LoRA) method~\cite{hu2021loralowrankadaptationlarge} (a type of Parameter Efficient Fine-Tuning (PEFT)~\cite{xu2023parameterefficientfinetuningmethodspretrained}), enabling significant customization while preserving computational efficiency. To optimize for on-board deployment, we apply 4-bit Activation-Aware Weight Quantization (AWQ)~\cite{lin2024awqactivationawareweightquantization}, compressing the VLM to increase inference speed without sacrificing too much accuracy. This combination of techniques ensures a responsive, on-board VLM suited to real-time response.

\paragraph{Dataset Collection}
We develop a specialized training dataset to fine-tune the Qwen-VL model~\cite{bai2023qwenvlversatilevisionlanguagemodel}, consisting of 10,000 semi-human-annotated image-text pairs. Each image, representing a traffic scene sourced from the NuScenes dataset~\cite{caesar_nuscenes_2020}, which includes numerous diverse traffic scenarios, is paired with a human-provided instruction and a corresponding action label in the form of a controller action matrix, guiding the model’s response in different traffic scenarios.

The human instructions are also very diverse, ranging from explicit commands like ``speed up'' to more implicit cues such as ``I am in an urgent situation.'' This diversity allows the VLM to interpret both clear and vague inputs, improving its ability to understand complex human intentions. To enhance the model's responsiveness to different driving styles, we annotate each image with three different instruction types (aggressive, moderate, and conservative), each paired with a corresponding action. This approach ensures that the VLM can adapt its behavior to match various driving styles, enabling it to respond flexibly and contextually across diverse traffic conditions.

\paragraph{LoRA Finetune}
We apply the LoRA method to fine-tune our Qwen-VL model. LoRA operates by freezing the pre-trained model weights and introducing trainable, low-rank decomposition matrices into each layer of the Transformer architecture. This approach significantly reduces the number of trainable parameters required, making fine-tuning more efficient.

The fine-tuning process of our VLM is conducted on a cluster of four NVIDIA A100 GPUs, each equipped with 40GB of memory. The model is trained over five epochs with a per-device batch size of two for training and one for validation, using a learning rate of 1e-5. Additionally, we implemented gradient accumulation with eight steps, allowing for effective larger batch processing. This setup enables the entire training process to be completed in approximately five hours, ensuring both accuracy and efficiency in model performance.

\paragraph{Compression and On-Board Deployment of VLM}

AWQ~\cite{lin2024awqactivationawareweightquantization} is a hardware-friendly technique for low-bit, weight-only quantization, specifically designed for VLM. AWQ minimizes quantization error by identifying the 1\% salient weights, which are then scaled using an equivalent transformation to preserve their precision. We apply AWQ to quantize our model to INT4, achieving improved quantization performance suited for on-board deployment. Additionally, we utilize the LMDeploy toolkit~\cite{2023lmdeploy} to optimize inference time. This enhancement is made possible through features such as persistent batching, blocked KV cache, tensor parallelism, and optimized CUDA kernels, all of which contribute to high-performance, low-latency operation.

\subsection{Real-World Experiment}
\label{sec:exper}
To comprehensively evaluate our system's performance, we conduct a series of experiments assessing its ability to provide safe, comfortable, reliable, and personalized driving experiences. We employ multiple evaluation metrics: a driving score to measure driving performance, including safety, comfort, and alignment with environmental conditions and human instructions; takeover frequency to assess personalization capabilities. Additionally, we perform an ablation study to examine the effectiveness of the memory module.  The demo for our real-vehicle experiment is publicly available\footnote{Real-vehicle experiment demo video with on-board VLM: \url{https://www.youtube.com/watch?v=IPZo2USvkAw}}.

\begin{table*}[ht!]
\caption{Driving Performance Validation. $\downarrow$: Lower Values are Better. $\uparrow$: Higher Values are Better.}
\label{tab:drivingperformancevlm}
\centering
\resizebox{2\columnwidth}{!}{%
\begin{tabular}{c|c|ccc|cccc|c|cc|c}
\toprule
\multirow{3}{*}{\begin{tabular}[c]{@{}c@{}}Driving \\ Scenario\end{tabular}} & \multirow{3}{*}{Model} & \multicolumn{3}{c|}{Safety Metrics} & \multicolumn{4}{c|}{Comfort Metrics} & Time Efficiency & \multicolumn{2}{c|}{Alignment} & \multirow{3}{*}{Driving Score$\uparrow$} \\ \cline{3-12}
&  & \begin{tabular}[c]{@{}c@{}}Time to \\ Collision($s$)$\uparrow$\end{tabular} & \begin{tabular}[c]{@{}c@{}} \first{$\sigma_{x}$} \\ ($m^2/s^2$)$\downarrow$\end{tabular} & \begin{tabular}[c]{@{}c@{}}
\first{$\sigma_{y}$} \\ ($10^{-2} m^2/s^2$)$\downarrow$\end{tabular} & \begin{tabular}[c]{@{}c@{}}$|\bar{\alpha_{x}}|$\\ ($m/s^2$)$\downarrow$\end{tabular} & \begin{tabular}[c]{@{}c@{}}$|\bar{J_{x}}|$\\ ($m/s^3$)$\downarrow$\end{tabular} & \begin{tabular}[c]{@{}c@{}}$|\bar{\alpha_{y}}|$\\ ($10^{-1} m/s^2$)$\downarrow$\end{tabular} & \begin{tabular}[c]{@{}c@{}}$|\bar{J_{y}}|$\\ ($m/s^3$)$\downarrow$\end{tabular} & \begin{tabular}[c]{@{}c@{}}VLM \\ Latency($s$)$\downarrow$\end{tabular} & \begin{tabular}[c]{@{}c@{}}Command\\ Alignment$\uparrow$\end{tabular} & \begin{tabular}[c]{@{}c@{}}Scenario\\ Alignment$\uparrow$\end{tabular} &  \\
\midrule
\multirow{3}{*}{Acceleration} & Baseline & 2.44& 28.8 & 0.36 & 0.78 & 3.27 & 0.46 & 0.44 & - & 92.0 & 60.0 & 75.6\\
& GPT-4o & 2.52 & 30.0 & 0.39 & 0.83 & 3.40 & 0.52 & 0.52 & 5.82 & 92.9 & \textbf{71.3} & 76.4\\
& \textbf{Ours} & 2.46 & 30.8 & 0.39 & 0.81 & 3.07 & 0.53 & 0.81 & 1.98 & \textbf{96.3} & 60.9 & \textbf{76.5} \\
\midrule
\multirow{3}{*}{\begin{tabular}[c]{@{}c@{}}Lane\\ Change\end{tabular}} & Baseline & 2.44 & 3.91 & 1.65 & 0.37 & 3.14 & 0.88 & 1.01 & - & 88.5 & 60.0 & 74.5\\
& GPT-4o & 2.71 & 3.88 & 2.23 & 0.53 & 4.38 & 1.13 & 1.39 & 4.84 & 90.4 & \textbf{88.6} & \textbf{78.4}\\
& \textbf{Ours} & 2.15 & 4.07 & 2.15 & 0.41 & 3.35 & 0.98 & 1.02 & 1.83 & \textbf{92.2} & 71.9 & 77.5\\
\midrule
\multirow{3}{*}{\begin{tabular}[c]{@{}c@{}}Left\\ Turn\end{tabular}} & Baseline & - & 1.12 & 7.52 & 0.22 & 1.81 & 1.29 & 1.36 & - & 88.0 & 60.0 & 70.4\\
& GPT-4o & - & 0.93 & 11.5 & 0.29 & 2.75 & 2.11 & 2.40 & 5.23 & \textbf{91.3} & \textbf{85.0} & 71.4\\
& \textbf{Ours} & - & 0.94 & 6.74 & 0.19 & 1.67 & 1.33 & 1.32 & 1.64 & 90.2 & 67.8 & \textbf{74.4}\\
\bottomrule
\end{tabular}%
}
\vspace{-5mm}
\end{table*}

\begin{figure}[t]
    \centering
    \includegraphics[width=\linewidth]{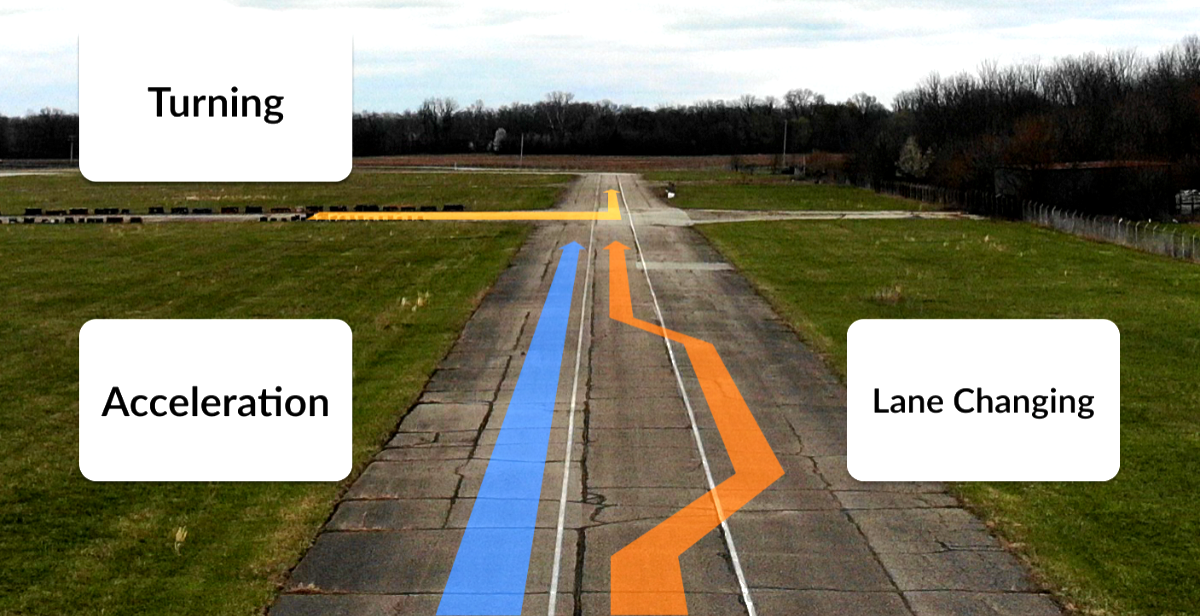}
    \caption{This figure provides an overview of the closed-loop test track layout used for personalized motion control real-world validation experiments.}
    \label{fig:test_track}
\end{figure}

\subsubsection{Experiment Setup}\label{subsec:setup}

The field experiments\footnote{All experiments conducted in this study satisfy all local traffic guidelines and guarantee the safety of the participants. A human always sits in the driver’s seat of the autonomous vehicle to monitor its status and get ready to take over.} aim at validating the real-world performance of our personalized motion control system. \first{We include three types of driving behaviors (accelerating, lane changing, and turning) to comprehensively validate control over steering, throttle, and braking.} An overview of the test track and driving behaviors is shown in Fig. \ref{fig:test_track}. For both acceleration and lane change scenarios, a lead vehicle is positioned 30 $m$ ahead of the ego vehicle, accelerating from static to 45 $km/h$ with an acceleration of 1.26 $m/s^2$. In the acceleration scenario, the ego vehicle accelerates from a complete stop to reach 50 $km/h$. In the lane change scenario, the ego vehicle maintains 50 $km/h$ while overtaking the lead vehicle. For the intersection turning scenario, the ego vehicle navigates a curve with a radius of 23.89 $m$ at a constant speed of 30 $km/h$. \first{This study includes seven participants with diverse demographic characteristics. The participants consisted of 61.4\% male and 28.6\% female drivers, with ages ranging from 23 to 30 years (mean = 26.42, std = 3.24). Their driving experience varies considerably (mean = 6.42 years, std = 4.27 years). All participants hold valid U.S. driving licenses.}

\subsubsection{System Driving Performance}

\begin{figure}[t]
    \centering
    \includegraphics[width=\linewidth]{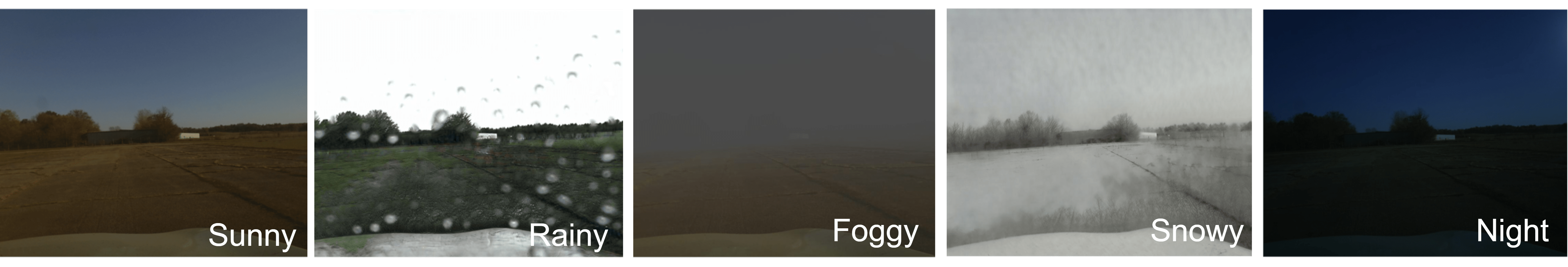}
    \caption{Sample vision inputs from different weather scenarios.}
    \vspace{-5mm}
    \label{fig:weather}
\end{figure}

To showcase our system's driving performance, we conduct comparative experiments against two systems: a baseline system using pre-defined controller parameters for general safety and comfort and a system utilizing GPT-4o with few-shot learning. We evaluate across three primary meta-driving scenarios (acceleration, lane changing, and turning), with each scenario tested under ten different commands and five weather conditions (sunny, rain, fog, snow, and night) to test the model's vision understanding as shown in Fig.~\ref{fig:weather}.

\paragraph{Evaluate Metrics} The models are then assessed based on four key aspects: safety, comfort, time efficiency, and alignment, as shown in Tab.~\ref{tab:drivingperformancevlm}. An overall driving score $Score$ is then calculated as a weighted sum of the individual metric scores, denoted as:
\begin{equation}
Score = \sum{w_k \cdot Score_k}
\end{equation}
where $k$ includes all ten metrics: Time to Collision ($\tau$), longitudinal and \first{lateral speed variance ($\sigma_x$ and $\sigma_y$)}, lateral and longitudinal mean absolute acceleration ($|\bar{\alpha_{x}}|$ and $|\bar{\alpha_{y}}|$), lateral and longitudinal mean absolute jerk ($|\bar{J_{x}}|$ and $|\bar{J_{y}}|$), VLM Latency, Command Alignment and Scenario Alignment.  All the scores $Score_k$ range from 0 to 100, while the weights of scores $w_k$ are empirically tuned for each driving scenario. For instance, longitudinal parameters have higher weights in acceleration scenarios, while lateral parameters are weighted more heavily in turning scenarios. The safety and comfort metrics are followed as defined in \cref{subsec:evalationmetrics}.


The alignment evaluation consists of two aspects. Command Alignment quantifies the deviation between the model output and the expected parameter range, calculated as a weighted average across six control parameters. For each parameter, we establish three ranges based on past experiments, corresponding to aggressive, conservative, or moderate driving styles.  Taking the PID controller's proportional parameter $K_p$ as an example, the score is calculated as:

\begin{equation}
\resizebox{0.4\textwidth}{!}{%
$
Score_{K_{p}}^{*} = 
\begin{cases} 
\frac{100(K_p - K_{p, \min})}{K_{p, \text{lower}} - K_{p, \min}}, & K_p \in [K_{p, \min}, K_{p, \text{lower}}), \\[10pt]
100, & K_p \in [K_{p, \text{lower}}, K_{p, \text{upper}}), \\[10pt]
\frac{100(K_{p, \max} - K_p)}{K_{p, \max} - K_{p, \text{upper}}}, & K_p \in [K_{p, \text{upper}}, K_{p, \max}], \\[10pt]
0, & \text{else}.
\end{cases}%
$
}
\end{equation}
where $K_{p, \min}$ and $K_{p, \max}$ are the minimum and maximum overall parameter range obtained through experiments, while $K_{p, \text{lower}}$ and $K_{p, \text{upper}}$ are determined by the command intention labeled by human experts. The scenario alignment score computes whether the model can capture details of the scene through vision input and act more aggressively or conservatively based on the current condition. It is calculated by tallying the percentage of instances where the model gives more conservative parameter sets in adverse weather conditions compared to the sunny clear scenarios. 

\paragraph{Result} As shown in Tab.~\ref{tab:drivingperformancevlm}, the command alignment score of our VLM model is similar to or greater than GPT-4o, showing our model has high reasoning capabilities regarding the personalization of control parameters. As for the scenario alignment, our model shows significantly better performance than baseline conditions in lane changing and left turn scenarios but a score very similar to the baseline scenario. We think this is mostly due to the model considering acceleration on a clear straight road in less dangerous situations and thus does not act too conservatively. In terms of the overall Driving Score, our model also surpasses the performance of baseline models and even GPT-4o in some scenarios, indicating our model provides the most well-rounded driving experience.

\subsubsection{Human-in-the-Loop Validation}

\begin{figure}[!t]
    \centering
    \includegraphics[width=0.7\linewidth]{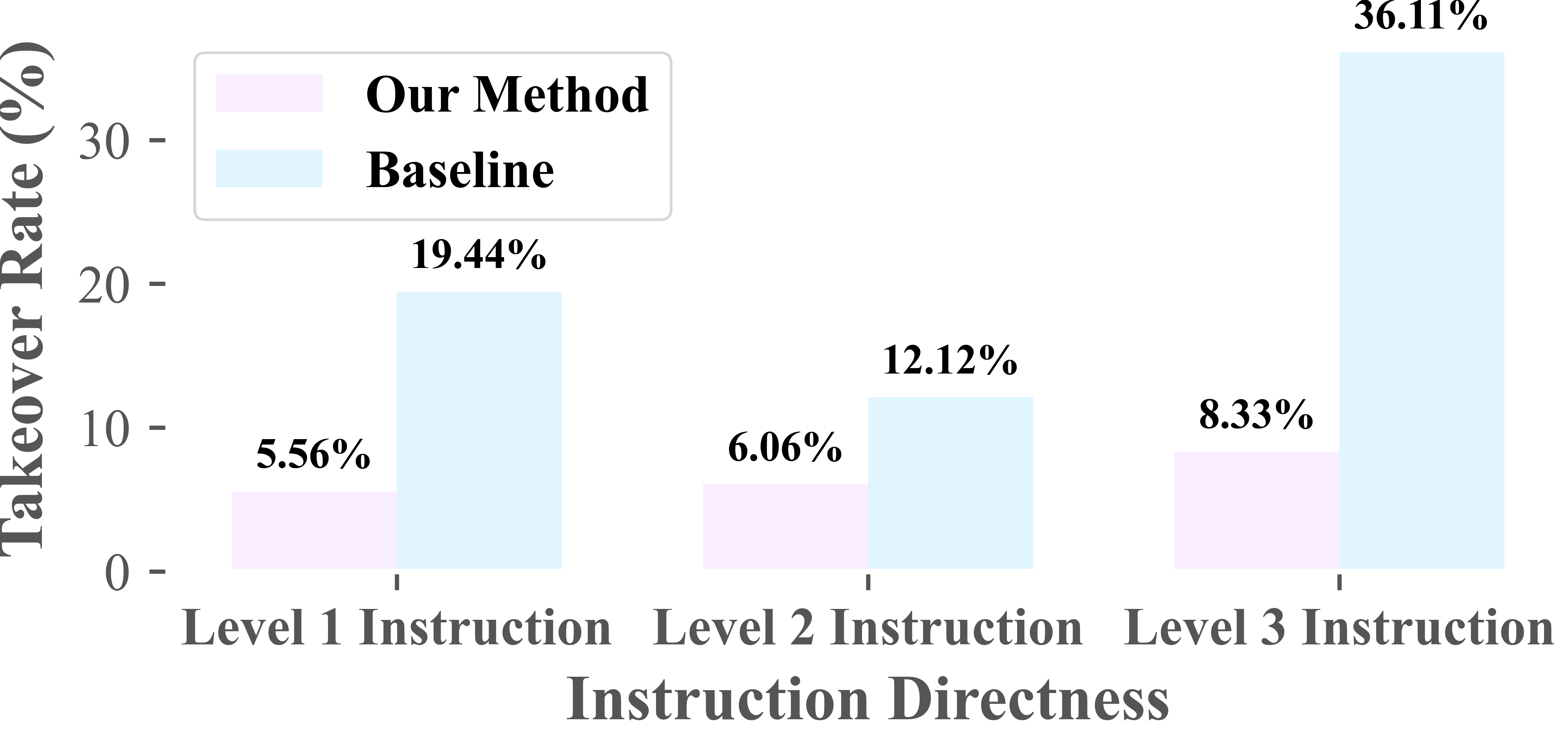}
    \caption{Takeover rate comparison between the baseline and our method.}
    \label{fig:takeover}
\end{figure}

This subsection evaluates the effectiveness of our method in reducing takeover rates compared to the baseline system. The baseline consists of the traditional autonomous vehicle controller with default settings, where two unified controllers manage vehicle operations on longitudinal and lateral respectively. We compare this conventional approach against our VLM-based adaptive motion control system. Throughout the experiments, participants could provide explicit instructions or implicit preferences/feedback with varying degrees of directness, prompting the system to make corresponding adjustments. The instructions were categorized into three levels of directness, as defined in \cref{tab:directlevel}.

Every participant is supposed to provide at least five instructions for each scenario. \first{For each instruction-scenario pair, participants completed two trips; one with the baseline system and one with our VLM solution. To ensure unbiased evaluation, participants are not informed which system they are using during each trip.} We use the takeover rate as our primary metric to evaluate the system's ability to provide personalized driving experiences.

The results demonstrate that our VLM-based method consistently outperforms the baseline system across all three levels of instruction directness. With Level 1 (direct) instructions, our method achieves a 5.56\% takeover rate versus the baseline's 19.44\%, representing a 71.4\% reduction. For Level 2 (moderately direct) instructions, the rates are 6.06\% versus 12.12\%, showing a 50\% improvement. Most notably, for Level 3 (indirect) instructions, our method achieves 8.33\% versus the baseline's 36.11\%, marking a 76.9\% reduction in takeover rate. The result demonstrates our system's better capability in interpreting and acting based on user preferences, regardless of how explicitly they are communicated.

\subsubsection{Ablation Study on Memory Module}

\begin{figure}[t!]
    \centering
    \includegraphics[width=0.7\linewidth]{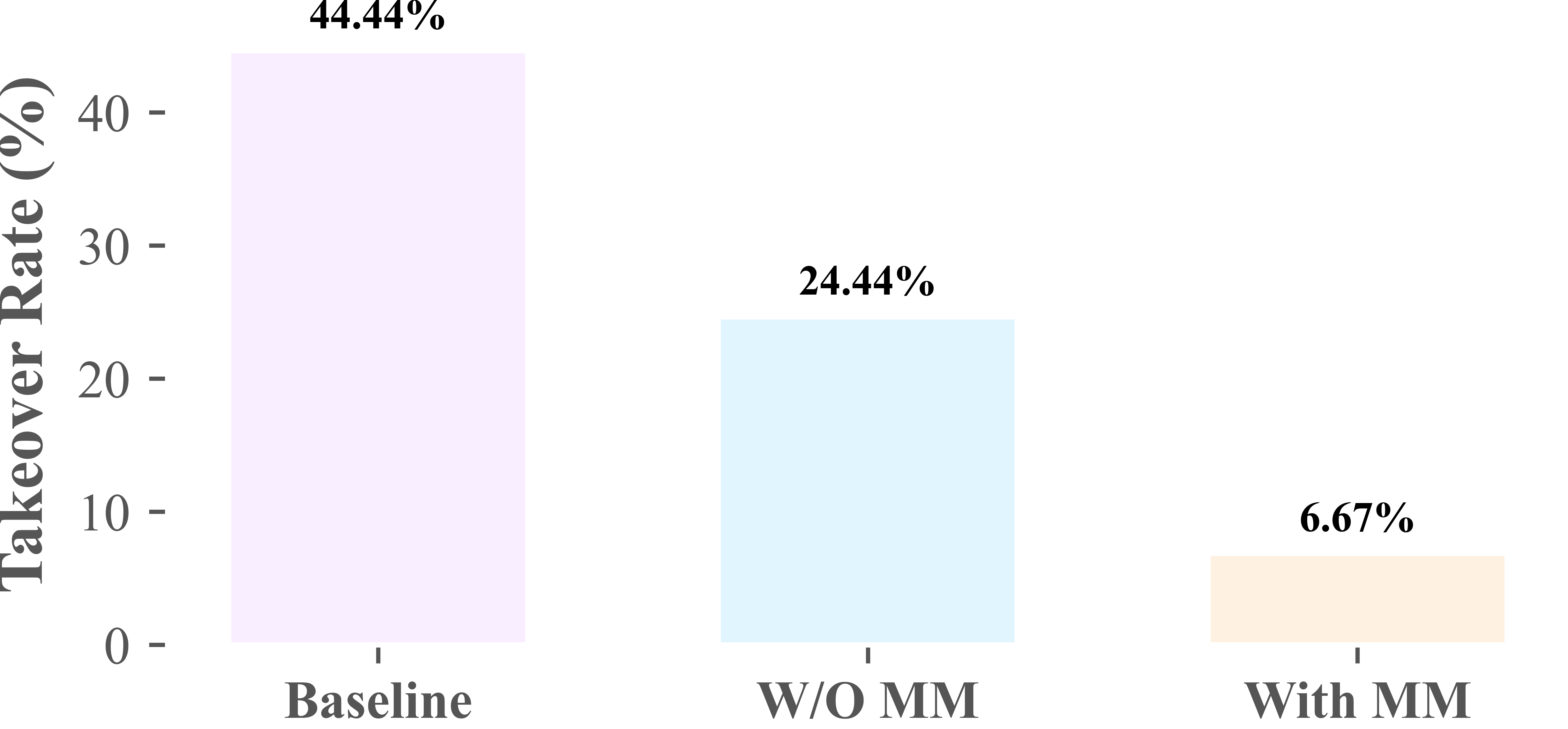}
    \caption{Effectiveness of memory modules in takeover rates.}
    \vspace{-7mm}
    \label{fig:rag_takeover}
\end{figure}

To further validate the effectiveness of our RAG-based Memory Module (MM),  we conduct an ablation study with three human participants, comparing three configurations: our complete system with MM, the system without MM (W/O MM), and the baseline. The results demonstrate the significant impact of the MM on reducing takeover rates.

As shown in Fig. \ref{fig:rag_takeover}, With three participants, our complete system with the memory module achieves the lowest average takeover rate of 6.67\%. When we remove the memory module while keeping other components the same, the average takeover rate increases substantially to 24.44\%. The baseline system performs worst with a 44.44\% average takeover rate. These results indicate a 72.7\% reduction in the takeover rate when adding the memory module to our base architecture and an 85\% overall reduction compared to the baseline. This significant improvement suggests that the RAG-based memory module plays a crucial role in maintaining personalized vehicle control by effectively utilizing historical interactions and user preferences.

\begin{figure*}[!t]
    \centering
    \includegraphics[width=\textwidth]{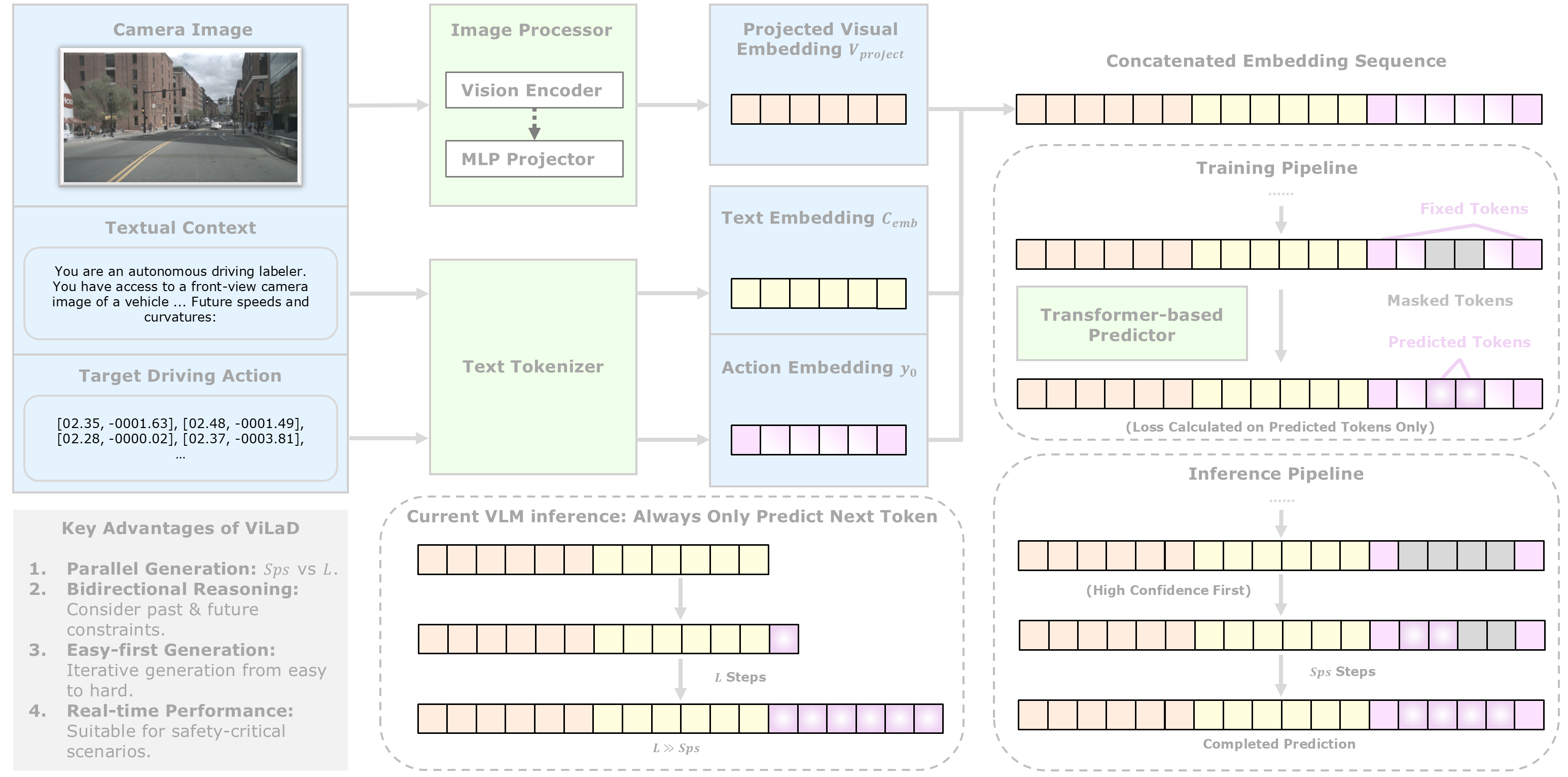}
    \caption{This diagram shows the complete ViLaD framework for end-to-end autonomous driving: The left side illustrates multimodal input processing, where camera images and textual context are encoded and concatenated with action embeddings. The right side demonstrates the core innovation with training and inference pipelines: during training, only driving decision tokens are masked while visual/textual inputs remain unchanged; during inference, the model progressively demasks predictions starting with high-confidence decisions first. The bottom highlights ViLaD's key advantages over autoregressive VLMs, including parallel generation, bidirectional reasoning, easy-first generation, and computational efficiency.}
    \label{fig:vilad_main}
\end{figure*}

\first{
\section{Future Trends of LLM4AD: Language Diffusion Models in Autonomous Driving}
\label{sec:vilad}
Current VLM-based end-to-end autonomous driving systems rely on autoregressive architectures that generate decisions through sequential next-token prediction. 
This autoregressive paradigm, while successful for many natural language processing tasks, presents some fundamental limitations when applied to the time-sensitive and spatially complex domain of real-world driving scenarios. 
These models generate decisions sequentially, token-by-token, which is \textit{computationally inefficient} and \textit{conceptually inflexible}: 
First, the sequential generation process creates significant inference latency, as each token must be generated one by one, making real-time decision-making challenging for safety-critical scenarios. Second, the left-to-right generation limits the model's ability to consider global spatial relationships and future trajectory implications simultaneously, potentially leading to suboptimal driving decisions that lack comprehensive scene understanding. In the dynamic and unpredictable environment of the road, this lack of foresight and adaptability may cause a significant challenge to safety and reliability.
Moreover, autoregressive VLMs suffer from the ``reversal curse'' phenomenon~\cite{berglund2024reversalcursellmstrained}, where models trained on sequential patterns struggle with tasks requiring bidirectional reasoning or reverse inference. In autonomous driving, this limitation causes difficulty in reasoning backward from desired outcomes (e.g., ``To avoid the congested traffic flow, I should exit the highway at the next exit'') or integrating future planning constraints into current decisions (e.g., ``Stop at the next intersection'').

To address these inherent limitations, we propose a paradigm shift in end-to-end autonomous driving from autoregressive VLMs to the Large Vision Language Diffusion (LVLD) models \cite{you2025lladavlargelanguagediffusion}. 
These models rely on a masked diffusion generation paradigm. Specifically, the text is generated by starting with a fully masked sequence, which is then iteratively filled in with tokens predicted by a model trained to reverse this masking process. 
Therefore, it offers several advantages in end-to-end autonomous driving:
\begin{itemize}
    \item \textbf{Efficient Parallel Generation}. Unlike sequential token prediction, diffusion models generate the decision sequences simultaneously, reducing inference latency.
    \item  \textbf{Bidirectional Reasoning}. The iterative bidirectional generation process enables models to consider both past context and future influences comprehensively.
    \item \textbf{Easy-First Generation Patterns}. The model can adaptively focus on easier aspects of the driving task first, then progressively address more complex aspects.
\end{itemize}

With the adoption of LVLD models, we introduce \textbf{ViLaD} (\textbf{Vi}sion \textbf{La}nguage \textbf{D}iffusion), a novel end-to-end autonomous driving framework that leverages the masked diffusion generation method. This technique trains a neural network model to reconstruct a complete sequence of driving actions from a version where parts of the output are deliberately ``masked.'' During inference, the model begins with a fully masked (except fixed tokens) sequence of decision tokens and iteratively ``unmasks'' the masked tokens in parallel over several steps to produce the final output. It demonstrates a successful end-to-end implementation of the LVLD autonomous driving system from the benchmark comprehensive experiment to a real-world vehicle deployment, which is the \textbf{first of its kind} to the best of our knowledge. 

\subsection{ViLaD: Vision Language Diffusion}

We present ViLaD (shown in Fig.~\ref{fig:vilad_main}), a novel architecture that leverages masked diffusion models for end-to-end autonomous driving. Our architecture consists of four key components: a vision encoder for camera-based perception, a multilayer perceptron (MLP) connector for multimodal alignment, and a diffusion-based language backbone for decision generation.

\paragraph{Overall Architecture Design} As shown in Fig.~\ref{fig:vilad_main}, ViLaD follows an encoder-decoder architecture where visual inputs are first processed through a vision encoder, then projected into the language embedding space through an MLP connector, concatenated with textual driving context, and finally processed by a transformer-based mask predictor to predict masks. Unlike traditional autoregressive VLMs that generate tokens sequentially, ViLaD generates complete driving action sequences through parallel diffusion-based generation, enabling real-time performance while maintaining bidirectional reasoning capabilities.

The architecture takes as input multimodal driving data: camera images $V$ and textual driving context $C$ (navigation instructions, traffic rules). The output consists of structured driving decisions $y$ representing waypoints, actions, and control commands. Mathematically, our model learns the conditional distribution $p_\theta(y|V, C)$ through the diffusion framework.

\paragraph{Vision Encoder and MLP Connector}
The vision encoder processes camera inputs to extract spatial-temporal features, which are then projected into the language model's embedding space for multimodal processing. The SigLP-2~\cite{tschannen2025siglip2multilingualvisionlanguage} is used as the base vision encoder to process input images $V$, producing visual tokens $\hat{V}$, where $N_v$ represents the number of visual tokens and $d_{model}$ is the hidden dimension. Then, the visual features $\hat{V}$ are projected into the language model's embedding space through a two-layer MLP connector, and the projected visual tokens $\hat{V}_{project} = MLP(\hat{V})$ are concatenated with textual driving context tokens $C_{emb}$ to form a unified multimodal input sequence:

\begin{equation}
    X_{input} = \text{Concat}([\hat{V}_{project}, C_{emb}]),
\end{equation}
where $C_{emb}$ represents the embedded textual driving context.

\paragraph{Diffusion Language Backbone}
The core of ViLaD is a masked diffusion transformer that processes the concatenated multimodal tokens through bidirectional self-attention, enabling joint reasoning over visual and textual information. 

\textbf{Fix Pattern Training.} The backbone employs a bidirectional transformer without causal masking, processing the concatenated multimodal sequence through transformer layers. The bidirectional self-attention enables the model to consider the complete context from both visual and textual modalities when predicting individual driving decisions.

During training, we utilize the masked diffusion approach specifically adapted for multimodal conditional generation and end-to-end autonomous driving tasks. 
Specifically, our input is a multimodal input $X_{input}$ consisting of projected visual features $\hat{V}_{project}$ and driving context embedding $C_{emb}$.
The target output $y_0$ is a fixed-pattern driving sequence that includes both control actions and punctuation marks, where punctuation marks always appear in predefined positions and the length of the $y_0$ is always consistent. This design ensures the model focuses solely on predicting action values rather than unnecessary formatting. We construct the model input as $X_t = \text{Concat}([X_{input}, y_0])$, applying masks only to the action tokens in $y_0$ while keeping $X_{input}$ and the punctuation tokens visible throughout training.

\begin{table*}[!t]
    \centering
    \small
    \caption{End-to-end planning performance on the nuScenes dataset (* results from ~\cite{xing2025openemma}).}
    \resizebox{0.9\linewidth}{!}{
    \begin{tabular}{l|c|llllll}
        \toprule
        Method & Learning & L2 (m) 1 s ($\downarrow$) & L2 (m) 2 s ($\downarrow$) & L2 (m) 3 s ($\downarrow$) & L2 (m) Avg ($\downarrow$) & Failure Rate (\%) ($\downarrow$)\\
        \midrule
        LLaVA-1.6-7B* & \multirow{4}{3em}{Zero-Shot} 
        & 1.66 & 3.54 & 4.54 & 3.24 & 4.06 \\
        Llama-3.2-11B* & 
        & 1.50  & 3.44 & 4.04 & 3.00  & 23.92 \\
        Qwen2-VL-7B* & 
        & 1.22 & 2.94 & 3.21 & 2.46  & 24.00\\
        \rowcolor{black!10} ViLaD-Zero & 
        & 1.09 & 2.51 & 3.74 & 2.45 & 0.03  \\
        \hline
        LLaVA-1.6-7B* & \multirow{4}{3em}{Open-EMMA} 
        & 1.49 & 3.38 & 4.09 & 2.98 & 6.12  \\
        Llama-3.2-11B* & 
        & 1.54 &3.31&3.91 & 2.92 & 22.00  \\
        Qwen2-VL-7B* & 
        & 1.45 &3.21 &3.76 &2.81 & 16.11 \\
        \rowcolor{black!10} ViLaD-CoT & 
        & 1.24 & 2.74 & 4.13 & 2.71  & 0.17 \\
        \hline
        LLaVA-1.6-7B & \multirow{4}{3em}{SFT} 
        & 0.91 & 2.50 & 3.44 & 2.28 & 55.25 \\
        Llama-3.2-11B & 
        & 0.80 & 2.31 & 3.10 & 2.07 &0.06  \\
        Qwen2-VL-7B & 
        & 1.32 & 2.94 & 3.98 & 2.74 & 0.03 \\
        \rowcolor{black!10} ViLaD-SFT & 
        & \textbf{0.79} & \textbf{1.92} & 2.83 & 1.85 & \textbf{0.00}  \\
        \hline
        \rowcolor{black!10} ViLaD-Opt & Efficient Optimized SFT
        & 0.81 & 1.93 & \textbf{2.69} & \textbf{1.81}  & \textbf{0.00} \\
        \bottomrule
    \end{tabular}
    }
    \label{tab:main-result}
\end{table*}

In each training step, for a training sample $X_{t} = \text{Concat}([X_{input}, y_0])$, we sample a noise level $t  \sim U(0,1)$ and independently mask each token in the target action tokens with probability t:
\begin{equation}
    q(y_t|y_0) = \prod_{i=1}^{L} q(y_t^i|y_0^i),
\end{equation}
where $L$ is the length of the driving sequence $y_0$. The projected visual features $\hat{V}_{project}$, textual context $C$, and punctuation marks remain unmasked throughout the process.

The training objective is:

\begin{equation}
    \mathcal{L}(\theta) = -\frac{1}{t} \sum_{i=1}^{L} \mathbb{1}[y_t^i = \text{[M]}] \log p_{\theta}(y_0^i | X_t),
\end{equation}
where $\mathbb{1}[y_t^i = \text{[M]}]$ ensures the loss is computed only on masked driving decision tokens, and $X_t$ represents the full concatenated sequence with masked driving responses.

\algrenewcommand\algorithmicrequire{\textbf{Input:}}
\algrenewcommand\algorithmicensure{\textbf{Output:}}

\begin{algorithm}[htbp]
\small
\caption{Remasking Process}\label{alg:remask}
\begin{algorithmic}[1]

    \Require Current masked sequence $y_t$, predictions $\hat{y}_0$, total number of tokens $L$, total steps $Sps$, threshold $\tau$
    \Ensure Next sequence state $y_s$

    
    \State $\textit{confidence}_{i} \gets p_{\theta}(\hat{y}_{0}^{i} \mid X_{t})$ for $i = 1, \dots, L$
    \State $\text{masked\_indices} \gets \{i \mid y_{t_i} = \text{[M]}\}$
   
    \State $n_{\text{unmask\_schedule}} \gets \lceil \frac{L}{Sps} \rceil$
    \Comment{Number of tokens to unmask based on schedule}
    
    \State $n_{\text{unmask\_threshold}} \gets |\{i \in \text{masked\_indices} \mid \textit{confidence}_i > \tau \}|$
    \Comment{Number of tokens to unmask based on confidence}
    
    \State $n_{\text{unmask}} \gets \max(n_{\text{unmask\_schedule}}, n_{\text{unmask\_threshold}})$
    
    \State $\text{conf}_{\text{masked}} \gets \{i \in \text{masked\_indices}| \textit{confidence}_i \in \text{topk}(\textit{confidence}, n_\text{unmask})\}$ \Comment{Get top-n confident tokens}
    \For{$i \in \text{conf}_{\text{masked}}$}
        \State $y_{s}^{i} \gets \hat{y}_{0}^{i}$ \Comment{Unmask the selected predictions}
    \EndFor
\end{algorithmic}
\end{algorithm}

\textbf{Fix Pattern Confidence-based Inference.}
During inference, we perform the reverse diffusion process starting from a partially masked sequence $y_1$, where all target action tokens are masked while punctuation tokens and length remain fixed, consistent with the training setup. The model then iteratively refines the masked tokens over up to $Sps$ denoising steps to generate the final driving action sequence.

One primary contribution we achieve is focusing on addressing quality degradation in parallel token generation caused by the conditional independence assumption in diffusion LLMs. When unmasking multiple token positions $i$ and $j$, MDMs sample these from $p_\theta(y^i_0|y_t) \cdot p_\theta(y^j_0|y_t)$ due to the conditional independence assumption, while the true joint probability is $p_\theta(y^i_0, y^j_0|y_t) = p_\theta(y^i_0|y_t) \cdot p_\theta(y^j_0|y_t, y^i_0)$. This discrepancy degrades generation quality. To address this, enlightened by Fast-dLLM~\cite{wu2025fastdllmtrainingfreeaccelerationdiffusion}, we propose confidence-aware parallel decoding: at each iteration, we compute confidence scores for each token and only unmask those exceeding the threshold $\tau$. If the number of tokens with confidence scores exceeding $\tau$ is less than $\lceil L/Sps \rceil $, we unmask the $\lceil L/Sps \rceil $ highest-confidence tokens to ensure progress.

The inference steps begin with sampling. At a step with the current sequence $X_t$, we use the model to predict all masked tokens simultaneously:

\begin{equation}
    \hat{y}_{0}^i = \underset{y_{0}^i}{\arg\max} \, p_{\theta}(y_{0}^i \mid X_{t}) \quad \text{for all } i.
\end{equation}

Then, following the threshold-based remasking strategy, we keep at least $\lceil L/Sps \rceil$ high-confidence predictions unchanged, and remask all the other tokens. The remask process is summarized in Algorithm \ref{alg:remask}.
After remasking, the resulting sequence is used as the input sequence in the next step. 
This iterative process lasts for at most $Sps$ steps until all tokens are masked, following the standard discretized reverse diffusion schedule. Both the number of sampling steps $Sps$ and the threshold $\tau$ can be adjusted to balance between generation quality and inference speed. The whole inference process is summarized in Algorithm \ref{alg:vilad_compact_modern}.

\begin{algorithm}[htbp]
\small
\caption{ViLaD Inference}\label{alg:vilad_compact_modern}
\begin{algorithmic}[1]
    \Require Visual features $\hat{V}$, driving context embedding $C_{emb}$, driving sequence length $L$, steps $Sps$
    \Ensure Predicted driving sequence $y_0$
    \State Initialize: $X_1 \leftarrow \text{Concat}([\text{MLP}(\hat{V}), C_{emb}, [\text{[M]}]^L])$
    \While{$y_s$ contains [M]} \Comment{Loop until all tokens are unmasked}
        \State $\hat{y}_0 \leftarrow \arg\max_{y_0} p_\theta(y_0 | X_t)$ \Comment{Parallel prediction}
        \State $y_s \leftarrow \text{Remask}(y_t, \hat{y}_0, p_\theta(\hat{y}_0^i | X_t))$ \Comment{Remask Process}
        \State $X_t \leftarrow \text{Concat}([V_{project}, C_{emb}, y_s])$
    \EndWhile
    \State \Return $y_0$
\end{algorithmic}
\end{algorithm}

\subsection{Experiment and Results}
\subsubsection{Experiment Setup}

We conduct experiments on the nuScenes dataset~\cite{caesar_nuscenes_2020} using only front camera images from the last frame and ground-truth trajectories for training and validation. Evaluation focuses on L2 error in the first 3 seconds and inference latency for real-time performance. Following Open-EMMA~\cite{xing2025openemma}, a prediction is considered a failure if the L2 distance exceeds 10 within the first second of the future trajectory. We compare ViLaD against three state-of-the-art VLMs: LLaVA-1.6-Mistral-7B~\cite{li_llava-med_2023}, Llama-3.2-11B-Vision-Instruct~\cite{zhang_video-llama_2023}, and Qwen2-VL-7B-Instruct~\cite{wang2024qwen2vlenhancingvisionlanguagemodels} under zero-shot, OpenEMMA prompting~\cite{xing2025openemma}, and supervised fine-tuning (SFT) settings. 

We evaluate multiple configurations of our proposed ViLaD framework to analyze different aspects of performance and capabilities. We utilize a pretrained LLaDA-V~\cite{you2025lladavlargelanguagediffusion} model as our base model, and fine-tune from it. Other ViLaD variants include ViLaD-SFT (full model with diffusion-based SFT), ViLaD-Zero (zero-shot), ViLaD-CoT (chain-of-thought reasoning using OpenEMMA~\cite{xing2025openemma}), and ViLaD-Opt (optimized inference), offering a broad evaluation of performance and capabilities.

\subsubsection{Optimization Strategy}
The outcome from dLLM-Cache~\cite{liu2025dllmcacheacceleratingdiffusionlarge} is used in our work. dLLM-Cache~\cite{liu2025dllmcacheacceleratingdiffusionlarge} introduces a training-free adaptive caching framework that combines long-interval prompt caching with partial response updates guided by feature similarity. These methods accelerate the inference speed of ViLaD and hence reduce the latency of the whole framework.

\begin{figure}[!t]
    \centering
        \includegraphics[width=0.85\linewidth]{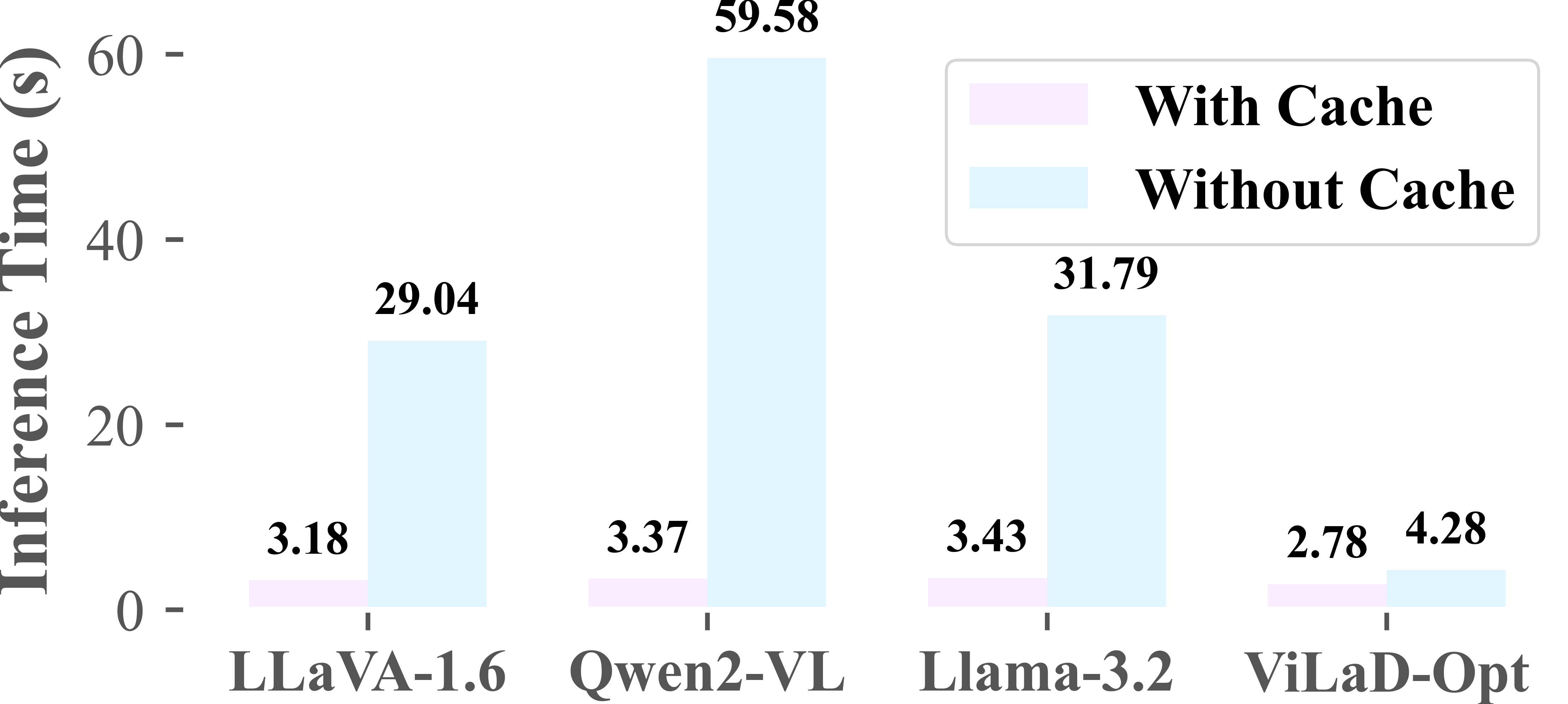}
    \vspace{-2mm}
    \caption{The comparison of inference time between VLMs and ViLaD-Opt.}
    \label{fig:te}
    \vspace{-7mm}
\end{figure}

\begin{table*}[t]
\caption{On-Board Model Performance Metrics for Interactive Parking.}
\centering
\resizebox{\textwidth}{!}{%
\begin{tabular}{c|cc|cc|cc}
\toprule
\multirow{2}{*}{Scenario} & \multicolumn{2}{c|}{Model Performance} & \multicolumn{2}{c|}{Safety Metrics} & \multicolumn{2}{c}{Comfort Metrics} \\
\cline{2-7}

& \begin{tabular}[c]{@{}c@{}}Latency (s)\end{tabular} & \begin{tabular}[c]{@{}c@{}}Success Rate (\%)\end{tabular} & \begin{tabular}[c]{@{}c@{}}Long. Vel. Var.( m²/s²)\end{tabular} & \begin{tabular}[c]{@{}c@{}}Lat. Vel. Var.($10^{-4}$ m²/s²)\end{tabular} & \begin{tabular}[c]{@{}c@{}}Max Long. Accel.(m/s²)\end{tabular} & \begin{tabular}[c]{@{}c@{}}Max Lat. Accel.(m/s²)\end{tabular} \\
\midrule
\textit{Parking Selection} & 0.18 & 95.18 & 1.07 & 2.66 & 1.43 & 1.12 \\
\textit{E2E Instruction Following} & 0.32 & 85.91 & 0.62 & 1.53 & 1.39 & 0.68 \\
\bottomrule
\end{tabular}%
}
\label{tab:onboard_performance}
\vspace{-6mm}
\end{table*}

\subsubsection{End-to-End Planning}
Tab.~\ref{tab:main-result} presents comprehensive comparison results across different methods. Our ViLaD framework demonstrates the best performance across all evaluation metrics compared to baseline VLM approaches. In the zero-shot setting, ViLaD-Zero achieves remarkable performance with L2 errors of 1.09 m, 2.51 m, and 3.74 m at 1 s, 2 s, and 3 s horizons, respectively, significantly outperforming the best baseline (Qwen2-VL-7B~\cite{wang2024qwen2vlenhancingvisionlanguagemodels}), which achieves 1.22 m, 2.94 m, and 3.21 m. Most notably, ViLaD-Zero maintains an extremely low failure rate of 0.03\% compared to baseline failure rates ranging from 4.06\% to 24.00\%, demonstrating the inherent robustness of our diffusion-based approach.

The OpenEMMA prompting framework shows consistent advantages of our approach. ViLaD-CoT achieves competitive performance with an average L2 error of 2.71 m while maintaining a near-zero failure rate (0.17\%), outperforming all baseline methods in this category. However, we found that the ViLaD model shows no improvement using the chain-of-thought prompting method. The chain-of-thought prompting method still needed to be explored.

With only the SFT setting (no fixed pattern training and fixed pattern confidence threshold inference), ViLaD-SFT achieves the best overall performance with L2 errors of 0.79 m, 1.92 m, and 2.83 m across the three time horizons, resulting in an average L2 error of 1.85 m with zero failure rate. This represents significant improvements over the best baseline (Llama-3.2-11B~\cite{zhang_video-llama_2023} SFT), which achieves 2.07 m average L2 error. Notably, while some baseline methods achieve competitive L2 errors, they suffer from substantially higher failure rates (up to 55.25\% for LLaVA-1.6-7B~\cite{li_llava-med_2023} SFT), highlighting the safety advantages of our approach.

ViLaD-Opt, our optimized deployment configuration with fixed pattern training and fixed pattern confidence threshold (0.5) inference, maintains the most competitive performance (1.81 m average L2 error) while achieving inference speedup, making it suitable for autonomous driving applications.


\subsubsection{Inference Efficiency Experiment}
To validate the real-time performance of our framework, we conducted a comprehensive inference efficiency experiment, comparing our approach against leading autoregressive VLM baselines that benefit from highly mature optimization technologies. The VLM baselines (LLaVA-1.6-7B~\cite{li_llava-med_2023}, Qwen2-VL-7B~\cite{wang2024qwen2vlenhancingvisionlanguagemodels}, and Llama-3.2-11B~\cite{zhang2023videollamainstructiontunedaudiovisuallanguage}) are accelerated using standard KV caching, while our ViLaD-Opt model, representing a still-developing class of vision language diffusion models, utilizes a dLLM cache. The results from the experiment, conducted on a single NVIDIA A100 GPU and presented in Fig. \ref{fig:te}, show the significant computational advantages of our diffusion-based method. Without caching optimizations, ViLaD-Opt achieves an inference time of 4.28 s, proving to be 6.5 to 14 times faster than the autoregressive models, which require between 28.10 s and 60.18 s. While caching accelerates all models, ViLaD-Opt maintains its superiority with an inference time of 2.78 s, still outperforming the fastest baseline (3.18 s). This experiment empirically confirms that the parallel generation paradigm of ViLaD provides a critical advantage in inference speed by avoiding the token-by-token processing bottleneck in autoregressive architectures, making it highly suitable for deployment in safety-critical, real-time applications.

\subsection{Real-World Case Studies: On-Board Interactive Parking}


We further validate the real-world applicability of our approach through two on-board case studies (interactive parking and instruction-following driving). Both of them deployed on the same vehicle platform as shown in Fig.~\ref{fig:Vehicle_Setup}.

\subsubsection{System Setup}



We evaluate system efficiency (inference latency) and driving quality (success rate, safety, and comfort). A trial is successful if the executed driving decision matches the instruction. Safety is measured by the variance of longitudinal and lateral velocities, and comfort by maximum longitudinal/lateral accelerations.

Two lightweight models fine-tuned from SMDM~\cite{nie2025scalingmaskeddiffusionmodels}, which are lightweight and support more on-board inference compared with the larger-scale models used in the full framework above, are used for the respective tasks. The parking dataset contains $\sim$3,000 command–action pairs covering diverse slot preferences, and the instruction-following dataset has a similar scale focused on short-horizon commands. Each task is tested with 300 unseen instructions. Demo videos are publicly available.\footnote{Real-vehicle experiment demo video with ViLaD: \url{https://youtu.be/pNfC7wrENyk}}

\subsubsection{Case Study I Interactive Parking}



Our first case focuses on selecting a parking slot based on natural language commands (e.g., ``Find a spot near the exit''). The target driving action predicted by our method includes the final waypoint corresponding to the chosen slot. As shown in Tab.~\ref{tab:onboard_performance}, ViLaD achieves 0.18~s latency and a 95.18\% success rate, with smooth dynamics (longitudinal/lateral variances: \SI{1.07}{\meter\squared\per\second\squared} / \SI{2.66e-4}{\meter\squared\per\second\squared}) and imperceptible accelerations (\SI{1.43}{\meter\per\second\squared}, \SI{1.12}{\meter\per\second\squared})~\cite{de2023standards,xu2015experimental}. These results confirm precise spatial reasoning and safe maneuvering in constrained spaces.

\subsubsection{Case Study II End-to-End Instruction Following}



For open-road commands (e.g., ``Turn left and stop near the crosswalk''), the vehicle autonomously executes the instruction over 28–38~m. ViLaD reaches an 85.91\% success rate with 0.32~s latency, demonstrating real-time performance. Stable motion (velocity variances: \SI{0.62}{\meter\squared\per\second\squared} / \SI{1.53e-4}{\meter\squared\per\second\squared}) and low accelerations (\SI{1.39}{\meter\per\second\squared}, \SI{0.68}{\meter\per\second\squared}) show reliable control and comfort. These results highlight ViLaD’s ability to generalize from parking to dynamic, instruction-driven driving.}

\section{Challenges} 
\label{sec:discussion}
\subsection{Latency}

Latency presents a fundamental challenge in deploying LLMs for autonomous driving applications. While traditional autonomous driving systems operate with millisecond-level response times, the integration of LLMs introduces significant latency concerns that must be carefully addressed. Our experimental work with cloud-based LLM APIs revealed multiple layers of latency challenges that impact system reliability.

Network latency forms the first critical bottleneck. Our real-world experiments under stable 4G-LTE conditions showed a baseline communication latency of approximately 300 ms - already an order of magnitude higher than typical autonomous driving control loops. This network dependency introduces additional vulnerabilities, as real-world driving conditions often involve varying network quality, dead zones, or connection interruptions that could compromise system reliability.

The LLM processing itself introduces substantial additional latency. Our empirical results with text-based GPT-4~\cite{openai_gpt-4_2023} or on-board \first{8B} VLM showed end-to-end latencies ranging from 1.2 to 1.8 seconds, while the visual-capable GPT-4o~\cite{openai2024gpt4ocard} exhibited even higher latencies around 5 seconds (assuming 300ms network latency). These latencies pose significant challenges for real-time decision-making in autonomous driving scenarios. While such response times might be acceptable for high-level decision-making or handling non-critical user requests, they become problematic for any safety-critical functions.

The latency challenge is further complicated by the variable nature of LLM response times. Input complexity, prompt length, and required reasoning depth can all significantly impact processing time. This variability makes it difficult to guarantee consistent response times - a crucial requirement for safety-critical systems. Additionally, the system must manage the trade-off between prompt effectiveness and latency, as more detailed prompts that might lead to better responses also increase processing time.

These latency constraints limit the role LLMs can play in autonomous driving architectures. Critical path functions requiring real-time responses (collision avoidance, emergency braking, immediate obstacle response) may still remain handled by traditional, low-latency control systems. This creates a complex architectural challenge: how to effectively partition decision-making responsibilities between high-latency LLM components and low-latency control systems while maintaining system safety and reliability.

\subsection{Deployment}
Another important consideration is the deployment of on-board LLMs in vehicles. While cloud-based LLM APIs offer flexibility and scalability, they rely on stable network connections, which may not always be available in all driving scenarios. To mitigate this issue, researchers are exploring the possibility of integrating LLMs directly into the vehicle's on-board systems. However, the deployment of LLMs in autonomous vehicles faces significant computational resource limitations. Automotive hardware typically has lower computational capabilities compared to development systems, creating a substantial gap between model requirements and available resources~\cite{cui2024onboardvisionlanguagemodelspersonalized}. This constraint affects all aspects of system performance, from basic perception tasks to complex decision-making processes. The challenge is particularly obvious for LLMs, which require substantial GPU computing power that often exceeds typical automotive hardware capabilities. 

Additionally, researchers can focus on optimizing LLMs specifically for in-vehicle deployment. This involves streamlining the models to reduce computational requirements, enabling them to run efficiently on the limited hardware resources available in automotive systems. Model distillation, which transfers knowledge from large teacher networks, such as multimodal LLMs, to smaller student architectures, is a promising yet underexplored direction in the autonomous driving domain for addressing computational constraints~\cite{sreenivas2024llm}. Similarly, plug-and-play approaches offer another compelling avenue for integrating foundation model capabilities into on-device autonomous driving networks without adding additional computational complexity~\cite{ma2023lampilot}.

Researchers must find a balance between model complexity and hardware constraints to ensure practical deployment in vehicles. As developments in hardware and model optimization continue, the feasibility of integrating powerful on-board LLMs directly into vehicles will likely increase, opening up new possibilities for enhancing the capabilities of autonomous driving systems.

\subsection{Security and Privacy}

The integration of LLMs into autonomous driving systems introduces significant security and privacy vulnerabilities that require careful consideration. While traditional autonomous vehicles already face cybersecurity challenges, the addition of LLMs creates new attack surfaces due to their natural language processing capabilities and complex decision-making processes.

A fundamental security concern lies in adversarial attacks targeting the LLM's language understanding capabilities~\cite{goodfellow2014explaining,carlini2017towards}. Unlike traditional autonomous systems, where inputs are structured and validated, LLMs must process natural language commands that could contain malicious intent. For instance, carefully crafted commands could potentially trigger the model's semantic understanding to trigger unsafe driving behaviors or override safety protocols. These attacks are particularly challenging to defend against because they operate at the semantic level, where distinguishing between appropriate and malicious commands becomes highly complex.

The risk of data poisoning and backdoor attacks~\cite{biggio2012poisoning,liu2018trojaning} presents another critical challenge, particularly given the continuous learning and adaptation capabilities of LLM-based systems. Attackers could potentially inject malicious training examples that create hidden triggers in the model's language understanding. For example, specific phrases or conversation patterns could be embedded as triggers that cause the system to deviate from safe operation. The detection of such backdoors becomes increasingly difficult as the model's decision-making process grows more sophisticated through personalization and learning from user interactions.

Prompt injection attacks~\cite{bagdasaryan2023ab} represent a unique threat in the context of autonomous driving. Since LLMs process natural language inputs to generate driving decisions and responses, attackers could potentially embed harmful instructions within seemingly innocent conversations. For instance, a casual conversation about route preferences could contain hidden commands that manipulate the vehicle's behavior in subtle but dangerous ways. 

Privacy concerns extend beyond traditional vehicular data protection. LLM-based autonomous systems necessarily process and retain extensive personal data, including individual communication patterns and preferences, regular routes and destinations, context-specific behaviors and habits, voice characteristics and linguistic patterns, emergency response preferences, and medical information. This rich personal data, combined with the system's cloud-based architecture, creates significant privacy risks. Traditional privacy-preserving techniques like differential privacy~\cite{abadi2016deep} face new challenges in this context, as they must balance data protection with the need for personalized and context-aware responses.

\first{Our real-world implementations (Sec.~\ref{sec:onboard} and~\ref{sec:real-vehicle}) mitigate some of these risks by storing user-specific interaction logs (e.g., commands $I$, policies $P$, feedback $F$) on the vehicle's on-board ECU in a local database, rather than in the cloud. However, this does not solve the challenge of defining clear data-governance policies. Future systems must move beyond principles and implement concrete policies for: (1) Data Retention (e.g., ``voice recordings are deleted immediately after transcription," ``interaction logs are deleted after 90 days"), and (2) User Control (e.g., a ``privacy mode" that disables logging, and a clear interface for users to review, export, or delete their entire stored profile).}

\subsection{Safety}
When LLMs are integrated into autonomous driving systems, safety is always the first priority. In our proposed LLM4AD concept, the LLMs themselves do not directly handle safety requirements because of the potential for hallucinations in LLM outputs and the reaction time required. Hallucinations could be dangerous in safety-critical situations, and LLMs may not react quickly enough to sudden events. In our solutions, we employ a rule-based safety guarantee mechanism where LLM outputs must fall within predefined safe ranges if outputs exceed these bounds or do not match the output format, execution is automatically blocked.


However, the challenge is that ensuring system safety and reliability requires extensive validation across a wide range of operational conditions. This includes testing system responses to edge cases, verifying behavior in degraded operating conditions~\cite{he2010single,zhu2024mwformer,10310160,li2024light}, and validating the effectiveness of fallback mechanisms. \first{This technical challenge is intrinsically linked to legal and ethical questions. For instance, if an LLM misinterprets an ambiguous user command and contributes to an accident, it creates a complex chain of liability. While our proposed architecture mitigates this by using rule-based safety checks to override or block unsafe LLM outputs, it does not solve the fundamental ethical question of responsibility. Defining the legal and ethical frameworks for accountability in these human-AI teaming scenarios remains a significant, unresolved challenge that is critical for public trust and future deployment.} The challenge extends to maintaining clear audit trails of system decisions and ensuring alignment with different regulatory requirements. This is particularly critical for systems implementing LLM-based components, where behavior must consistently align with defined ethical guidelines while maintaining safety guarantees.

\subsection{Trust and Transparency}
A critical challenge in integrating LLMs into autonomous driving systems lies in establishing genuine trust and transparency between human users and the autonomous vehicle. While LLMs can generate natural language explanations for vehicle decisions~\cite{marcu2024lingoqavisualquestionanswering}, several fundamental challenges emerge.

First, there is a risk of explanation mismatch: The LLM may generate plausible-sounding explanations that do not actually match the true decision-making process of the underlying autonomous system. This could create a false sense of transparency and potentially dangerous misconceptions about the vehicle's capabilities and limitations. When the LLM generates both executable policies and natural language reasoning, we must ensure perfect alignment between the two to avoid misleading explanations. \first{In our proposed frameworks, the explanation (output thought $R$) is generated concurrently with the executable policy ($P$). This explanation is intended to be shown on the vehicle's human-machine interface (HMI) display during execution, allowing the human to understand the system's intent. However, detecting a mismatch between the plausibility of the explanation and the technical correctness of the policy remains an open challenge. Our current human-in-the-loop feedback $F$ provides a basic mechanism for users to flag such discrepancies, but an automated verification layer for this ``explanation-action alignment" is a critical area for future research.}

Apart from that, the real-time nature of autonomous driving poses significant challenges for LLM-based explanations. The system must balance between providing detailed, comprehensible explanations and maintaining rapid response times for critical driving decisions. Too much detail could overwhelm users in time-sensitive situations, while oversimplified explanations might omit crucial context.

Additionally, the black-box nature of LLMs themselves raises questions about the trustworthiness of their explanations. Users might question whether they can trust explanations generated by a system that operates in ways that are not fully transparent. This creates a meta-level transparency challenge that needs to be carefully addressed in the system design.

\subsection{Personalization}
Although one of the primary focuses of our research is integrating LLMs to achieve a higher level of personalization in autonomous vehicles by learning and adapting to individual drivers' habits, preferences, and communication styles over time,  it also introduces significant technical and safety challenges. 

A critical challenge lies in the interpretation of natural language commands and emotional states. While LLMs can process natural language, accurately inferring the driver's true intent and appropriate response is non-trivial. For instance, if a driver expresses urgency with ``I am on my way to the hospital,'' the system must carefully balance between accommodating the emergency situation and maintaining safe driving practices. Simply adopting a more aggressive driving style could introduce unnecessary risks, particularly in complex traffic conditions where the safest course of action might conflict with the inferred urgency.

The long-term learning and adaptation process also raises important questions about system boundaries and safety constraints. As the system builds a personalized profile over time, it must maintain clear limits on which aspects of driving behavior can be personalized and which must remain constrained by safety parameters. This creates a complex balance between personalization and standardized safety requirements, particularly in scenarios where learned preferences might conflict with optimal safety decisions.

\first{A further challenge is the scalability of the personalization memory and the resolution of conflicting user preferences over time. A simple approach of logging all historical interactions becomes computationally unmanageable and can lead to ambiguous or contradictory guidance for the model. However, our on-board VLM framework (Section~\ref{sec:onboard}) is designed to address this challenge using a RAG-based memory module. This architecture avoids feeding an unmanageably large history into the model's context. Instead, it leverages a vector database to perform a similarity search, retrieving only the top-k most relevant prior interactions based on the current command and context. This RAG-based approach directly addresses scalability by ensuring the VLM reasons over a small, contextualized subset of data, even from a history of thousands of interactions. This method also provides a mechanism for resolving conflicts; by retrieving the most similar past scenarios, the VLM can arbitrate between a few relevant (and potentially conflicting) examples to best match the user's preference in the current specific situation .}

\section{Conclusions}
\label{sec:conclusion}

\first{In this paper, we presented a comprehensive framework for integrating Large Language Models into autonomous driving, referred to as LLM4AD. We introduced the overall concept and architecture, reviewed the recent advances, proposed standardized benchmarks including LaMPilot-Bench and NuPlanQA to evaluate instruction-following and multimodal reasoning abilities, and conducted extensive experiments in both simulation and real-world environments. Our real-vehicle studies demonstrated the feasibility of deploying on-cloud and on-edge LLMs for personalized decision-making and motion control, highlighting their ability to interpret human intent and adapt to diverse driving contexts.

Looking forward, we explored the emerging Vision-Language Diffusion Models (ViLaD) as a promising future trend for LLM4AD, offering new directions for reasoning, planning, and generative scene understanding. Despite significant progress, challenges remain in achieving real-time inference, safety assurance, explainability, and personalization. We believe that future development in model efficiency, multimodal reasoning, and human-autonomy teaming will enable next-generation autonomous vehicles to be more adaptive, interpretable, and human-centric.}



%




\ifCLASSOPTIONcaptionsoff
  \newpage
\fi




\printbibliography
%


%


\begin{IEEEbiography}
[{\includegraphics[width=1in,height=1.25in,clip,keepaspectratio]{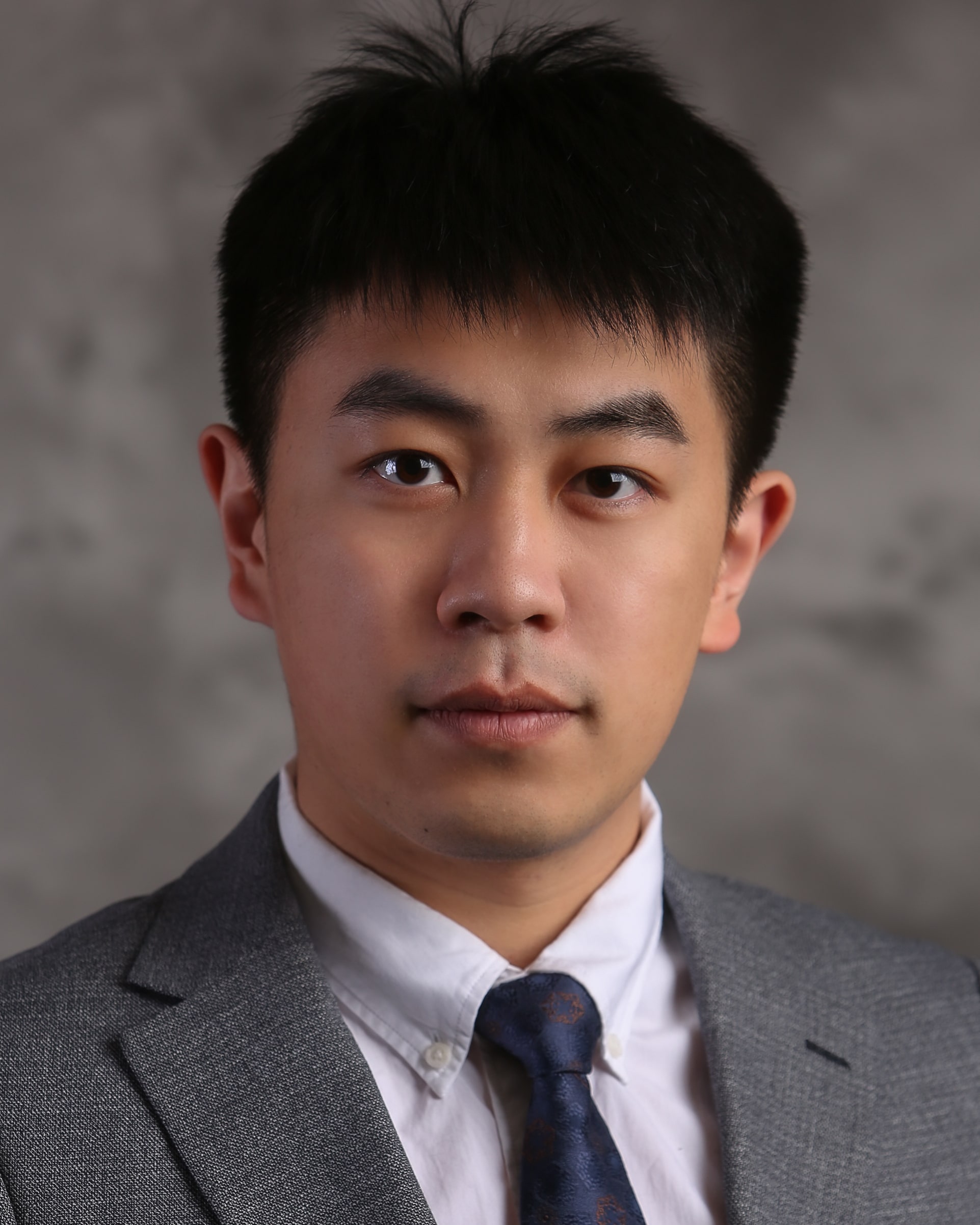}}]
{Can Cui}
(Student Member, IEEE) received his M.S. degree in Electrical and Computer Engineering from the University of Michigan, Ann Arbor in 2022. He is now pursuing a Ph.D. degree in the College of Engineering at Purdue University. His research pursuits encompass automated driving, Advanced Driver-Assistance Systems (ADAS), applied machine learning, controls, and digital twin.
\end{IEEEbiography}
\vspace{-10pt}

\begin{IEEEbiography}[{\includegraphics[width=1in,height=1.25in,clip,keepaspectratio]{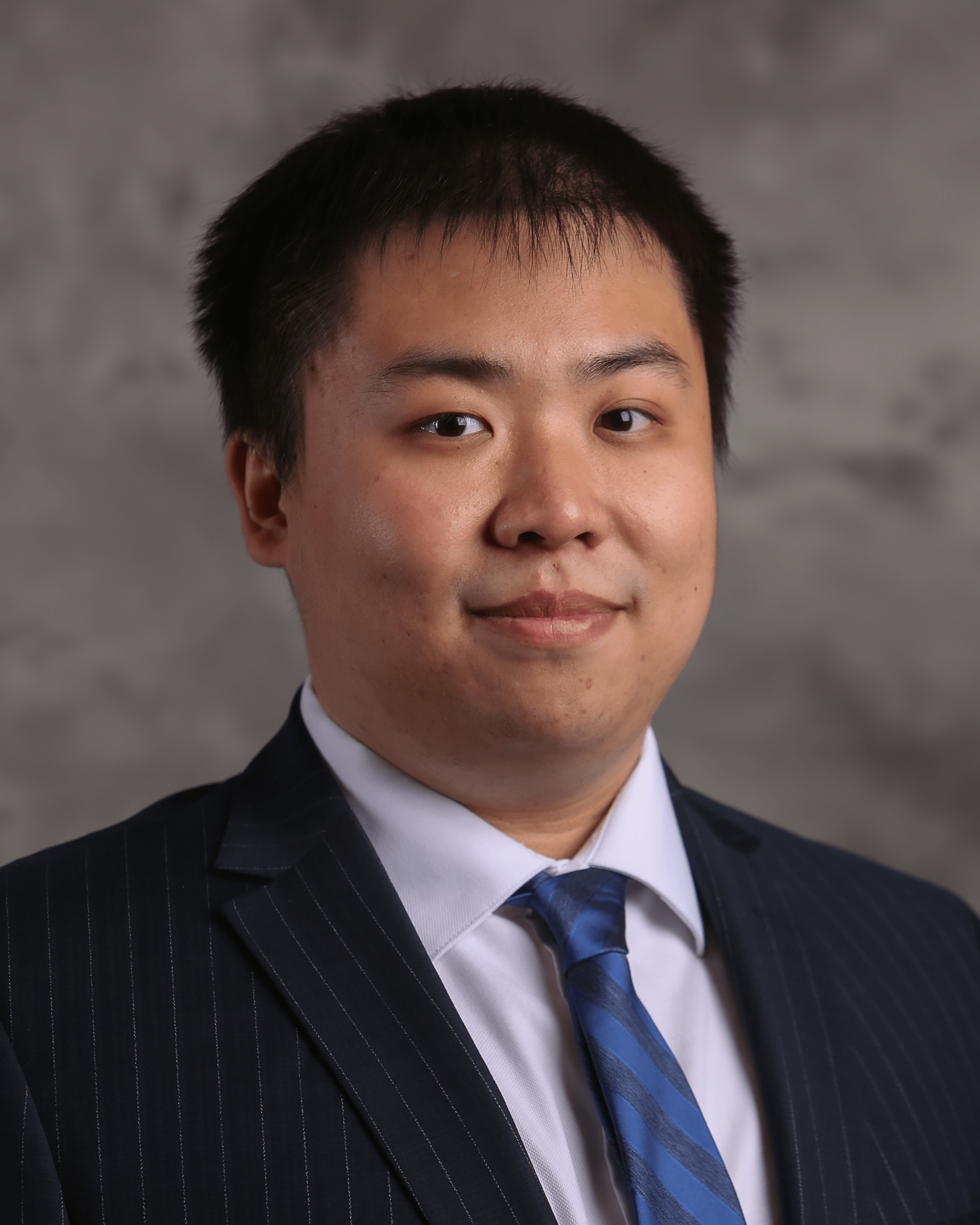}}]{Yunsheng Ma}
(Student Member, IEEE) received his M.S. degree in Computer Science from New York University in 2022. He is now pursuing a Ph.D. degree in the College of Engineering at Purdue University, with a research focus on autonomous driving, embodied AI, and digital twins. He has publications in top-tier conferences and journals, including CVPR, AAAI, ITSC, and IEEE Transactions on Intelligent Vehicles.
\end{IEEEbiography}
\vspace{-10pt}

\begin{IEEEbiography}[{\includegraphics[width=1in,height=1.25in,clip,keepaspectratio]{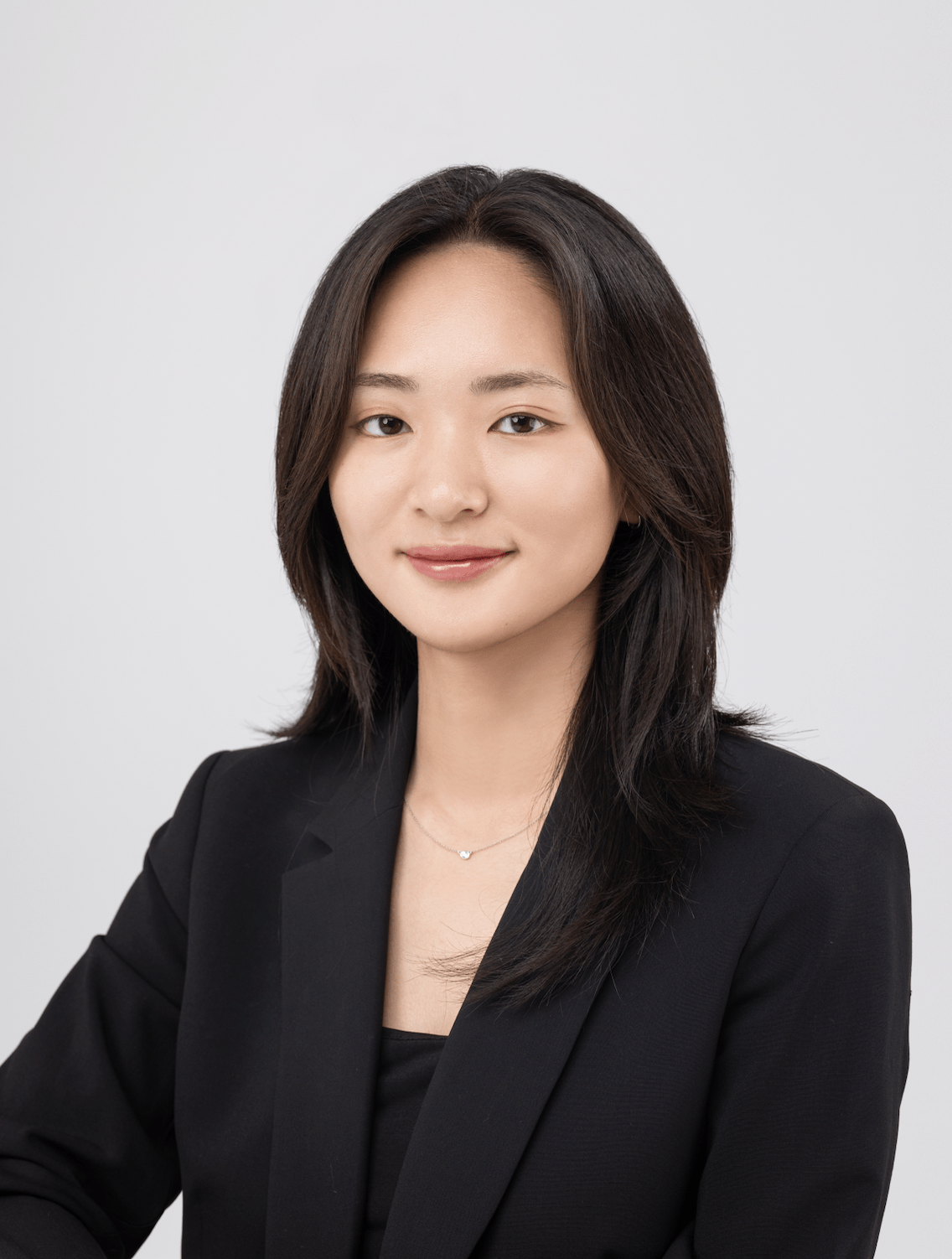}}]{Sung-Yeon Park} received her M.S. degree in Civil Engineering from Seoul National University in 2023. She is now pursuing a Ph.D. degree in the College of Engineering at Purdue University, with a research focus on autonomous driving for 3D/4D scene understanding and spatial reasoning.
\end{IEEEbiography}
\vspace{-10pt}

\begin{IEEEbiography}[{\includegraphics[width=1in,height=1.25in,clip,keepaspectratio]{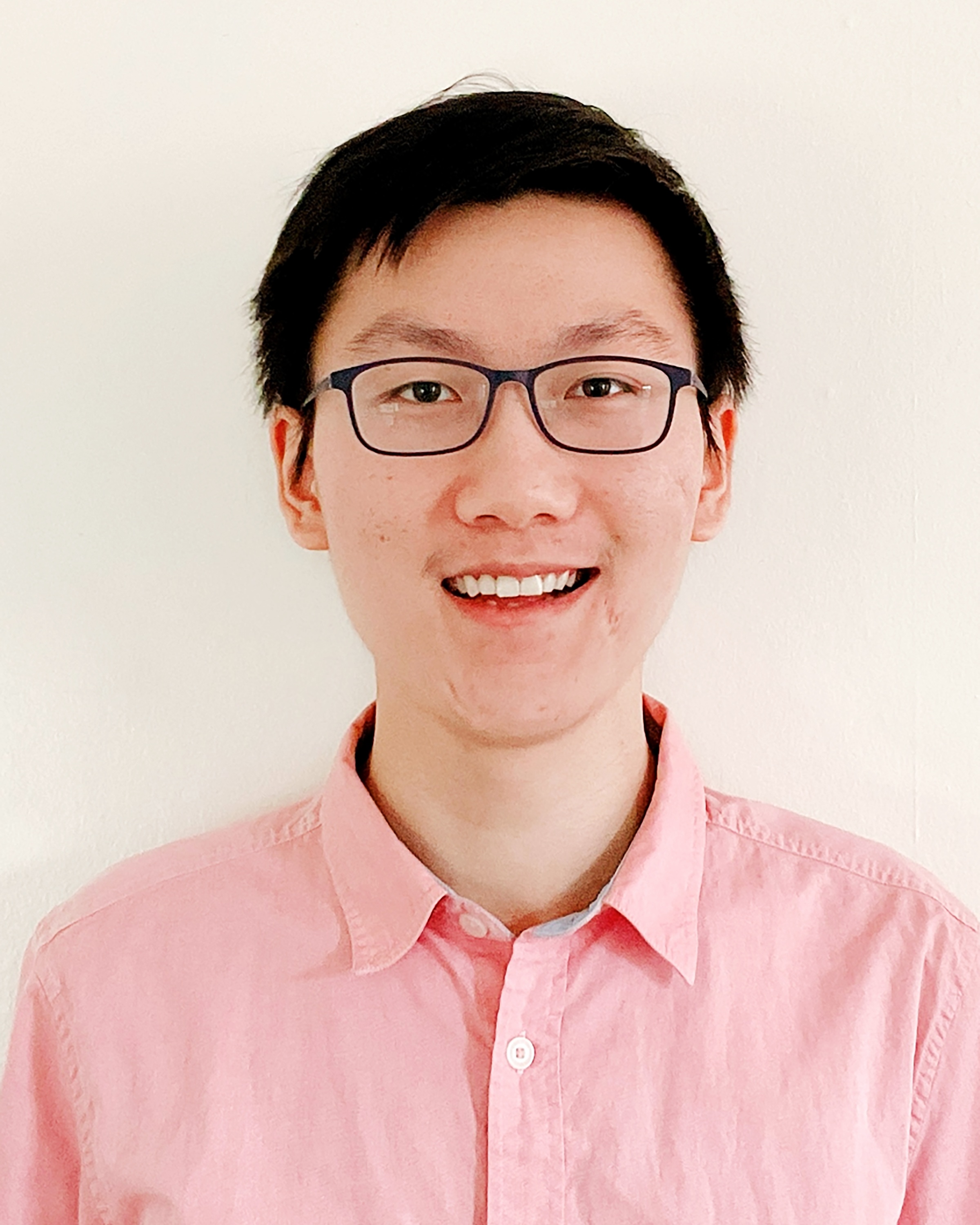}}]{Zichong Yang}
received his M.S. degree in Mechanical Engineering from Carnegie Mellon University in 2023. He is now pursuing a Ph.D. degree in the College of Engineering at Purdue University, with a research focus on computer vision, human pose estimation, and human trajectory prediction.
\end{IEEEbiography}
\vspace{-10pt}

\begin{IEEEbiography}[{\includegraphics[width=1in,height=1.25in,clip,keepaspectratio]{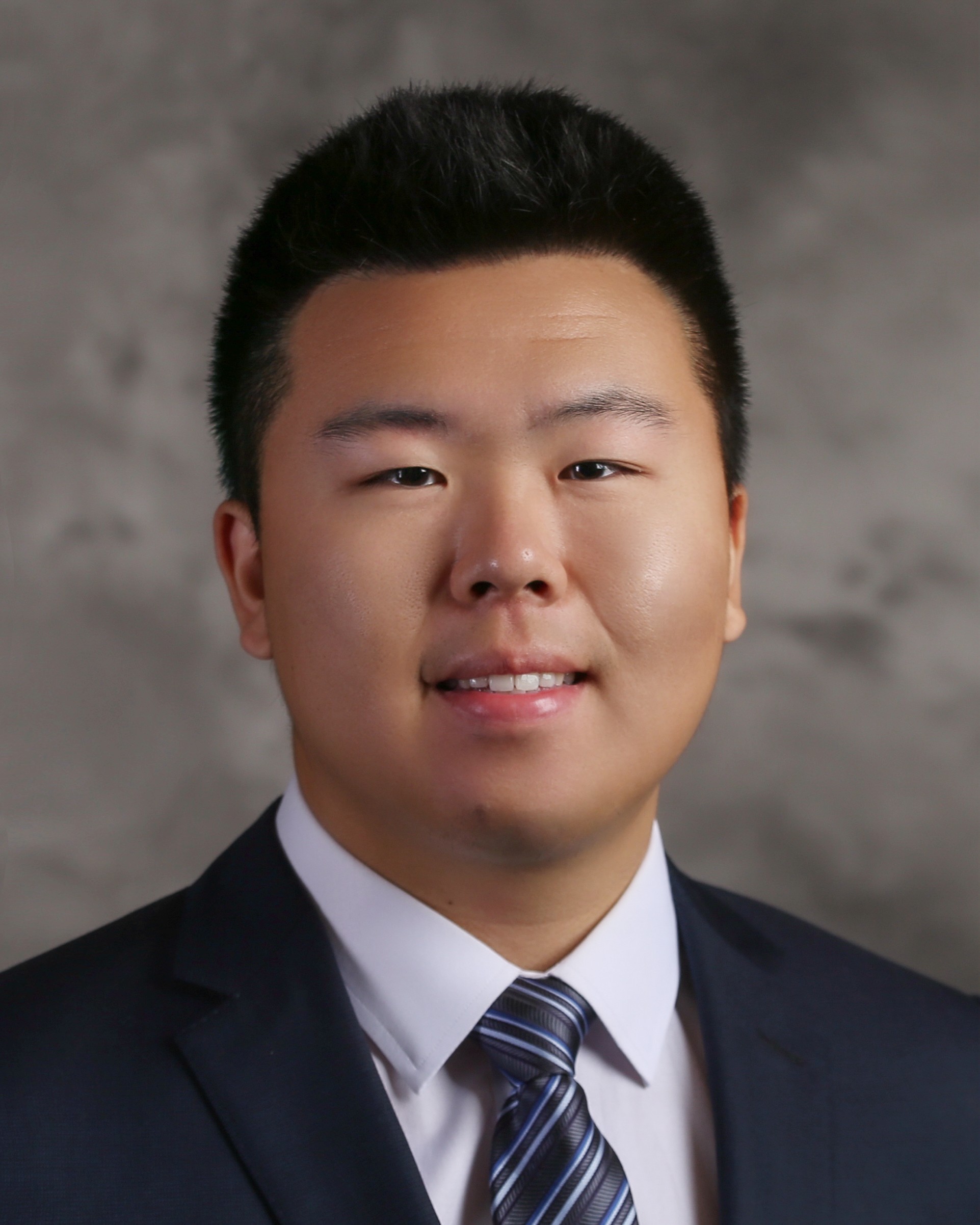}}]{Yupeng Zhou}
(Student Member, IEEE) received his M.S. degree in Mechanical Engineering from Purdue University, West Lafayette, in 2024. He is now a Ph.D. student in the College of Engineering at Purdue University, with a research focus on applied robotics, autonomous vehicles, and controls. 
\end{IEEEbiography}
\vspace{-10pt}

\begin{IEEEbiography}
[{\includegraphics[width=1in,height=1.25in,clip,keepaspectratio]{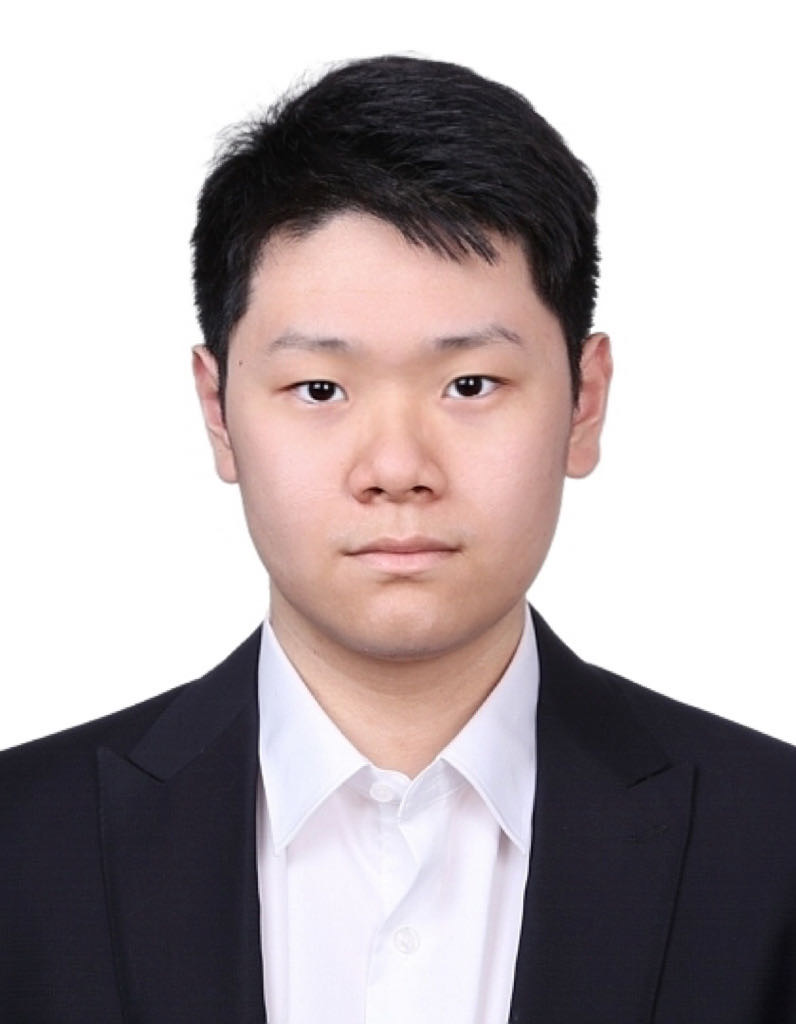}}]{Peiran Liu}
received a B.S. degree in Information Systems from the University of Florida in 2023. He is currently pursuing his Ph.D. degree in the College of Engineering at Purdue University. His research interests include computer vision, human behavior predictions, and digital twins.
\end{IEEEbiography}
\vspace{-10pt}

\begin{IEEEbiography}
[{\includegraphics[width=1in,height=1.25in,clip,keepaspectratio]{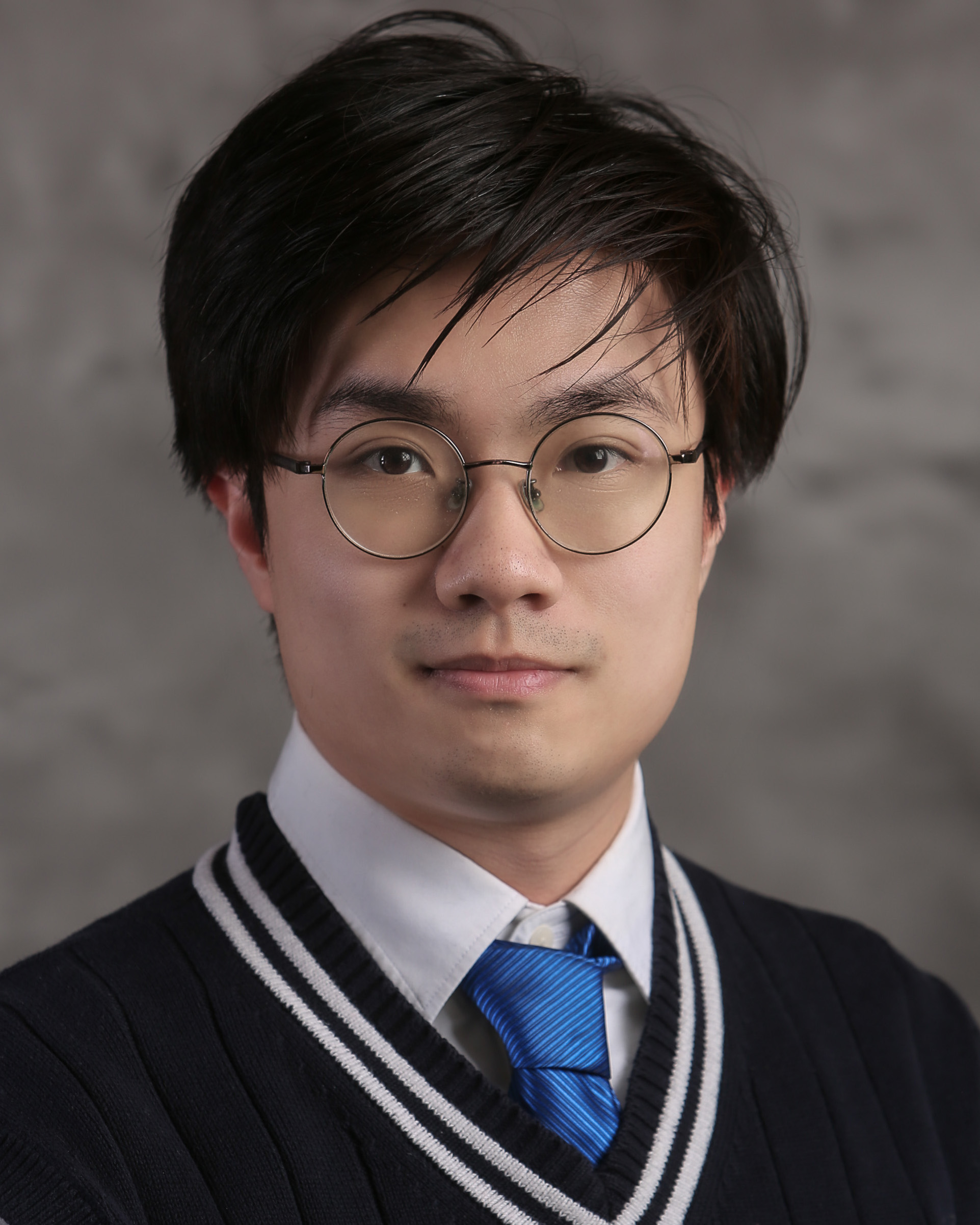}}]
{Juanwu Lu}
(Student Member, IEEE) received an M.S. degree in Civil Engineering from the University of California, Berkeley in 2022. He is currently pursuing his Ph.D. degree in the College of Engineering at Purdue University. His research interests include statistical machine learning and its application for generalizable and interpretable microscopic traffic modeling.
\end{IEEEbiography}
\vspace{-10pt}

\begin{IEEEbiography}
[{\includegraphics[width=1in,height=1.25in,clip,keepaspectratio]{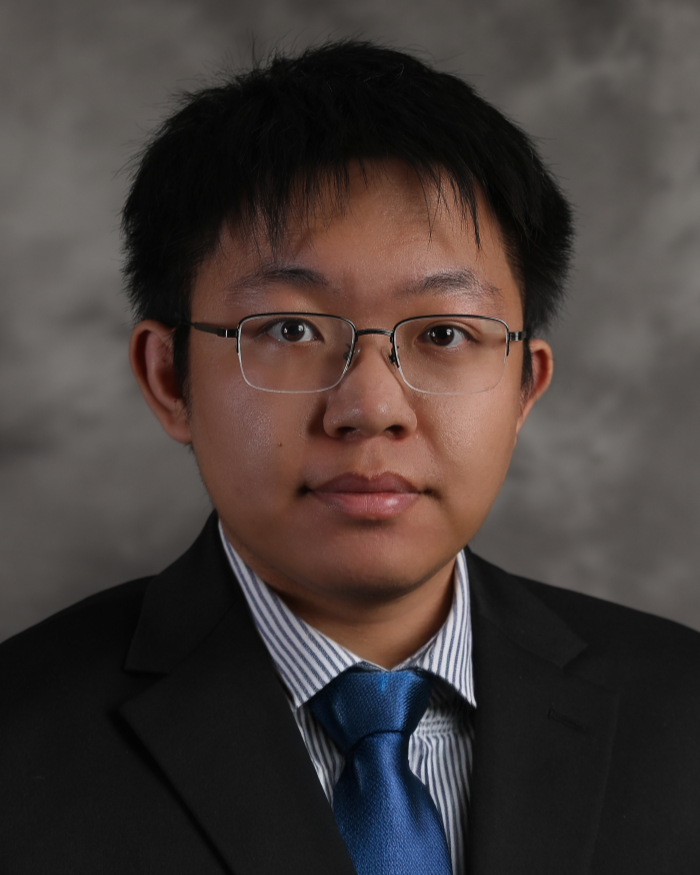}}]
{Juntong Peng}
(Student Member, IEEE) received his B.Eng. degree in Information Engineering from Shanghai Jiao Tong University in 2024. He is now pursuing his Ph.D. degree in the College of Engineering at Purdue University, with a research focus on autonomous driving, collaborative perception, and multi-agent intelligence.
\end{IEEEbiography}
\vspace{-10pt}

\begin{IEEEbiography}
[{\includegraphics[width=1in,height=1.25in,clip,keepaspectratio]{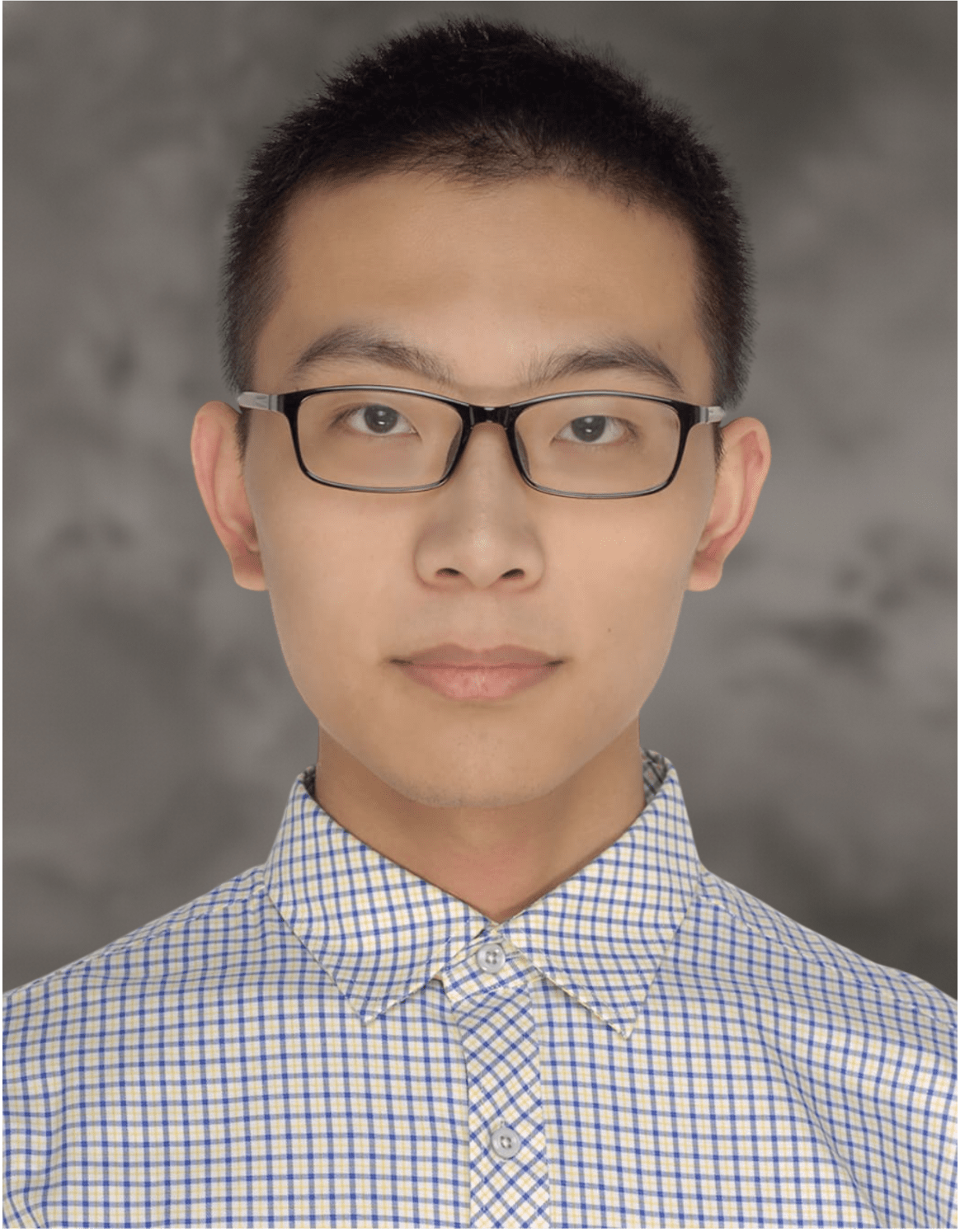}}]
{Jiaru Zhang} received his Ph.D. degree in Computer Science from Shanghai Jiao Tong University in 2024. He is currently a postdoctral scholar in the Institute for Physical Artificial Intelligence at Purdue University. His research interests include Bayesian deep learning, generative models, and their applications in autonomous driving.
\end{IEEEbiography}
\vspace{-10pt}

\begin{IEEEbiography}
[{\includegraphics[width=1in,height=1.25in,clip,keepaspectratio]{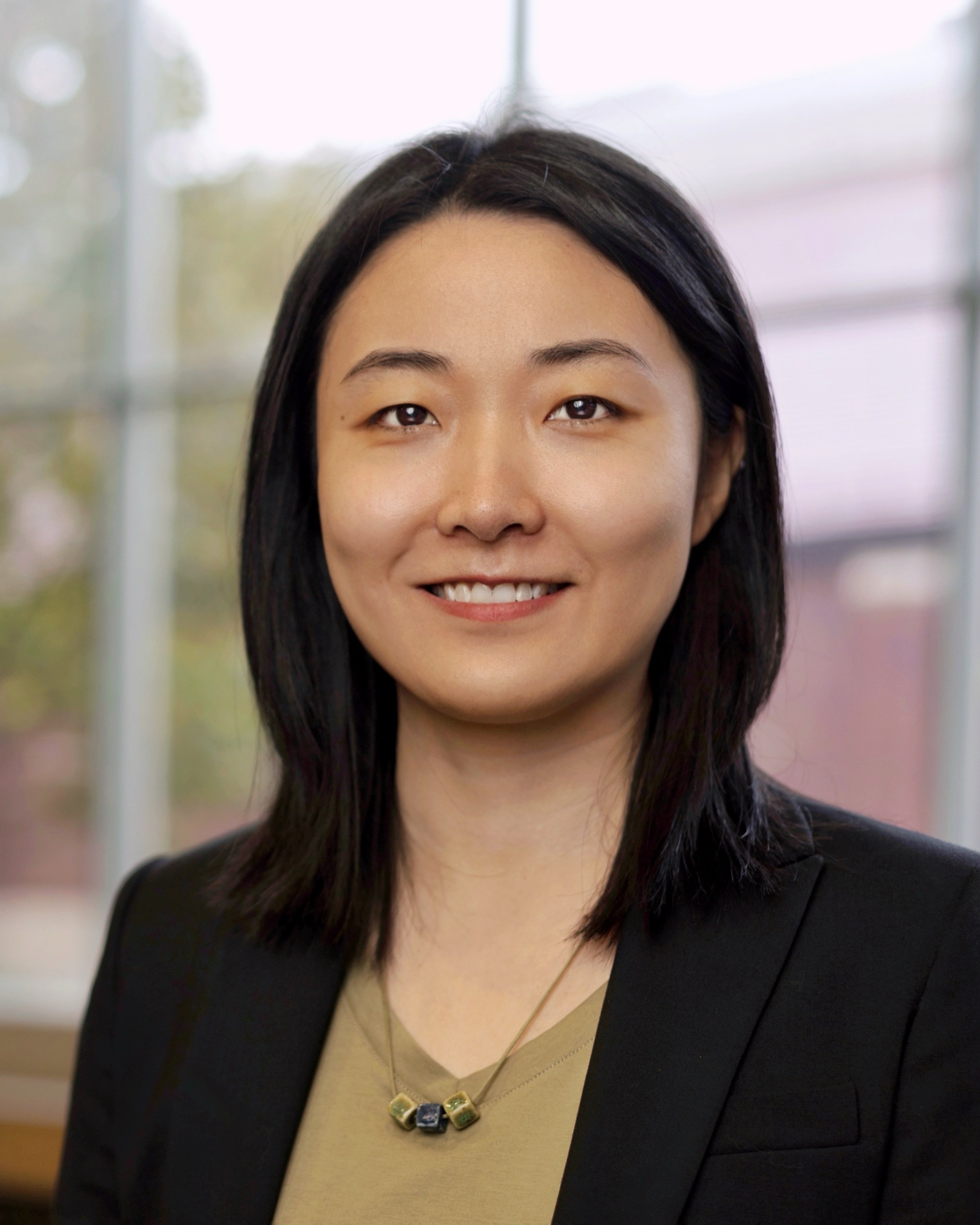}}]
{Ruqi Zhang} received her Ph.D. degree from Cornell University. She is currently an assistant professor in the Department of Computer Science at Purdue University. Before that, she was a postdoctoral fellow at the Institute for Foundations of Machine Learning, UT Austin. Her research focuses on machine learning, deep learning, generative modeling, probabilistic methods, and trustworthy AI.
\end{IEEEbiography}
\vspace{-10pt}

\begin{IEEEbiography}
[{\includegraphics[width=1in,height=1.25in,clip,keepaspectratio]{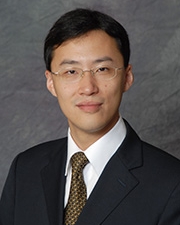}}]
{Lingxi Li}
(Senior Member, IEEE) received his Ph.D. degree in Electrical and Computer Engineering from the University of Illinois at Urbana–Champaign in 2008. He is a professor in Electrical and Computer Engineering at Purdue University. His research interests include complex systems, intelligent transportation systems, connected and automated vehicles, discrete event dynamic systems, human-machine interaction, digital twins, and parallel intelligence.
\end{IEEEbiography}
\vspace{-10pt}

\begin{IEEEbiography}
[{\includegraphics[width=1in,height=1.25in,clip,keepaspectratio]{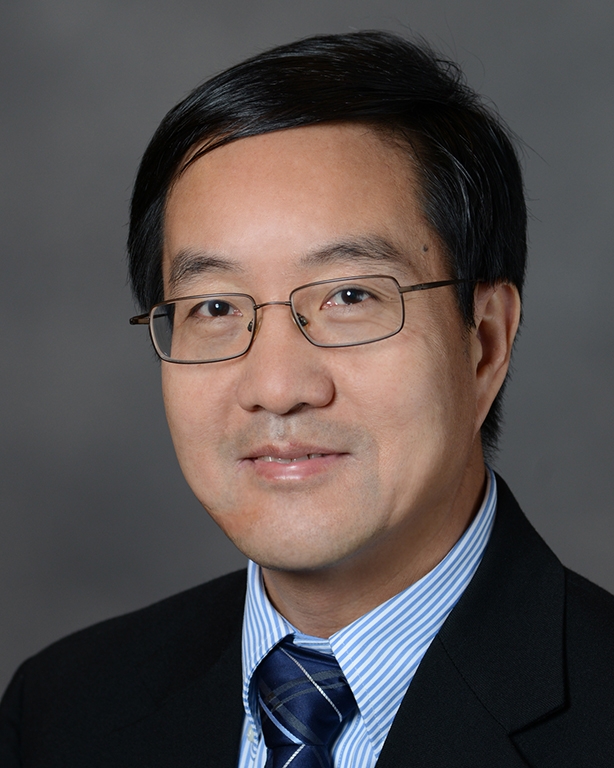}}]
{Yaobin Chen}
(Senior Member, IEEE) received his Ph.D. degree in Electrical and Computer Engineering from Rensselaer Polytechnic Institute in 1988. He is a professor in Electrical and Computer Engineering at Purdue University. His areas of expertise include automated driving and autonomous/connected vehicles, vehicle active safety systems, EV and HEV, intelligent controls and robotics, and advanced energy and power systems.
\end{IEEEbiography}
\vspace{-10pt}

\begin{IEEEbiography}
[{\includegraphics[width=1in,height=1.25in,clip,keepaspectratio]{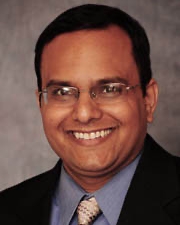}}]
{Jitesh Panchal}
received his Ph.D. degree in Mechanical Engineering from Georgia Institute of Technology in 2005. He is a professor in Mechanical Engineering at Purdue University. His areas of expertise include computational design of socio-technical systems, secure design and manufacturing, engineering design by self-organized virtual communities, and integrated products and materials design.
\end{IEEEbiography}
\vspace{-15pt}

\begin{IEEEbiography}
[{\includegraphics[width=1in,height=1.25in,clip,keepaspectratio]{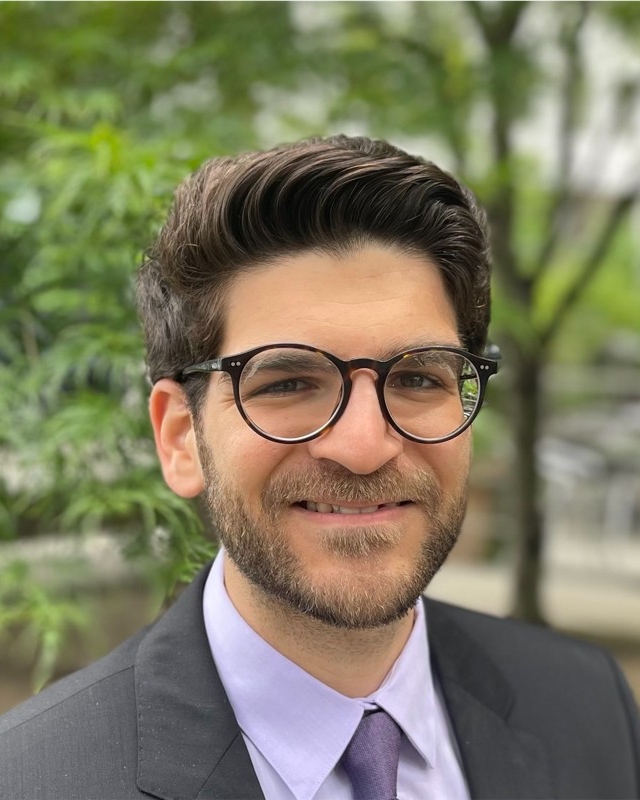}}]
{Amr Abdelraouf}
(Member, IEEE) received his Ph.D. degree in Transportation Engineering from the University of Central Florida in 2022. He was a research scientist at Toyota Motor North America, InfoTech Labs. His research interests include applied AI/ML, advanced driver assistance systems, connected and automated vehicles, intelligent transportation systems, and digital twins.
\end{IEEEbiography}
\vspace{-15pt}

\begin{IEEEbiography}
[{\includegraphics[width=1in,height=1.25in,clip,keepaspectratio]{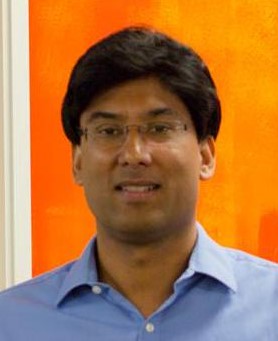}}]
{Rohit Gupta}
received the Ph.D. degree in Mechanical Engineering from the University of California at Santa Barbara. He is currently a principal researcher at Toyota Motor North America, InfoTech Labs. Prior to joining Toyota, he worked in various industries: semiconductors, telecommunications, finance, and software. His research interests include cloud and edge computing architectures, connected and autonomous vehicles, AI/ML, ``Digital Twin'' of human-vehicle interactions, and high-speed mobility.  
\end{IEEEbiography}
\vspace{-15pt}

\begin{IEEEbiography}
[{\includegraphics[width=1in,height=1.25in,clip,keepaspectratio]{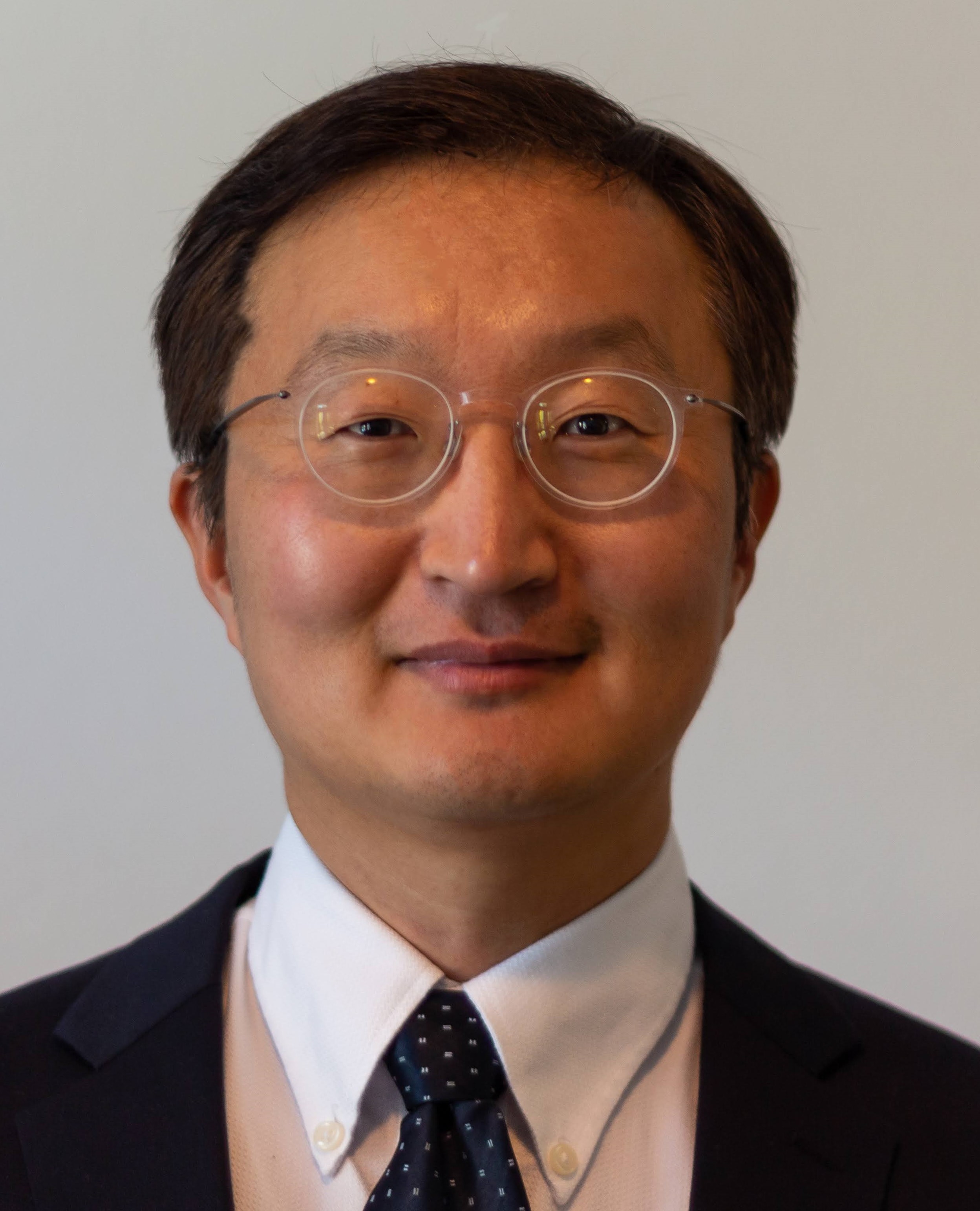}}]
{Kyungtae (KT) Han}
(Senior Member, IEEE) received the Ph.D. degree in Electrical and Computer Engineering from The University of Texas at Austin in 2006. He is currently a senior principal scientist at Toyota Motor North America, InfoTech Labs. Prior to joining Toyota, Dr. Han was a Research Scientist at Intel Labs and a Director at Locix Inc. His research interests include cyber-physical systems, connected and automated vehicle techniques, and intelligent transportation systems.  
\end{IEEEbiography}
\vspace{-15pt}

\begin{IEEEbiography}
[{\includegraphics[width=1in,height=1.25in,clip,keepaspectratio]{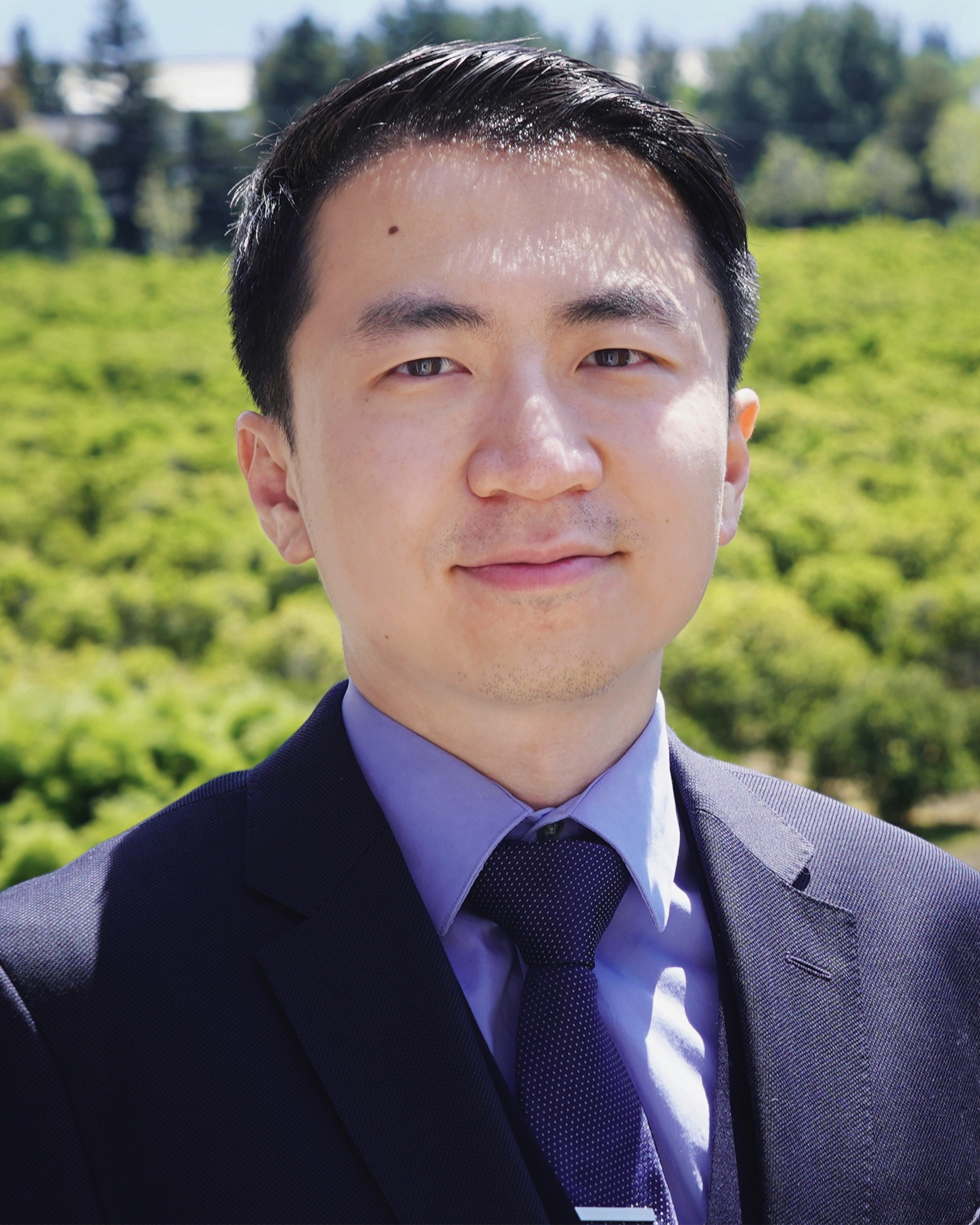}}]
{Ziran Wang}
(Senior Member, IEEE) received his Ph.D. degree in Mechanical Engineering from the University of California, Riverside in 2019. He is an assistant professor in the College of Engineering at Purdue University, and was a principal researcher at Toyota Motor North America R\&D. His research focuses on automated driving, human-autonomy teaming, and digital twins. He is the Founding Chair of the IEEE Technical Committee on Internet of Things in Intelligent Transportation Systems.
\end{IEEEbiography}
\vspace{-10pt}

\end{document}